\documentclass[sn-mathphys-num]{sn-jnl}

\usepackage[T1]{fontenc}
\usepackage{graphicx}%
\usepackage{multirow}%
\usepackage{amsmath,amssymb,amsfonts}%
\usepackage{amsthm}%
\usepackage{mathrsfs}%
\usepackage[title]{appendix}%
\usepackage{xcolor}%
\usepackage{textcomp}%
\usepackage{manyfoot}%
\usepackage{booktabs}%
\usepackage{algorithm}%
\usepackage{algorithmicx}%
\usepackage{algpseudocode}%
\usepackage{listings}%
\usepackage{fullpage}
\usepackage{setspace}
\usepackage{parskip}
\usepackage{titlesec}
\usepackage[section]{placeins}
\usepackage{xcolor}
\usepackage{breakcites}
\usepackage{lineno}
\usepackage{hyphenat}
\usepackage{pifont}
\usepackage{caption}
\usepackage{array}
\usepackage{makecell}
\usepackage{multirow}
\usepackage{float}
\usepackage{tikz}
\usepackage{graphicx}

\theoremstyle{thmstyleone}%
\theoremstyle{thmstyletwo}%

\theoremstyle{thmstylethree}%

\newcolumntype{P}[1]{>{\centering\arraybackslash}p{#1}}
\def\checkmark{\tikz\fill[scale=0.4](0,.35) -- (.25,0) -- (1,.7) -- (.25,.15) -- cycle;} 

\begin{document}

\title[History, Development, and Principles of Large Language Models—An Introductory Survey]{History, Development, and Principles of Large Language Models—An Introductory Survey}











\author[1]{\fnm{Zichong} \sur{Wang}\textsuperscript{1}}

\author[1]{\fnm{Zhibo} \sur{Chu}\textsuperscript{1}}

\author[1]{\fnm{Thang Viet} \sur{Doan}\textsuperscript{1}}

\author[2]{\fnm{Shiwen} \sur{Ni}\textsuperscript{2}}

\author[2]{\fnm{Min} \sur{Yang}\textsuperscript{2}}

\author{\fnm{Wenbin} \sur{Zhang} \textsuperscript{1}\,\footnote{Corresponding author. Email:\{ziwang, tdoan011, wenbin.zhang\}@fiu.edu}} 

\affil[1]{\orgname{Florida International University}, \orgaddress{\street{11200 SW 8th Street}, \city{Miami}, \postcode{33199}, \state{FL}, \country{USA}}}

\affil[2]{\orgname{Shenzhen Institutes of Advanced Technology, Chinese Academy of Sciences}, \orgaddress{\city{Shenzhen}, \state{Guangdong}, \country{China}}}


\abstract{Language models serve as a cornerstone in natural language processing (NLP), utilizing mathematical methods to generalize language laws and knowledge for prediction and generation. Over extensive research spanning decades, language modeling has progressed from initial statistical language models (SLMs) to the contemporary landscape of large language models (LLMs). Notably, the swift evolution of LLMs has reached the ability to process, understand, and generate human-level text. Nevertheless, despite the significant advantages that LLMs offer in improving both work and personal lives, the limited understanding among general practitioners about the background and principles of these models hampers their full potential. Notably, most LLM reviews focus on specific aspects and utilize specialized language, posing a challenge for practitioners lacking relevant background knowledge. In light of this, this survey aims to present a comprehensible overview of LLMs to assist a broader audience. It strives to facilitate a comprehensive understanding by exploring the historical background of language models and tracing their evolution over time. The survey further investigates the factors influencing the development of LLMs, emphasizing key contributions. Additionally, it concentrates on elucidating the underlying principles of LLMs, equipping audiences with essential theoretical knowledge. The survey also highlights the limitations of existing work and points out promising future directions.}




\maketitle

\section*{1. Introduction}\label{sec1}

Language stands as humanity's most potent tool, enabling the expression of thoughts and emotions and facilitating communication with others~\cite{pinker1994language,turing2000computing}. However, machines lack the intrinsic ability to comprehend and communicate in human language, necessitating powerful Artificial Intelligence (AI) algorithms to acquire such capabilities. The ultimate aim of AI research is the creation of machines capable of reading, writing, and conversing like humans~\cite{dwivedi2023so,zhao2023survey,jin2023rethinking}. Natural language processing (NLP) emerges as the AI branch dedicated to realizing this goal. At the core of NLP are language models (LMs), tasked with predicting or generating the probability or likelihood of linguistic units (\textit{e.g.,} words, phrases, sentences) based on context. The evolutionary stages of LMs unfold chronologically from the initial statistical language models (SLMs) to subsequent neural language models (NLMs), progressing to pre-trained language models (PLMs), and reaching the current state of large language models (LLMs). Specifically, SLMs employ simple probability distributions to model word sequences, while NLMs utilize neural networks to grasp complex patterns and representations of language. PLMs leverage large-scale corpora and self-supervised learning to capture general linguistic knowledge, and LLMs extend PLMs by incorporating massive data, computation, and algorithms, resulting in more expressive, sophisticated, and adaptable LMs~\cite{fu2022does,wei2022emergent}.
 
\quad In recent years, significant advancements in NLP have been achieved with LLMs, such as the generative pre-trained transformer (GPT) series of models~\cite{radford2018improving,radford2019language,brown2020language,achiam2023gpt}. Extensively trained on textual data, these models demonstrate the ability to generate human-level texts and perform language-based tasks with exceptional precision. Despite these advancements having greatly benefited various aspects of work and life, LLMs might not be widely understood by practitioners, particularly those without a background in NLP. To provide practitioners with a fundamental understanding of LLMs, this review introduces it across six dimensions: history, development, principles, applications, drawbacks and future directions. This survey distinguishes itself from other reviews on PLMs or LLMs~\cite{zhao2023survey,liu2023pre,han2021pre,shanahan2024talking}, which often concentrate on specific aspects using technical language; this survey aims to elucidate LLMs and their principles in a more accessible manner, intending to maximize its full potential.

\quad To achieve a comprehensive survey of LLMs, we employed a meticulous review process guided by a structured search strategy. We searched databases such as IEEE Xplore, ACM Digital Library, Scopus, and Google Scholar using keywords including ``large language models'', ``large language models survey'', and ``large language models for \textit{X}'' where \textit{X} includes software engineering, drug discovery, finance, medical, legal, and education. Our review focused on publications from major conferences and journals about NLP and LLMs' applications, including EMNLP~\cite{dodge2021documenting, sonkar2023class, kim2021changes, tay2023transcending}, ACL~\cite{ahmad2021unified, muennighoff2023crosslingual, bang2023multitask, sheng2021societal}, ICLR~\cite{fried2022incoder, nijkamp2022codegen, wei2021finetuned, sanh2022multitask, tay2022ul2, zeng2022glm, he2020deberta, merity2022pointer}, ICML~\cite{du2022glam, biderman2023pythia, wang2022language, yu2023codeipprompt}, ICSE~\cite{steenhoek2023empirical, yin2024improving}, ICAIF~\cite{li2023large, pagliaro2021investor}, ICDM~\cite{zhang2021fair, wang2023mitigating, chinta2023optimization}, IJCAI~\cite{zhang2019faht,zhang2021farf,saxena2024unveiling}, AAAI~\cite{zhang2022longitudinal,zhang2024fairness,zhang2023censored}, ECAI~\cite{zhang2023individual,wang2024group, wang2024individual1}, BEA~\cite{xiao2023evaluating}, Financial Innovation~\cite{gupta2020comprehensive}, PLOS Digital Health~\cite{kung2023performance, mozafari2020hate}, Applied Sciences~\cite{jin2021disease}, Interspeech~\cite{kombrink2011recurrent, mikolov2010recurrent, stolcke2002srilm}, Journal of Machine Learning Research~\cite{chung2024scaling, chowdhery2023palm, fedus2022switch, raffel2020exploring}, Journal of AI~\cite{baidoo2023education}, etc., with an emphasis on studies from the past decade. Additionally, we reviewed references from selected papers and cross-checked our findings against existing LLMs surveys~\cite{zhao2023survey, chen2024survey, yao2024survey, huang2023survey}. Our process involved identifying relevant studies and applying screening criteria based on LLMs relevance, citation count, and the quality of the publication venue. Papers with minimal contributions or lower relevance were excluded. The remaining studies were then evaluated for methodological rigor and clarity of conclusions. Key data from these studies were extracted and synthesized to highlight trends, gaps, and emerging themes in LLM research. This synthesis offers a comprehensive overview of current advancements, ongoing challenges, and future research directions, providing valuable insights for both academic and industrial applications.

\quad The subsequent sections of this review are structured as follows: Section 2 details the history of LLMs and the contributing factors behind their rapid growth, covering the evolution of LMs, increased data diversity, computational advancements, and algorithmic innovations. Section 3 provides an overview of the principles of LLMs, using the GPT family of models as an accessible example to enhance understanding of LLMs' principles. Additionally, we present a comparison of state-of-the-art models across various types to assist readers in selecting the most suitable model for their specific applications. Moving on, Section 4 explores the wide-ranging applications of LLMs across various industries and professional domains. Section 5 critically examines the drawbacks of current state-of-the-art LLMs, and finally, Section 6 concludes the survey.

\section*{2. HISTORY AND DEVELOPMENT OF LARGE LANGUAGE MODELS}
{\label{chap:2}}

In this section, we will explore the history of the LMs and analyze the developmental dynamics and influences shaping LLMs. 
\subsection*{2.1 History of LLMs}
\begin{figure}[H]
    \centering
    \includegraphics[width=0.95\linewidth]{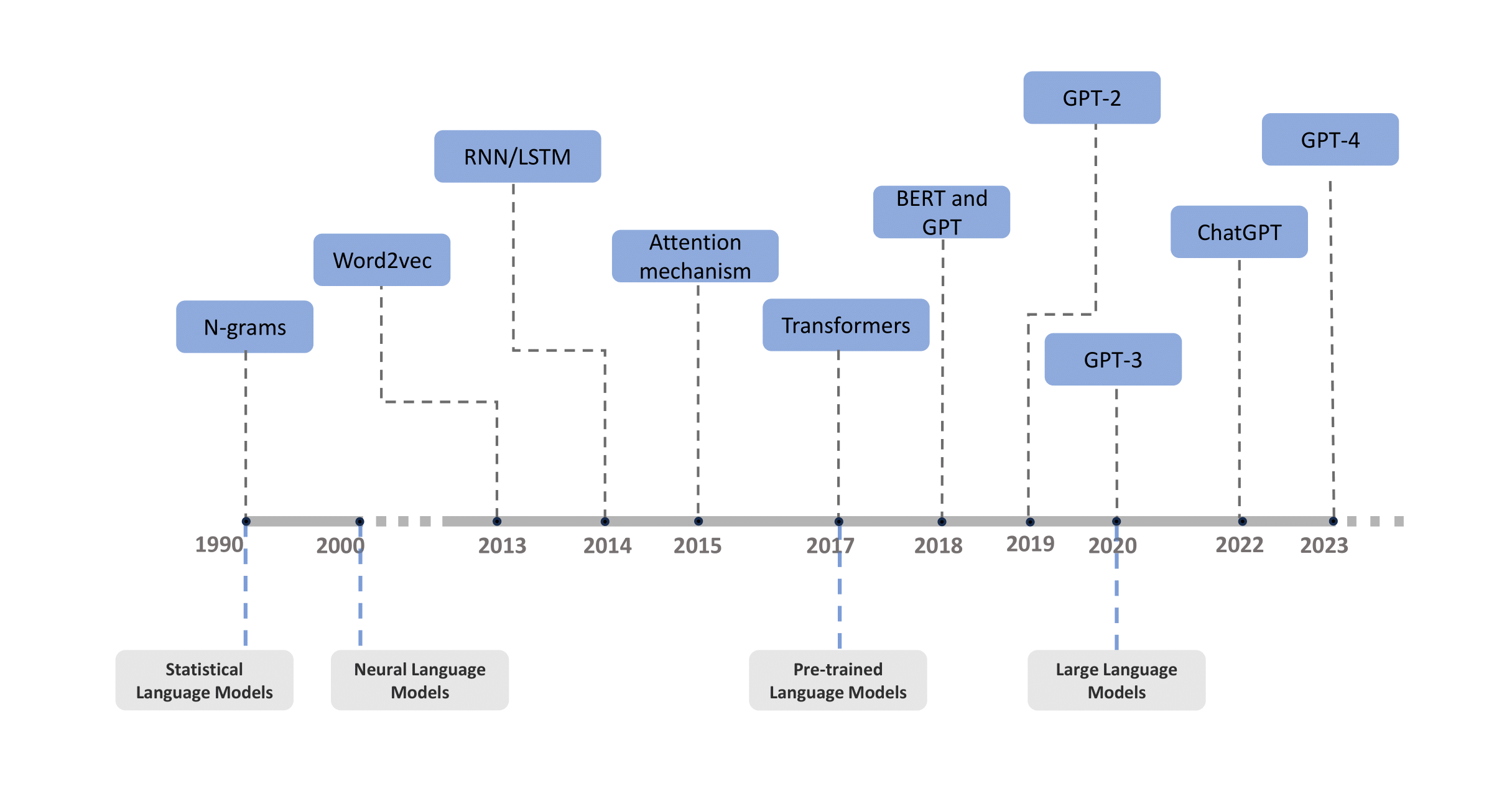}
    \caption{History and development of language models.}
    \label{fig:1}
\end{figure}

{\label{chap:2.1}}
To better understand LLMs, this section will follow the developmental stages of LMs and introduce SLMs, NLMs, PLMs, and LLMs. Figure \ref{fig:1} provides a visual map of the history of the development of the LMs.

\begin{itemize}
    \item Statistical Language Models: SLMs~\cite{jelinek1998statistical,stolcke2002srilm,rosenfeld2000two} originated in the 1990s as mathematical models addressing contextually relevant properties of natural language from a probabilistic statistical perspective. The essence of statistical language modeling lies in ascertaining the probability of a sentence occurring within a text. Considering $S$ as the sentence ``I am very happy'', $P(\omega_i)$ signifies the probability of the $i$-th word in the sentence: $\omega_1$ as ``I'', $\omega_2$ as ``am'', $\omega_3$ as ``very'', and $\omega_4$ as ``happy''. Now, the objective is to ascertain the likelihood of $S$ appearing in the text, denoted as $P(S)$:
    
   \begin{equation}
    P(S)=P(\omega_1,\omega_2,\omega_3,\omega_4)=P(I,am,very,happy)
   \end{equation}
   
To calculate this probability, the conditional probability can be employed: 
 \begin{equation}
    P(I,am,very,happy)=P(I)\cdot P(am \mid I)\cdot P(very \mid I,am)\cdot P(happy \mid I,am,very)
\end{equation}
\noindent where $P(I)$ represents the probability of the word ``I'' appearing and $P(am|I)$ stands for the probability of ``am'' appearing given that ``I'' has appeared. When we multiply $P(am|I)$ by $P(I)$, it fulfills the condition of ``I'' appearing in $P(am|I)$, resulting in the probability of ``I am'' appearing together as $P(I, am) = P(I) \cdot P(am|I)$. Now, the question arises: how do we calculate the conditional probability of the occurrence of each word? The answer lies in Maximum Likelihood Estimation, enabling us to estimate probabilities by substituting them with frequencies when the sample size is sufficiently large, given by:
    \begin{equation}
    P(w_{i}|w_{1}w_{2}\cdots w_{i-1})=\frac{P(w_{1}\cdots w_{i-1}w_{i})}{P(w_{1} w_{2}\cdots w_{i-1})}=\frac{{\bf C}(w_{1}w_{2}\cdots w_{i})}{{\bf C}(w_{1}w_{2}\cdots w_{i-1})}
    \end{equation}
\noindent where $\mathbf{C(\cdot)}$ represents the count of occurrences of the subsequence in the training set. Using this formula, we can calculate the likelihood of each word as the $i$-th word given the preceding $i-1$ words. Then, we select the $i$-th word by choosing the word associated with the highest probability. The previous discussion assumes that the $n$-th word is related to the initial $n-1$ words, consistent with our intuition and commonly referred to as the n-gram model. In order to efficiently compute conditional probabilities, it is necessary to pre-compute and save $C(X)$ required for the conditional probability computation, where $X$ is a sentence of length $n$. The number of possible sentences $X$ grows exponentially with the size of the vocabulary. For instance, with $1000$ different words, there exist ${1000}^n$ potential sequences of length $n$. However, excessively large values of $n$ pose storage limitations. Typically, $n$ is confined to $2$ or $3$, causing each word to relate to only its first $1$ or $2$ preceding words, ultimately leading to a reduction in the model's accuracy.

    \item Neural Language Models: NLMs~\cite{bengio2000neural,mikolov2010recurrent,kombrink2011recurrent} leverage neural networks to predict the probabilities of subsequent words within sequences. They effectively handle longer sequences and mitigate the limitations associated with small $n$ in SLMs. Before delving into neural networks, let's grasp the concept of word vectors. Humans effortlessly comprehend word meanings. For instance, ``cat'' and ``dog'' share closer semantic connections than ``cat'' and ``tree'' since they both represent animals. But how does a computer accomplish this? Computers operate using binary code — 0s and 1s — so human language needs translation into binary. Word vectors can accomplish this, which are numerical representations of human language, where each word corresponds to a distinct vector. These vectors usually possess fixed dimensions and can simulate word relationships through the angles between them. An intriguing example is the angle between the word vectors for ``cat'' and ``dog'' which is smaller than the angle between the word vector for ``cat'' and ``tree''. Another illustrative example is ``$Messi - Argentina + Portugal = Cristiano\ Ronaldo$''. Word2Vec~\cite{mikolov2013distributed} is a widely recognized tool for computing word vectors. Its function involves converting words into dense numerical representations, making words with similar semantics closer together in the vector space. 
   
    \begin{figure}[h]
        \centering
        \includegraphics[width=0.75\linewidth]{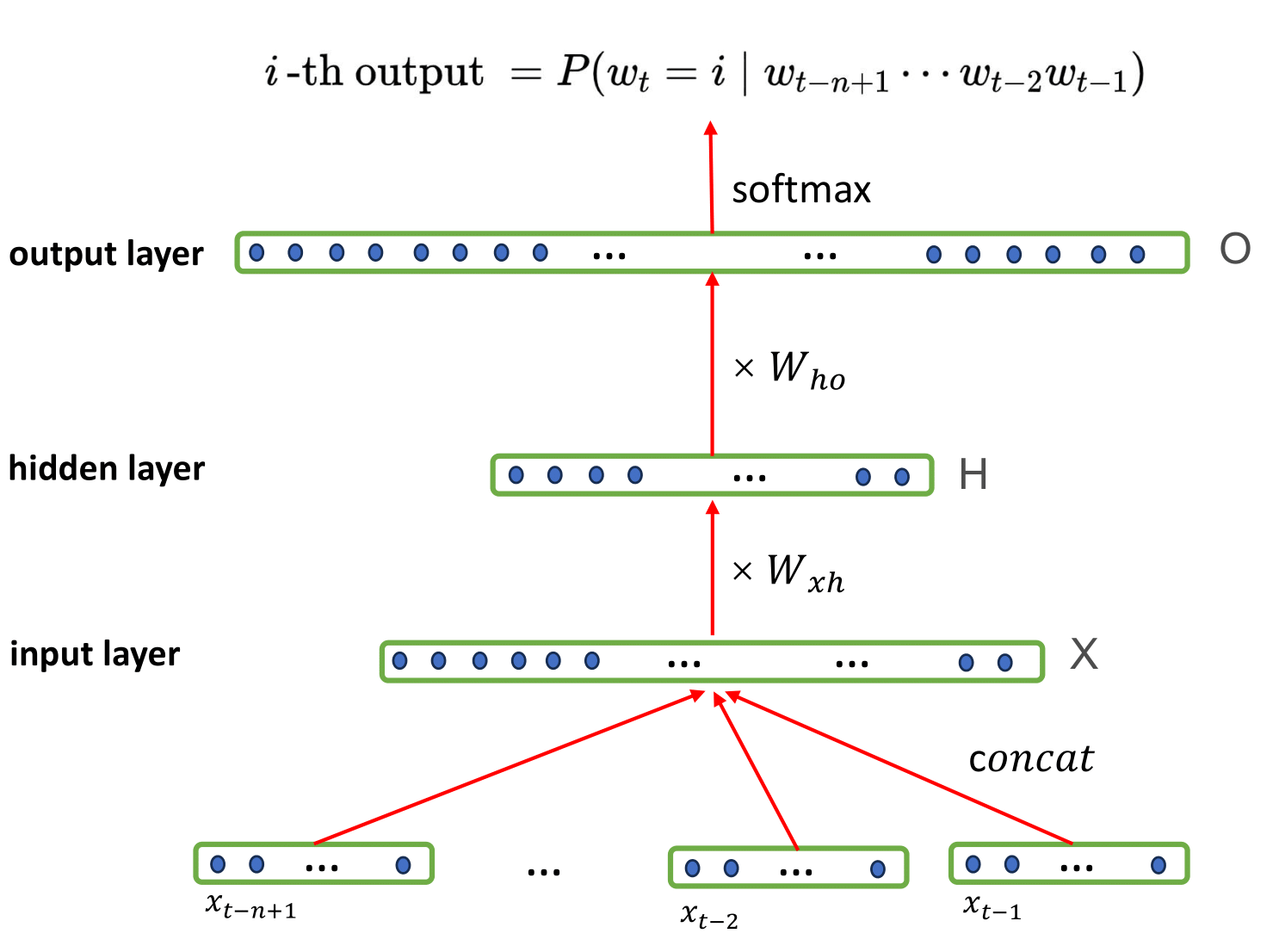}
        \caption{Neural language model.}
        \label{fig:2}
    \end{figure}
    
\quad The efficacy of Word2Vec is notably influenced by its utilization of neural networks, which mirrors the structure and functionality of the human brain, comprising interconnected simple units known as neurons organized into different layers. Figure \ref{fig:2} provides a visualization of a simple NLM structure, composed of input, hidden, and output layers. Within this structure, each word vector $x_{t}\in \mathbb{R}^{m}$, the matrix $W_{xh}\in \mathbb{R}^{m(n-1) \times h}$, and the matrix $W_{ho}\in \mathbb{R}^{h \times |V|}$, where $h$ signifies the number of neurons in the hidden layer, and $V$ is the vocabulary containing all words recognized by the model. NLMs operate akin to the n-gram concept, assuming that the probability of each word depends only on its previous n-1 words. The first layer—the input layer—concatenates the word vectors of n-1 words, forming a vector $X \in \mathbb{R}^{m \times (n-1)}$. Subsequently, the second layer—the hidden layer—derives the output $H_{}$ by applying a non-linear activation function, like sigmoid or tanh, to the matrix product of $X_{}$ and $W_{xh}$. Following this, the third layer—the output layer—aims to forecast the subsequent word based on the hidden layer's output. With $|V|$ neurons within the output layer, the resulting output vector $O \in \mathbb{R}^{|V|}$ is computed by multiplying $H_{}$ and $W_{ho}$. This resultant vector then undergoes the Softmax function, producing a vector $O’_{}$ containing the probability value assigned to each word. The Softmax function, concerning the output of the $i$-th neuron, is defined as follows:

\begin{equation}
O_{i}^{\prime}=\operatorname{Softmax}\left(O’_{i}\right)=\frac{\exp \left(O_{i}\right)}{\sum_{j=1}^{|V|} \exp \left(O_{j}\right)},
\end{equation}
where $ O_{i}^{\prime} $ denotes the output value of the $i$-th node, and $|V|$ represents the count of output nodes, corresponding to the classification categories. Utilizing the Softmax function allows the transformation of output values from multiclass classification into a probability distribution, ensuring a sum of 1 within the range [0, 1]. 
    
    \item Pre-trained Language Models: PLMs undergo initial training using an extensive volume of unlabeled text, enabling them to grasp fundamental language structures such as vocabulary, syntax, semantics, and logic — a phase termed pre-training. Subsequently, this comprehensive LM can be applied to various NLP tasks like machine translation, text summarization, and question-answering systems. To optimize its performance, models need to be trained a second time on a smaller dataset customized for a specific downstream task — a phase known as fine-tuning. This is the ``pre-training and fine-tuning'' learning paradigm. We can use a visual example to understand the ``pre-training and fine-tuning'', as follows: in martial arts novels, a person who wants to become a martial arts master needs to have a solid foundation of internal martial arts, which can be acquired by training on a large variety of techniques. Then, the person can learn a specific skill, such as a sword or a palm strike, and master it quickly and effectively by applying internal martial arts principles. A large number of studies on PLMs have been built on this paradigm, which introduces different architectures~\cite{devlin2018bert,fedus2022switch} (\textit{e.g.,} GPT-2~\cite{radford2019language} and Bert~\cite{devlin2018bert}).

    \item Large Language Models: LLMs are trained on massive text corpora with tens of billions (or more) of parameters, such as GPT-3~\cite{brown2020language}, GPT-4~\cite{achiam2023gpt}, PaLM~\cite{chowdhery2023palm}, and LLaMA~\cite{touvron2023llama}. The goal of LLMs is to enable machines to understand human commands and adherence to human values. Their hallmark lies in the consolidation of two stages: initial pre-training on a vast general-purpose corpus followed by alignment with human values, rather than transitioning to a different domain. LLMs exhibit remarkable adaptability compared to PLMs, transitioning from specialized to general-purpose models. The substantial increase in model size, dataset volume, and computational prowess has resulted in significant enhancements across various tasks and unveiled remarkable capabilities absent in smaller models. For example, GPT-3 has the capability to leverage contextual information, a functionality that GPT-2~\cite{radford2019language} lacks. This means that when GPT-3 is prompted with task-related examples, it utilizes them to enhance its problem-solving capabilities. The number of parameters for LLMs typically exceeds a hundred billion, and the training data is usually in the range of a few hundred GB to a few TB. A concrete example is that there are multiple versions of the GPT-2 model, with the largest version having 1.5 billion parameters and using 40GB of text data for training. In contrast, the largest version of the GPT-3 model has 175 billion parameters and uses 570GB of text data for training. This example illustrates the significant discrepancy between LLMs and PLMs concerning parameter count and training data volume. 
    
\end{itemize}

\subsection*{2.2 Factors Propelling Large Language Models}

{\label{chap:2.2}}
In the past few years, various factors have played a significant role in the swift advancement of LLMs:

\begin{itemize}
    \item Data Diversity: Data diversity has emerged as a crucial catalyst for the development of LLMs. Recent years have witnessed a surge in the availability and diversity of extensive internet-based data sources, furnishing these models with training data containing rich linguistic and worldly insights. The quality and origins of this data significantly influence the efficacy and capacity of LLMs.  Earlier studies trained language models on a single domain of text, such as news articles~\cite{jozefowicz2016exploring}, Wikipedia~\cite{merity2022pointer}, or fiction books~\cite{kiros2015skip}. However, the proliferation of vast and heterogeneous text corpora across the web—including blogs, social media, forums, and reviews, among others—has spurred researchers to pivot towards training models on multi-domain texts. This shift enhances the model's aptitude for generalization and multitasking. An exemplary instance is Meta AI's open-source large language model, LLaMA~\cite{touvron2023llama}, trained on publicly accessible datasets like CommonCrawl\footnote{https://commoncrawl.org/get-started}, C4~\cite{raffel2020exploring}, Github\footnote{https://github.com/}, Wikipedia\footnote{https://huggingface.co/datasets/wikipedia}, Books~\cite{gao2020pile}, ArXiv\footnote{https://arxiv.org/}, and StackExchange\footnote{https://stackexchange.com/}.
    
    \item Computational Advancement: LLMs are neural network models with a large number of parameters. With the progression of deep learning (DL) technology in recent years, the scale and effectiveness of LLMs have notably escalated. Simultaneously, they present serious challenges due to their substantial computational demands~\cite{kaplan2020scaling}. Computational power signifies a computer system's capability to execute computational tasks, often quantified in terms of floating-point operations per second. According to OpenAI calculations\footnote{https://openai.com/research/ai-and-compute}, since 2012, the amount of computation used for global AI training has shown exponential growth, doubling every 3.43 months on average, and now the amount of computation has expanded 300,000 times, far outpacing the rate of computational power growth. Fortunately, hardware innovators like Nvidia continually invent specialized devices such as GPUs, TPUs, and NPUs with amplified computational prowess, LLMs can be trained quickly. For instance, the launch of GPT-3 by OpenAI in May 2020~\cite{brown2020language} underscores this reliance on robust hardware. Training GPT-3 on a single NVIDIA Tesla V100 GPU would require an estimated 355 years. Conversely, leveraging 1024×A100 GPUs, researchers estimated the potential to train GPT-3 in as little as 34 days. This exemplifies the indispensability of potent hardware in facilitating the development and training of LLMs.
    
    \item Algorithmic Innovation: Algorithmic innovation stands as the pivotal driving force behind LLMs, shaping their structure and functionality. From the initial rule- and statistics-based approach to the later DL-based approach, the algorithms of LMs have continuously evolved and improved. Presently, all LLMs are based on the transformer architecture~\cite{vaswani2017attention}, which employs a self-attention mechanism (see discussion in Section 3). This architecture offers distinct advantages over traditional recurrent~\cite{mikolov2010recurrent} and convolutional neural networks~\cite{lecun1998gradient}, enabling fully parallel computation and adeptly capturing long-distance dependencies. Moreover, there exist numerous iterations and enhancements of the transformer, such as Transformer-XL~\cite{cui2023chatlaw}, XLNet~\cite{yang2019xlnet}, ALBERT~\cite{lan2019albert}, each targeting specific aspects for optimization. These variations aim to enhance the attention mechanism, refine pre-training objectives, and curtail parameter counts, thereby advancing the transformer architecture in diverse dimensions.

\end{itemize}
\par\null
\section*{3. PRINCIPLES AND TAXONOMY OF LARGE LANGUAGE MODELS}   
\label{chap:3}
This section strives to clarify the fundamental principles and taxonomy of LLMs, intending to provide general practitioners with a more profound understanding of the operational mechanisms employed by these models and help readers choose the most suitable LLMs for their specific applications. 

\begin{figure}[h]
    \centering
\includegraphics[width=0.85\linewidth]{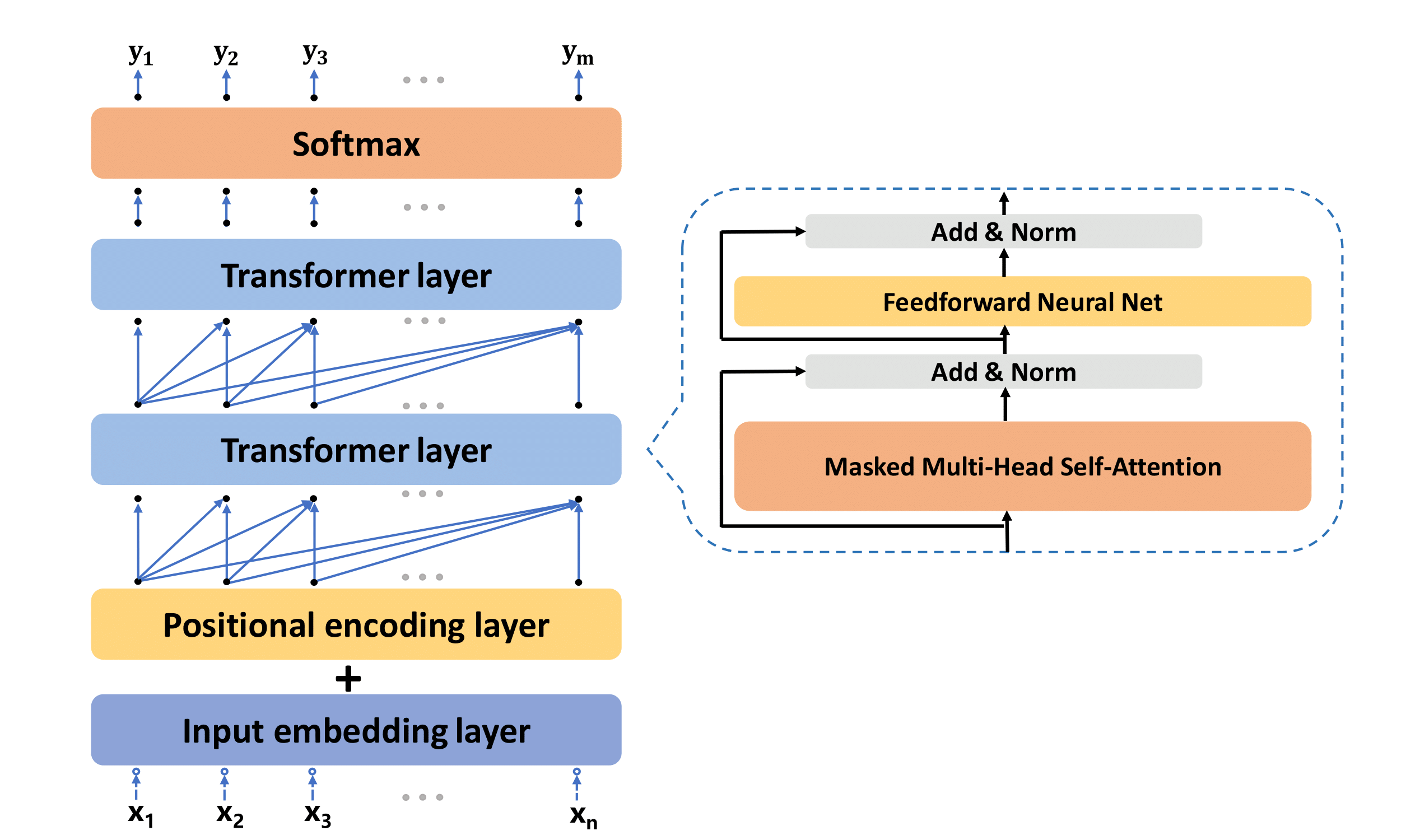}
    \caption{The GPT-3 model architecture. Note that only two transformer layers are illustrated for simplicity and illustrative purposes, and the actual model consists of 12 transformer layers.}
    \label{fig:3}
\end{figure}

Fundamental to LLMs is the utilization of deep neural networks to grasp the statistical patterns and semantic nuances of language, enabling the understanding and generation of language. The GPT series, a notable example among these models, incorporates the transformer architecture~\cite{vaswani2017attention} and utilizes autoregression to predict consecutive words. The following discussion will concentrate on the GPT-3~\cite{brown2020language} as an exemplar, providing insights into multiple facets that elucidate the principles of LLMs, as illustrated in Figure \ref{fig:3}.

\begin{itemize}
    \item Input: The GPT model architecture takes a sequence of symbol representations $(x_1, \ldots, x_n)$ as input. This sequence comprises $n$ words, also known as tokens, with GPT-3 explicitly defining the input sequence length as 2048 words.

    \item Encoding: The significance of encoding words into vectors is heightened by the fact that the machine learning model operates on numeric inputs. Within the input embedding layer, the initial step involves constructing a comprehensive vocabulary that encompasses all words and assigning each word a unique position within this established vocabulary. For instance, ``a'' could be designated as 0, ``an'' as 1, and so forth. In the case of GPT-3, its vocabulary encompasses 50,257 words. Following this, every word can be transformed into a one-hot vector of 50,257 dimensions, where only a single element has the value of one while the rest are zeros. For example, ``a'' would be represented as [1,0,0,0,…], and ``an'' as [0,1,0,0,0,…]. This process leads to the encoding of the input sequence into a matrix denoted as $I_{E}\in \mathbb{R}^{2048\times 50257}$.

    \item Embedding: The encoding process above transforms every input word into a one-hot vector of 50,257 dimensions. However, this vector exhibits high sparsity, primarily consisting of zeros, leading to significant space inefficiency. To address this drawback, the subsequent step in the input embedding layer employs an embedding matrix, denoted as $W_{E}$, to condense these 50,257-dimensional input vectors into shorter numeric vectors of length $n$ ($n$ is set to 12,288 for GPT-3). This method condenses information about word meanings into a more compact space, effectively reducing the vector's length. Specifically,  the transformation is represented as: $X_{WordEmbedding} = I_{E} \times W_{E}$, where $X_{WordEmbedding}\in \mathbb{R}^{2048\times 12288}$, $I_{E}\in \mathbb{R}^{2048\times 50257}$ and $W_{E}\in \mathbb{R}^{50257\times 12288}$.

    \item Positional Encoding: The position and order of elements in a sequence are very important for understanding and generating sequences. Demonstrated through the sentences, ``He is a good person and does not do bad things'' and ``He is a bad person and does not do good things'', a slight alteration in word order - the transition from ``good'' to ``bad'' - results in a significant change in the sentence's semantic meaning. Within the architectural framework of the GPT model, the positional encoding layer serves as a complement to the input embedding layer. This augmentation facilitates the transformer layer in comprehending both the positional and sequential cues within the data. GPT-3 employs the following implementation for positional encodings:
    \begin{equation}
    P E(\text {pos},\text{ 2i}) = \sin \left(\frac{\text {pos}}{10000^{2 i / d_{\text {model}}}}\right)
    \end{equation}
    \begin{equation}
    P E(\text {pos},\text{ 2i+1})=\cos \left(\frac{\text {pos}}{10000^{2 i / d_{\text {model}}}}\right)
    \end{equation}
\noindent where $pos$ is the position of the current word within the sentence (\textit{e.g.,} in the sentence ``A survey of Large Language Model'', $pos_{A}$=0, $pos_{Large}$=3), $d_{model}$ is the length of the embedding vector 12288 and $i$ is the dimension. However, why delve into such intricate computations for positional encodings? Let's explore a more straightforward approach to conceptualize it. Initially, considering a text of length $T$, the most basic approach of positional encoding is counting. This method utilizes $[0,1,2...T-1]$ as positional encodings for each word in the text. For instance, the positional encodings for the $2$-th word would be $[1,1,...,1]$, aligning in length with the embedding vector. However, this encoding strategy presents a notable drawback: within longer sentences, subsequent words hold considerably larger positional encoding values compared to the positional encodings of the initial words. As a result, this disparity inevitably introduces numerical biases in the merged features when combined with word embeddings. To address this issue, normalization becomes a viable consideration. The most straightforward normalization method involves dividing directly by $T$, resulting in positional encodings such as $[0, \frac{1}{T}, \ldots, \frac{T-1}{T}]$ (\textit{e.g.,} the 2nd word's positional encodings  = $[\frac{1}{T}, \frac{1}{T}, \ldots, \frac{1}{T}]$). While this method restricts all positional encodings to the [0,1] range, it introduces a significant inconsistency: the positional values are contingent on the text's length, causing the positional distinctions between adjacent words in shorter texts to differ from those in longer texts. As illustrated in Figure \ref{fig:4}, the positional encodings discrepancy between adjacent word positions in sentence \ding{172} is 0.5, while in sentence \ding{173}, neighboring word positions are coded with a difference of 0.125. Consequently, neither of these methods proves to be satisfactory. To address this issue, Google devised an alternative approach involving trigonometric functions~\cite{vaswani2017attention}. As a result, a positional encoding matrix denoted as $X_{ PositionalEncoding}\in \mathbb{R}^{2048 \times 12288}$.

    \FloatBarrier
    \begin{figure}[h]
        \centering
\includegraphics[width=0.90\linewidth]{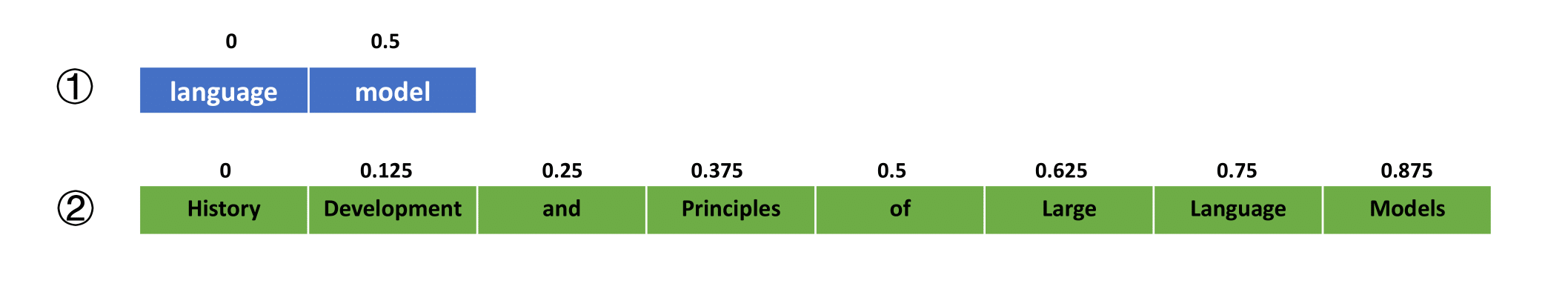}
        \caption{Incorrect position encoding.}
        \label{fig:4}
    \end{figure}

    \item Input Matrix: The input sequence $(x_1, \ldots, x_n)$ undergoes processing through both the input embedding layer and the positional embedding layer to generate the input matrix.  Specifically, input matrix $X_{}$ $=$ embedding matrix $X_{WordEmbedding}$ $+$ positional encoding matrix $X_{PositionalEncoding}$, where $X\in \mathbb{R}^{2048\times 12288}$, $X_{WordEmbedding}\in \mathbb{R}^{2048\times 12288}$ and $X_{PositionalEncoding}\in \mathbb{R}^{2048\times 12288}$.
    
    \item Masked Multi-Head Self-Attention: Masked multi-head self-attention involves combining a masking technique and a multi-head mechanism within the framework of self-attention. The following discussion commences with self-attention before delving into masked multi-head self-attention for enhanced comprehension.
   
    \quad Self-Attention: Attention is a crucial cognitive function for human beings, allowing us to concentrate on significant aspects within a complex environment. Within the transformer layer, self-attention mirrors this cognitive functionality by forecasting which input words warrant attention for every output in the sequence, subsequently assessing their significance. For instance, when analyzing the input ``I am very'' and predicting the output ``happy'', the relative contributions of the three words ``I am very'' to the output ``happy'' might be quantified as 0.7, 0.1, and 0.2, respectively. The specific steps are as follows:

    Step 1: Initiation commences with the creation of three projection matrices $W_{q}\in \mathbb{R}^{12288 \times 128}$, $W_{k}\in \mathbb{R}^{12288 \times 128}$, and $W_{v}\in \mathbb{R}^{12288 \times 128}$. These matrices are subsequently multiplied with the input matrix $X\in \mathbb{R}^{2048 \times 12288}$ to obtain three different matrices $Q$, $K$, and $V$, which respectively represent the query, key, and value.

    Step 2: Following this, a matrix is derived using the formula $Softmax~(Q \times K^T)$. This matrix then multiplied with $V_{}$ to obtain the output $Y$, represented as $Y=Softmax~(Q\times K^T)\times V$, where $Q\in \mathbb{R}^{2048\times 128}$, $K^T\in \mathbb{R}^{128\times2048}$, $V\in \mathbb{R}^{2048\times 128}$, and $Y\in \mathbb{R}^{2048\times 128}$. An illustrative analogy can be drawn to searching for a song within a music software - here, $Q$ denotes the queried song's name, $K$ encompasses all song names within the software database, and $V$ includes the corresponding audio data of all songs in the software. Thus, $Y$ signifies the audio data of the sought-after song. 
    
    \begin{figure}[h]
        \centering
    \includegraphics[width=0.95\linewidth]{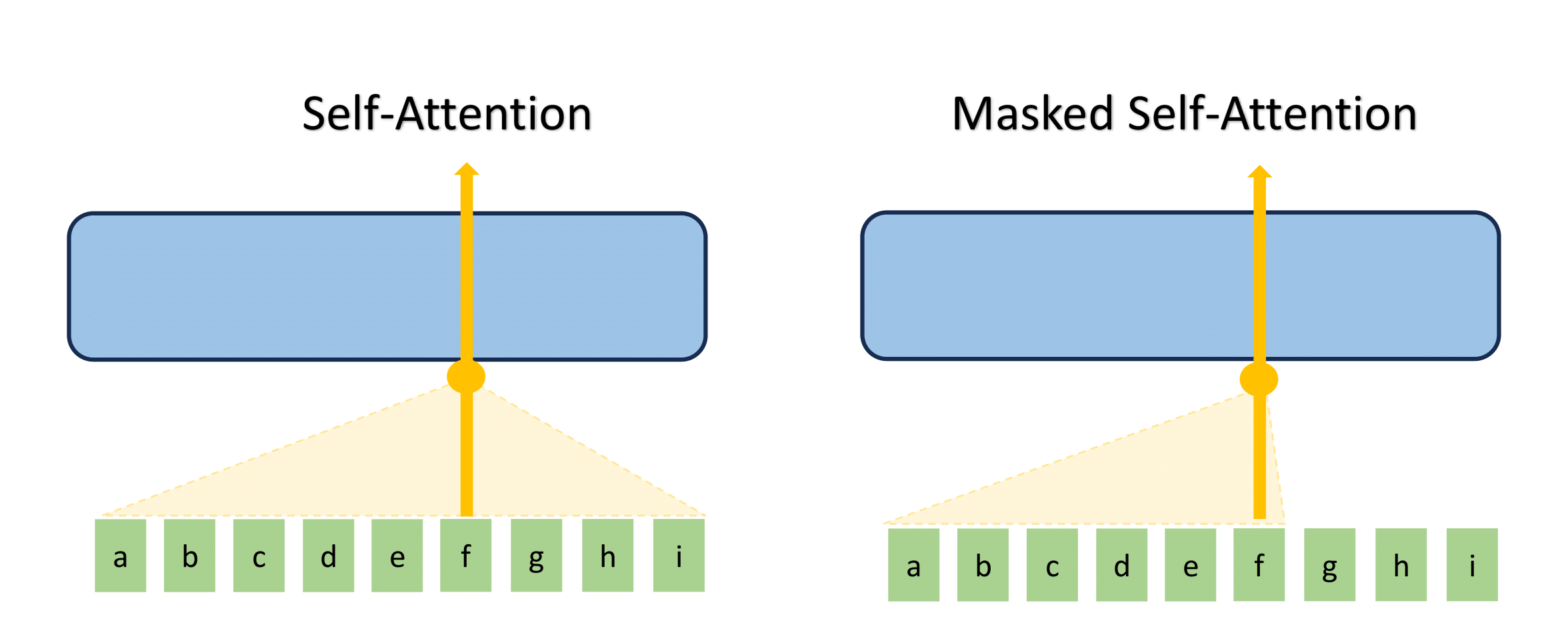}
        \caption{Self-Attention and Musk Self-Attention: a normal self-attention block allows a position to attend to all words to its right, while masked self-attention prevents a position from attending to words that come after it in the sequence.}
        \label{fig:5}
    \end{figure}
    
    \quad Masked Self-Attention: Masked self-attention is a modified version of self-attention designed to prevent the model from accessing future words during the prediction of the next word by incorporating a masking technique. For example, with the corpus ``abcde'', we can train GPT to predict the next word four times by feeding in ``a'', ``ab'', ``abc'' and ``abcd'' separately. However, this involves segmenting and inputting the sentence four times. The mask technique trick streamlines this process, enabling the model to receive the entire sentence at once and generate an output for each word. The challenge arises from the model peeking at subsequent words, which necessitates the use of masking. When predicting the next word of ``a'' GPT will only see ``a'', and when predicting the next word of ``ab'', it will only see ``ab''. It's essential to differentiate between self-attention (utilized in BERT~\cite{devlin2018bert}) and masked self-attention (utilized in GPT-3~\cite{brown2020language}). Figure~\ref{fig:5} provides a visual illustration. 
    
    \quad Multi-Head Self-Attention: Multi-head self-attention subdivides the self-attention mechanism into $h$ independent heads to compute in parallel (for GPT-3, $h$ is 96). Each head focuses on different aspects and connections within the sequence, concatenating their outcomes afterward. This approach facilitates the model to collectively concentrate on information derived from diverse representation sub-spaces, thereby enhancing its capacity for expression. Each head possesses its distinctive projection matrix, each with dimensions $d = d_{model}/h = 12288/96=128$. The outcome from each head is symbolized as $Y_{s}$, and when these individual head outputs are combined through concatenation ($96Y_{s} = 12288 = 96 \times 128$), the resultant amalgamation is denoted as $Y_{}$.
   
    \item Feedforward Neural Net: In addition to masked multi-head self-attention sub-layers, each of the transformer layers in the GPT model architecture contains a fully connected feedforward neural net. This neural network serves as a mechanism designed to non-linearly transform its inputs, thereby enhancing the model's overall expressive capabilities.
    
    \item Add \& Norm: The add \& norm modules are incorporated into each transformer layer on the GPT model architecture, where ``add'' represents the skip connection~\cite{kaiming2016deep} and ``norm'' represents layer normalization~\cite{ba2016layer}. The skip connection helps alleviate the issue of gradient vanishing in deep models, enabling the transformer layer to be stacked to a considerable depth. Additionally, layer normalization serves as a regularization strategy to prevent network overfitting. It calculates the mean and variance on each sample independently, without considering other data, thereby mitigating the impact of varying batch sizes.
    
    \item Decoding \& Output: After 12 transformer layers, the input matrix $X\in \mathbb{R}^{2048 \times 12288}$ transforms into an information-rich matrix $Y\in \mathbb{R}^{2048 \times 12288}$, representing the final representation of the input sequence. To achieve the desired result, we employ $Softmax(Y \cdot W)$, where $W\in \mathbb{R}^{12288 \times 50257}$, with 50,257 as the vocabulary size. This process generates probabilities assigned to each word in the vocabulary, enabling the selection of the most likely output as GPT's prediction for each word.
\end{itemize}

To help readers choose the most suitable model for their specific applications, we further organize LLMs based on their structure. Specifically, we classify LLMs into three types: \textit{encoder-only}, \textit{encoder-decoder}, and \textit{decoder-only} models. This taxonomy is based on the transformer architecture, the backbone of LLMs, comprising core components like the encoder, decoder, and their variations, which together enable their capabilities~\cite{vaswani2017attention}. Specifically, encoders are designed to process and interpret input sequences, while decoders generate output sequences based on this interpretation~\cite{wang2019language}. In some models, both components are integrated into an encoder-decoder framework to handle a diverse set of tasks effectively~\cite{wang2022language}. Conversely, other models focus on specific applications by using either the encoder alone for understanding tasks or the decoder alone for text generation. The detail information about each type of models is as outlined below:

\begin{itemize}
    \item Encoder-only models, such as BERT~\cite{devlin2018bert}, primarily focus on understanding and interpreting input data rather than generating it. They excel in tasks that require strong contextual understanding, such as sentiment analysis, named entity recognition, and text classification. These models are highly effective for analyzing text but are less suitable for text generation tasks due to their limited generation capabilities and computational intensity when handling large texts.
    
    \item Encoder-decoder LLMs, like T5~\cite{raffel2020exploring}, combine the strengths of both encoders and decoders, making them versatile and powerful for a range of applications. These models are particularly effective for tasks that require both text understanding and generation, such as machine translation, text summarization, and question answering. The encoder-decoder architecture provides a unified approach to these tasks, allowing for a comprehensive understanding and generation of text. However, the complexity of this dual architecture can result in more challenging training processes and slower inference times compared to other model types.
    
    \item Decoder-only LLMs, such as GPT-3~\cite{brown2020language}, are optimized for generating text, making them ideal for content creation, conversational AI, and creative writing. These models are known for their excellent text-generation capabilities and are effective in few-shot learning scenarios, where they can generate coherent and contextually relevant responses from minimal input. However, decoder-only models can face limitations in understanding context deeply and are often resource-intensive, requiring significant computational power to generate high-quality text outputs.
\end{itemize}

With the proposed taxonomy, table~\ref{table:llm_recent_years} provides a comparison of state-of-the-art LLMs across different types, with models in each category organized in ascending order of parameter size. This structure enables a systematic examination of key aspects such as architecture, parameter count, provider, release date, open-source status, tunability status, subsequently fine-tuning applied for the model (Adaptation), pre-training data scale, and evaluation methods in their original papers. Analyzing these elements together provides a thorough understanding of the models' capabilities and differences. Specifically, the type of each model indicates its architecture, whereas the parameter size reflects both the model's complexity and the resources required for fine-tuning. Next, the provider describes the model's development context and goals, and the release date places the model within the timeline of technological advancements. Furthermore, open-source status indicates the model's accessibility for future research and adaptation, whereas tunability status indicates whether the model has been customized after release. Last but not least, the pre-training data scale shows the model's exposure to various types of information, whereas the evaluation methods assess its performance against benchmarks. As an example, Codex~\cite{chen2021evaluating}, a decoder-only model with 12 billion parameters created by OpenAI and released in July 2021, was pre-trained on 100 billion tokens and supports in-context learning. While it is customizable and open-source, Codex does not incorporate more advanced techniques such as chain-of-thought methods or reinforcement learning with human feedback, which limits its ability to handle certain complex tasks.

\par\null 
\section*{4. APPLICATIONS OF LARGE LANGUAGE MODELS}

\begin{table*}[h]
\centering
\caption{Statistics of large language models in recent years. IT: instruction tuning. RLHF: reinforcement learning with human feedback. ICL: in-context learning. COT: chain-of-thought}
\label{table:llm_recent_years}
\resizebox{\textwidth}{!}{%
\begin{tabular}{|P{1.3cm}|P{3.2cm}|P{1cm}|P{2cm}|P{1.3cm}|P{1cm}|P{1cm}|P{0.55cm}P{0.7cm}|P{2cm}|P{0.53cm}P{0.53cm}|}
\hline
\multirow{2}{*}{Type} &
  \multirow{2}{*}{Model} &
  \multirow{2}{*}{Size} &
  \multirow{2}{*}{Provider} &
  \multirow{2}{*}{Release} &
  \multirow{2}{*}{\begin{tabular}[c]{@{}c@{}}Open-\\ sourced\end{tabular}} &
  \multirow{2}{*}{\begin{tabular}[c]{@{}c@{}}Tuna-\\ bility\end{tabular}} &
  \multicolumn{2}{c|}{Adaptation} &
  \multirow{2}{*}{\begin{tabular}[c]{@{}c@{}}Pre-train\\ Data scale\end{tabular}} &
  \multicolumn{2}{c|}{Evaluation} \\ 
 &
   &
   &
   &
   &
   &
   &
  \multicolumn{1}{c|}{IT} &
  RLHF &
   &
  \multicolumn{1}{c|}{ICL} &
  CoT \\ \hline
  
\multirow{6}{*}{\begin{tabular}[c]{@{}c@{}}Encoder-\\only\end{tabular}} 
& ALBERT~\cite{lan2019albert} & 0.012 & Google AI & 02/2020 & \checkmark & \checkmark  & - & - & 16GB & - & -\\ 
& DeBERTa~\cite{he2020deberta} & 0.1 & Microsoft & 06/2020 & \checkmark & \checkmark   & - & - & 80GB & - & -\\
& ELECTRA~\cite{clark2020electra} & 0.11 & Google AI & 03/2020 & \checkmark & \checkmark  & - & - & - & - & -\\ 
& BERT~\cite{devlin2018bert} & 0.11 & Google AI & 10/2018 & \checkmark & \checkmark & - & - & 3.3B words & - & - \\
& RoBERTa~\cite{zhuang2021robustly} & 0.125 & Meta AI & 07/2019 & \checkmark & \checkmark & - & - & 160GB & - & - \\ 
& XLM-RoBERTa~\cite{conneau2019unsupervised} & 0.27 & Meta AI & 07/2020 & \checkmark & \checkmark & - & - & 2.5TB & - & -\\ 
\hline

\multirow{21}{*}{\begin{tabular}[c]{@{}c@{}}Encoder-\\decoder\end{tabular}} 
& ERNIE 3.0~\cite{sun2021ernie} & 10 & Baidu & 07/2021 & x & \checkmark &- & -& 300B tokens & \checkmark & - \\ 
& T5~\cite{raffel2020exploring} & 11 & Google AI & 10/2019 & \checkmark & \checkmark & - & - & 1T tokens & \checkmark & - \\
& T0~\cite{sanh2022multitask} & 11 & BigScience & 10/2021 & \checkmark & \checkmark & \checkmark & - & - &\checkmark & - \\ 
& Flan-T5~\cite{chung2024scaling} & 11 & Google AI & 10/2022 & \checkmark & \checkmark & \checkmark & - & - & \checkmark & \checkmark \\ 
& UL2~\cite{tay2022ul2} & 20 & Google AI & 05/2022 &  \checkmark & \checkmark & - & - & 1T tokens & \checkmark & \checkmark \\ 
& AlexaTM~\cite{soltan2022alexatm} & 20 & Amazon & 08/2022 & x & \checkmark & - & - & 1.3T tokens & \checkmark & \checkmark\\ 
& AlphaCode~\cite{li2022competition} & 41 & Google AI & 02/2022 & x  & x & - & - & 967B tokens & - & - \\ 
& Anthropic~\cite{askell2021general} & 52 & Anthropic & 12/2021 & x & \checkmark & - & - & 400B tokens & \checkmark & - \\ 
& NLLB~\cite{costa2022no} & 54.5 & Meta AI & 07/2022 & \checkmark & \checkmark & - & - & - & \checkmark & - \\
& LLaMa~\cite{touvron2023llama} & 65 & Meta AI & 02/2023 & \checkmark & \checkmark & - & - & 1.4T tokens & \checkmark & -\\ 
& FLAN~\cite{wei2021finetuned} & 67 & Google AI & 09/2021 & x & \checkmark & \checkmark & - & - & \checkmark & - \\ 
& Sparrow~\cite{glaese2022improving} & 70 & Google AI & 09/2022 & x & \checkmark & - & \checkmark & - & \checkmark & \checkmark \\ 
& Chinchilla~\cite{hoffmann2022training} & 70 &  Google AI & 03/2022 & x & \checkmark & - & - & 1.4T tokens & \checkmark & - \\
& OPT-OML~\cite{iyer2022opt} & 175 & Meta AI & 12/2022 & \checkmark & x & \checkmark & - & - & \checkmark & \checkmark \\ 
& Gopher~\cite{rae2021scaling} & 280 & Google AI & 12/2021 & x & \checkmark & - & - & 300B tokens & \checkmark & - \\ 
& PaLM~\cite{chowdhery2023palm} & 540 & Google AI & 04/2022 & \checkmark & \checkmark & - & - & 780B tokens & \checkmark & \checkmark \\ 
& U-PaLM~\cite{tay2023transcending} & 540 & Google AI & 10/2022 & x & x  &- & - & - & \checkmark & \checkmark\\
& GLaM~\cite{du2022glam} & 1200 & Google AI & 12/2021 & x  & \checkmark & - & - & 280B tokens & \checkmark & - \\ 
\hline
\multirow{27}{*}{\begin{tabular}[c]{@{}c@{}}Decoder-\\ only\end{tabular}}
& Codex~\cite{chen2021evaluating} & 12 & OpenAI & 07/2021 & x & x & - & - & 100B tokens & \checkmark & - \\ 
& Pythia~\cite{biderman2023pythia} & 12 & Eleuther AI & 04/2023 & \checkmark & x & - & - & 300B tokens & \checkmark & - \\ 
& CodeGeeX~\cite{zheng2023codegeex} & 13 & Tsinghua U. & 09/2022 & \checkmark & \checkmark & - & - & 850B tokens & \checkmark & - \\ 
& Skywork~\cite{wei2023skywork} & 13 & Kunlun Inc & 10/2023 & \checkmark & \checkmark & - & - & 3.2T tokens & \checkmark & - \\ 
& QWEN~\cite{bai2023qwen} & 14 & Alibaba & 09/2023 & \checkmark & \checkmark & \checkmark & \checkmark & 3T tokens & \checkmark & - \\ 
& CodeGen2~\cite{nijkamp2023codegen2} & 16 & Salesforce AI & 05/2023 & \checkmark & \checkmark & - & - & 400B tokens & \checkmark & - \\
& GPT-NeoX-20B~\cite{black2022gpt} & 20 & EleutherAI & 04/2022 & \checkmark & x & - & - & 825GB & \checkmark & - \\
& Gemini 1.5~\cite{reid2024gemini} & 20 & Google AI & 02/2024 & x & \checkmark & - & - & - & \checkmark & - \\ 
& LlaMa 2~\cite{touvron2023llama} & 70 & Meta AI & 07/2023 & \checkmark & \checkmark & \checkmark & \checkmark & 2T tokens & \checkmark & - \\ 
& LlaMa 3~\cite{dubey2024llama} & 70 & Meta AI & 04/2024 & \checkmark & \checkmark &\checkmark & \checkmark& 15T tokens & \checkmark & - \\ 
& HyperCLOVA~\cite{kim2021changes} & 82 & Naiver & 09/2021 & x & \checkmark & - & - & 300B tokens & \checkmark & - \\
& Galactia~\cite{taylor2022galactica} & 120 & Meta AI & 11/2022 & \checkmark & \checkmark & - & - & 106B tokens & \checkmark & \checkmark \\ 
& GLM~\cite{zeng2022glm} & 130 & Tsinghua U. & 10/2022 & \checkmark & \checkmark & - & - & 400B tokens & \checkmark & - \\ 
& LaMDA~\cite{thoppilan2022lamda} & 137 & Google AI & 01/2022 & x & \checkmark & - & - & 768B tokens & \checkmark & - \\ 
& Jurassic-1~\cite{lieber2021jurassic} & 178 & AI21 Labs & 08/2021 & x & \checkmark & - & - & 300B tokens & \checkmark & - \\   
& GPT-3~\cite{brown2020language} & 175 & OpenAI & 05/2020 & x & x & - & - & 300B tokens & \checkmark & - \\ 
& WebGPT~\cite{nakano2021webgpt} & 175 & OpenAI & 12/2021 & x & x & - & - & - & \checkmark & - \\ \ 
& InstructGPT~\cite{ouyang2022training} & 175 & OpenAI & 03/2022 & x & x & \checkmark & \checkmark & - & \checkmark & - \\  
& BLOOM~\cite{le2023bloom} & 176 &  BigScience & 11/2022 & \checkmark & \checkmark & - & - & 366B tokens & \checkmark & - \\ 
& BLOOMZ~\cite{muennighoff2023crosslingual} & 176 & BigScience & 11/2022 & \checkmark & x  & \checkmark & - & - & \checkmark & - \\ 
& Llama 3.1~\cite{vavekanandllama} & 405 & Meta AI & 06/2024 & \checkmark & \checkmark &\checkmark & \checkmark & 15T tokens & \checkmark & - \\
& GPT-4~\cite{achiam2023gpt} & 450 & OpenAI & 03/2023 & x & \checkmark & \checkmark & \checkmark & 1.76T & \checkmark & \checkmark \\
& MT-NLG~\cite{smith2022using} & 530 & Microsoft & 01/2022 & x & \checkmark & - & - & 967B tokens & \checkmark & - \\ 
& Gemma~\cite{team2024gemma} & 2700 & Google AI & 02/2024 & \checkmark & \checkmark & - & - & 6T tokens & \checkmark & -  \\ 
 \hline
\end{tabular}%
}
\end{table*}

{\label{350277}}
LLMs have been instrumental across diverse domains, and this section aims to illustrate exemplary applications of LLMs in various fields.

\textbf{Software Engineering}: LLMs have demonstrated remarkable capabilities across various software engineering tasks such as defect prediction~\cite{steenhoek2023empirical}, code review~\cite{li2022automating}, code generation~\cite{ahmad2021unified}, and analysis of software-related questions~\cite{he2024representation, he2022ptm4tag, yang2022answer}. These models significantly aid in improving code quality, debugging, and reducing human involvement in many aspects of software development. For instance, Codex~\cite{chen2021evaluating} is a pioneering model in this domain, utilizing a large generative pre-trained model with up to 12 billion parameters to aid developers by generating code, suggesting optimizations, and identifying and correcting errors. Building on this foundation, several other models have been developed to address various facets of coding and software development. Notable examples include DeepMind's AlphaCode~\cite{li2022competition}, which specializes in generating code for programming competitions, and Meta's InCoder~\cite{fried2022incoder} and Code Llama~\cite{roziere2023code}, as well as Salesforce's CodeRL~\cite{le2022coderl} and CodeGen~\cite{nijkamp2022codegen, nijkamp2023codegen2}. The BigCode project has introduced StarCoder~\cite{le2022coderl}, and OpenAI has made significant strides with its GPT~\cite{achiam2023gpt} and ChatGPT series, enhanced through Reinforcement Learning from Human Feedback (RLHF). These developments expand the scope and potential of automated code generation, setting a new standard in software engineering that offers both opportunities and challenges for the industry.

\textbf{Drug Discovery}: The field of drug discovery is characterized by substantial challenges and elevated costs. The utilization of computational methodologies, such as AI, DL, and quantum mechanical techniques, holds the promise of accelerating the identification of potential drug candidates and predicting their properties~\cite{sadybekov2023computational,gorgulla2022emerging}. Among recent advancements in this arena, the integration of LLMs, such as ChatGPT, emerges as a valuable addition, adept at generating and analyzing natural language texts~\cite{savage2023drug}. LLMs contribute significantly to the initial phase of drug discovery, particularly in identifying suitable drug targets for diverse diseases~\cite{haley2020functional,paananen2020omics}. They offer preliminary insights into the structure and functionality of protein-based drug targets, subject to further validation by alternative methodologies. Furthermore, LLMs play a pivotal role in subsequent drug discovery steps, aiding in the design and screening of small molecules capable of interacting with these drug targets~\cite{savage2023drug}.

\textbf{Finance}: LLMs are driving significant advancements in the finance industry, with significant financial applications including trading and portfolio management~\cite{zhang2020deep}, financial risk modeling~\cite{mashrur2020machine}, financial text mining~\cite{gupta2020comprehensive,pagliaro2021investor}, and financial advisory and customer services~\cite{shahfinaid}. AI integration in financial advisory and customer services is an emerging and rapidly expanding field. According to~\cite{misischia2022chatbots}, AI-driven chatbots already provide more than 37\% of supporting functions in various e-service scenarios. Within the financial field, these chatbots are embraced as cost-effective alternatives to human customer service, a trend underscored in the ``Chatbots in Consumer Finance'' report\footnote{https://www.consumerfinance.gov/data-research/research-reports/chatbots-in-consumer-finance/chatbots-in-consumer-finance}. With the advent of LLMs, more and more tasks that were previously considered difficult to handle have become possible, further expanding the potential applications of AI in the financial industry~\cite{li2023large}. Notably, there are already some specialized LLMs like BloombergGPT~\cite{wu2023bloomberggpt}, a 50-billion-parameter model excelling in financial tasks while maintaining strong performance in general LM benchmarks.

\textbf{Medical}: LLMs are now at the forefront of medical AI with immense potential to improve the efficiency and effectiveness of clinical, educational, and research work~\cite{thirunavukarasu2023large,chinta2024ai}. One example is the performance of LLMs on the MedQA dataset~\cite{jin2021disease}, which serves as a widely used biomedical question-answering dataset consisting of the United States Medical Licensing Examination (USMLE)-style questions. In this dataset, ChatGPT attains human passing levels and~\cite{kung2023performance}. Additionally, in a span of fewer than six months, Med-PaLM2 achieves proficiency levels close to human experts~\cite{singhal2023towards}. The strong performance of GPT-4~\cite{achiam2023gpt} and Med-PaLM2 in medical tests suggests that LLMs may be useful teaching tools for students currently attaining a lower level in such tests~\cite{thirunavukarasu2023large}. GPT-4’s meta-prompt feature allows users to explicitly describe the desired role for the chatbot to take on during conversation; a useful example is a ``Socratic tutor mode''\footnote{https://sites.highlands.edu/tutorial-center/tutor-resources/online-tutor-training/module-4/the-socratic-method/}, which encourages students to think for themselves by pitching questions at decreasing levels of difficulty until students can work out solutions to the fuller question at hand. This mode can foster critical thinking and problem-solving skills among students~\cite{thirunavukarasu2023large}. ChatGPT demonstrates superior performance in tasks not necessitating specialized knowledge or when such expertise is available in user prompts~\cite{arora2023promise}. For example, consider discharge summaries as a case in point - they entail repetitive tasks involving information interpretation and compression, often requiring minimal specialized knowledge and user prompts~\cite{touvron2023llama}. 

\textbf{Legal}: LLMs exhibit robust capabilities across various legal tasks, particularly in case prediction and generating legal texts. For instance, ChatGPT achieved a 50.3\% accuracy rate on NCBE MBE practice exams, significantly surpassing the 25\% random guess level, and achieved passing scores in both Evidence and Torts~\cite{bommarito2022gpt}. The integration of ChatGPT holds the potential to reshape the entire legal industry. An Australian law firm experimented with ChatGPT to draft a statement of claim based on the 1992 Mabo case, yielding results akin to those of a first-year lawyer\footnote{\href{https://www.abc.net.au/news/science/2023-01-25/chatgpt-midjourney-generative-ai-and-future-of-work/101882580}{How ChatGPT and other new AI tools are being used by lawyers, architects and coders - ABC News} }. \cite{iu2023chatgpt} investigation explored the possibility that ChatGPT could replace litigators by evaluating its drafting and research capabilities. While recommending ChatGPT as a complement to litigators rather than a complete replacement, this suggestion stemmed from ChatGPT's limitations in meeting the precise accuracy and timeliness demands of legal texts, especially in ensuring continuous access to updated legal research. More recently, advancements such as DeLilegal\footnote{https://data.delilegal.com/lawQuestion} addressed this issue by integrating with external law databases to enhance retrieval and minimize hallucination. Meanwhile, LLMs in the legal field have emerged in recent months such as ChatLaw~\cite{cui2023chatlaw}.

\textbf{Education}: With the rapid advancement of AI technology, its applications in education are expanding, fundamentally transforming teaching and learning methods. Researchers have recognized the remarkable capabilities of LLMs, such as ChatGPT, and have explored their substantial potential to impact various educational settings. These models can be used in diverse ways, such as simulating students to help train teachers~\cite{lee23generative, markel2023gpteach, chinta2024fairaied} and acting as virtual instructors to provide personalized instruction to students~\cite{tu2023littlemu, sonkar2023class, chen2023empowering}. LLMs can also function as personal tutors, offering real-time feedback~\cite{zentner2022applied} and tailored evaluations and suggestions based on individual learning progress~\cite{baidoo2023education, zhang2023preparing}. Furthermore, they can enhance the overall learning experience by improving student engagement and promoting greater autonomy in learning~\cite{dwivedi2023opinion, xiao2023evaluating}. In addition to supporting individualized learning, LLMs can address broader educational challenges, such as the low teacher-student ratio, by supplementing human instructors and providing additional support to students~\cite{chen2023artificial}. The integration of LLMs into education presents a unique opportunity to personalize and scale educational resources, making quality education more accessible to a broader range of learners. 

\section*{5. DRAWBACKS AND FUTURE DIRECTIONS OF LARGE LANGUAGE MODELS}
{\label{880788}}
While LLMs offer significant advantages in improving both professional and personal aspects of life, it is essential to acknowledge their accompanying drawbacks. In the subsequent discussion, we will identify and explore some of these limitations inherent in current LLMs, which concurrently provide insights into potential directions for future advancements.


\textbf{Privacy}: LLMs are widely used across various sectors, but they also present significant privacy concerns~\cite{yan2024protecting, yao2024survey}. One major issue is the risk of data leakage, where LLMs may inadvertently memorize and reproduce sensitive or personal information from their training datasets, potentially leading to privacy breaches. This risk is particularly severe in domains like healthcare and finance, where exposing private data can have serious consequences~\cite{carlini2021extracting}. Additionally, the training datasets used for LLMs are often not fully anonymized, meaning personal identifiers and other sensitive information could be embedded within the model~\cite{fredrikson2015model}. This increases the risk of violating privacy regulations, such as the General Data Protection Regulation~\cite{leboukh2023balancing}. Recent developments have further complicated the privacy landscape, as tools like FraudGPT and WormGPT have emerged. These tools, designed for cybercrime, highlight the misuse potential of LLMs by generating fraudulent emails, suggesting malicious link placements, and facilitating cyberattacks. FraudGPT~\cite{falade2023decoding, amos2023fraudgpt} and WormGPT~\cite{delley2023wormgpt} were trained on datasets focused on malware and fraud, enabling cybercriminals to conduct sophisticated attacks such as Business Email Compromise (BEC). This illustrates the urgent need for robust privacy safeguards and regulatory frameworks to prevent the misuse of LLMs and protect sensitive information from being exploited maliciously.

\textbf{Fairness}: LLMs are widely employed for decision-making across various domains, impacting individuals and society. However, these decisions might exhibit bias against marginalized populations~\cite{saxena2023missed,wang2024individual,wang2024toward,wang2023fg2an,wang2023preventing}. For instance, when utilized in screening resumes for programming positions, LLMs might exhibit a bias favoring male candidates, demonstrating evident gender discrimination~\cite{chu2024fairness,doan2024fairness1,zhang2023fairness,doan2024fairness,zhang2024inpractice}. This bias stems from the training of LLMs on extensive and unstructured Internet data, where they inherit stereotypes, misrepresentations, derogatory language, and exclusionary behaviors. These aspects disproportionately affect already disadvantaged and marginalized groups~\cite{bender2021dangers,dodge2021documenting,sheng2021societal}. The commonly employed strategy to mitigate bias in LLMs is to remove biased data from the training dataset. However, this strategy does not entirely eliminate bias and often diminishes the effectiveness of language modeling~\cite{meade2021empirical}. Saxena et al.~\cite{saxena2023missed} discussed that the root causes of these problems lie not only in technology but also in socially acceptable definitions of fairness and meaningful interventions to ensure the long-term well-being of all groups. Researchers should consider the needs of disadvantaged groups from the outset and design these technologies proactively, rather than simply reacting to the biases present in their designed systems~\cite{wang2023mitigating,gallegos2024bias,wang2024advancing}. The documented bias in LLMs, as highlighted by studies like~\cite{blodgett2017racial,mei2023bias,mozafari2020hate}, underscores the imperative for future efforts in this direction.

\textbf{Safety}: LLMs have diverse applications in industries such as finance and healthcare. However, these applications also present significant security challenges. LLMs may generate outputs that diverge from the provided context, user input, or factual knowledge, a phenomenon referred to as ``hallucination''~\cite{bang2023multitask,huang2023survey}. This undermines their reliability in real-world scenarios, emphasizing the vital necessity for precision, particularly in critical domains like medicine~\cite{dash2023evaluation,umapathi2023med}. As a primary way to avoid these problems, well-aligned LLMs can be developed by including humans in the training loop and using reinforcement learning from human feedback (RLHF)~\cite{ouyang2022training,gao2020pile}. To improve model safety, it is also important to include safety-related cues in the RLHF process, as shown in GPT-4~\cite{achiam2023gpt}. However, RLHF relies heavily on high-quality human feedback data from professional annotators, which is costly and time-consuming. As a result, enhancing the RLHF framework is essential to alleviate the burden on human annotators. Additionally, exploring more efficient annotation methods with assured data quality, such as LLMs, can assist in the annotation process and enhance overall safety.

\textbf{Intellectual Property}: LLMs possess the capability to create content resembling human-generated material, potentially encroaching upon users' intellectual property rights. These models rely on training and optimizing using users' textual data, which may encompass personal information, proprietary knowledge, patented technology, and more. Notably, AI-generated code exemplifies this issue. Code-generation models trained on open-source repositories can produce programs resembling existing ones, potentially disregarding relevant licenses~\cite{yu2023codeipprompt}. For instance, legal action has been taken against Microsoft, GitHub, and OpenAI, alleging copyright infringement due to Copilot reproducing licensed code without adherence to licensing terms~\footnote{https://www.reuters.com/technology/microsoft-attracting-users-its-code-writing-generative-ai-software-2023-01-25/}~\footnote{https://www.reuters.com/legal/litigation/openai-microsoft-want-court-toss-lawsuit-accusing-them-abusing-open-source-code-2023-01-27/}. Additionally, there are concerns about potential copyright infringements with LLM-generated images and texts~\cite{dzuong2024uncertain,yazdani2024comprehensive}, as highlighted in recent lawsuits by a group of novelists against OpenAI in July 2023 for allegedly using their books without permission to train models~\cite{novelistLawsuit}, and by The New York Times against OpenAI and Microsoft in December 2023 for scraping articles to train their generative models without consent~\cite{Stempel_2023}. Consequently, there's a pressing need for heightened attention and regulation concerning intellectual property rights concerning LLMs. This is essential to safeguard the legitimate rights and interests of original creators, users, and the public while fostering and respecting the innovation and advancement of LLMs technology~\cite{li2023protecting}.

\section*{6. CONCLUSIONS}
{\label{880798}}

In recent times, Large Language Models (LLMs) have garnered significant attention from the fields of science, industry, society, and beyond. However, despite the substantial advantages that LLMs offer in enhancing both professional and personal aspects of life, a limited understanding among general practitioners regarding the background and principles of these models impedes their full potential. In light of this, the survey at hand systematically delves into the evolution of LLMs, presenting key concepts and techniques for a comprehensive understanding of these models. It introduces a detailed taxonomy of LLMs and compares state-of-the-art models to assist readers in selecting the most suitable model for their specific needs. Additionally, the survey highlights the limitations of current LLMs and identifies promising areas for future research. Notably, setting itself apart from other surveys on LLMs, this study prioritizes empowering practitioners, irrespective of their background knowledge, to ensure their proficient utilization in both scientific research and daily tasks.

\section*{Acknowledgement}

This work was supported in part by the National Science Foundation (NSF) under Grant No. 2245895. 



\bibliography{main}


\begin{thebibliography}{174}
\ifx \bisbn   \undefined \def \bisbn  #1{ISBN #1}\fi
\ifx \binits  \undefined \def \binits#1{#1}\fi
\ifx \bauthor  \undefined \def \bauthor#1{#1}\fi
\ifx \batitle  \undefined \def \batitle#1{#1}\fi
\ifx \bjtitle  \undefined \def \bjtitle#1{#1}\fi
\ifx \bvolume  \undefined \def \bvolume#1{\textbf{#1}}\fi
\ifx \byear  \undefined \def \byear#1{#1}\fi
\ifx \bissue  \undefined \def \bissue#1{#1}\fi
\ifx \bfpage  \undefined \def \bfpage#1{#1}\fi
\ifx \blpage  \undefined \def \blpage #1{#1}\fi
\ifx \burl  \undefined \def \burl#1{\textsf{#1}}\fi
\ifx \doiurl  \undefined \def \doiurl#1{\url{https://doi.org/#1}}\fi
\ifx \betal  \undefined \def \betal{\textit{et al.}}\fi
\ifx \binstitute  \undefined \def \binstitute#1{#1}\fi
\ifx \binstitutionaled  \undefined \def \binstitutionaled#1{#1}\fi
\ifx \bctitle  \undefined \def \bctitle#1{#1}\fi
\ifx \beditor  \undefined \def \beditor#1{#1}\fi
\ifx \bpublisher  \undefined \def \bpublisher#1{#1}\fi
\ifx \bbtitle  \undefined \def \bbtitle#1{#1}\fi
\ifx \bedition  \undefined \def \bedition#1{#1}\fi
\ifx \bseriesno  \undefined \def \bseriesno#1{#1}\fi
\ifx \blocation  \undefined \def \blocation#1{#1}\fi
\ifx \bsertitle  \undefined \def \bsertitle#1{#1}\fi
\ifx \bsnm \undefined \def \bsnm#1{#1}\fi
\ifx \bsuffix \undefined \def \bsuffix#1{#1}\fi
\ifx \bparticle \undefined \def \bparticle#1{#1}\fi
\ifx \barticle \undefined \def \barticle#1{#1}\fi
\bibcommenthead
\ifx \bconfdate \undefined \def \bconfdate #1{#1}\fi
\ifx \botherref \undefined \def \botherref #1{#1}\fi
\ifx \url \undefined \def \url#1{\textsf{#1}}\fi
\ifx \bchapter \undefined \def \bchapter#1{#1}\fi
\ifx \bbook \undefined \def \bbook#1{#1}\fi
\ifx \bcomment \undefined \def \bcomment#1{#1}\fi
\ifx \oauthor \undefined \def \oauthor#1{#1}\fi
\ifx \citeauthoryear \undefined \def \citeauthoryear#1{#1}\fi
\ifx \endbibitem  \undefined \def \endbibitem {}\fi
\ifx \bconflocation  \undefined \def \bconflocation#1{#1}\fi
\ifx \arxivurl  \undefined \def \arxivurl#1{\textsf{#1}}\fi
\csname PreBibitemsHook\endcsname

\bibitem[\protect\citeauthoryear{Pinker}{1994}]{pinker1994language}
\begin{botherref}
\oauthor{\bsnm{Pinker}, \binits{S.}}:
The language instinct: How the mind creates.
Language. New York: Harper Collins
(1994)
\end{botherref}
\endbibitem

\bibitem[\protect\citeauthoryear{Turing et~al.}{2000}]{turing2000computing}
\begin{barticle}
\bauthor{\bsnm{Turing}, \binits{A.M.}},
\bauthor{\bsnm{Geirsson}, \binits{H.}},
\bauthor{\bsnm{Losonsky}, \binits{M.}}:
\batitle{Computing machinery and intelligence}.
\bjtitle{Artificial Intelligence: Critical Concepts}
\bvolume{2}(\bissue{236}),
\bfpage{19}
(\byear{2000})
\end{barticle}
\endbibitem

\bibitem[\protect\citeauthoryear{Dwivedi et~al.}{2023}]{dwivedi2023so}
\begin{barticle}
\bauthor{\bsnm{Dwivedi}, \binits{Y.K.}},
\bauthor{\bsnm{Kshetri}, \binits{N.}},
\bauthor{\bsnm{Hughes}, \binits{L.}},
\bauthor{\bsnm{Slade}, \binits{E.L.}},
\bauthor{\bsnm{Jeyaraj}, \binits{A.}},
\bauthor{\bsnm{Kar}, \binits{A.K.}},
\bauthor{\bsnm{Baabdullah}, \binits{A.M.}},
\bauthor{\bsnm{Koohang}, \binits{A.}},
\bauthor{\bsnm{Raghavan}, \binits{V.}},
\bauthor{\bsnm{Ahuja}, \binits{M.}}, \betal:
\batitle{“so what if chatgpt wrote it?” multidisciplinary perspectives on opportunities, challenges and implications of generative conversational ai for research, practice and policy}.
\bjtitle{International Journal of Information Management}
\bvolume{71},
\bfpage{102642}
(\byear{2023})
\end{barticle}
\endbibitem

\bibitem[\protect\citeauthoryear{Zhao et~al.}{2023}]{zhao2023survey}
\begin{botherref}
\oauthor{\bsnm{Zhao}, \binits{W.X.}},
\oauthor{\bsnm{Zhou}, \binits{K.}},
\oauthor{\bsnm{Li}, \binits{J.}},
\oauthor{\bsnm{Tang}, \binits{T.}},
\oauthor{\bsnm{Wang}, \binits{X.}},
\oauthor{\bsnm{Hou}, \binits{Y.}},
\oauthor{\bsnm{Min}, \binits{Y.}},
\oauthor{\bsnm{Zhang}, \binits{B.}},
\oauthor{\bsnm{Zhang}, \binits{J.}},
\oauthor{\bsnm{Dong}, \binits{Z.}}, et al.:
A survey of large language models.
arXiv preprint arXiv:2303.18223
(2023)
\end{botherref}
\endbibitem

\bibitem[\protect\citeauthoryear{Jin et~al.}{2023}]{jin2023rethinking}
\begin{botherref}
\oauthor{\bsnm{Jin}, \binits{H.}},
\oauthor{\bsnm{Wei}, \binits{W.}},
\oauthor{\bsnm{Wang}, \binits{X.}},
\oauthor{\bsnm{Zhang}, \binits{W.}},
\oauthor{\bsnm{Wu}, \binits{Y.}}:
Rethinking learning rate tuning in the era of large language models,
112--121
(2023).
IEEE
\end{botherref}
\endbibitem

\bibitem[\protect\citeauthoryear{Fu et~al.}{2022}]{fu2022does}
\begin{botherref}
\oauthor{\bsnm{Fu}, \binits{Y.}},
\oauthor{\bsnm{Peng}, \binits{H.}},
\oauthor{\bsnm{Khot}, \binits{T.}}:
How does gpt obtain its ability? tracing emergent abilities of language models to their sources.
Yao Fu’s Notion
(2022)
\end{botherref}
\endbibitem

\bibitem[\protect\citeauthoryear{Wei et~al.}{2022}]{wei2022emergent}
\begin{botherref}
\oauthor{\bsnm{Wei}, \binits{J.}},
\oauthor{\bsnm{Tay}, \binits{Y.}},
\oauthor{\bsnm{Bommasani}, \binits{R.}},
\oauthor{\bsnm{Raffel}, \binits{C.}},
\oauthor{\bsnm{Zoph}, \binits{B.}},
\oauthor{\bsnm{Borgeaud}, \binits{S.}},
\oauthor{\bsnm{Yogatama}, \binits{D.}},
\oauthor{\bsnm{Bosma}, \binits{M.}},
\oauthor{\bsnm{Zhou}, \binits{D.}},
\oauthor{\bsnm{Metzler}, \binits{D.}}, et al.:
Emergent abilities of large language models.
Transactions on Machine Learning Research
(2022)
\end{botherref}
\endbibitem

\bibitem[\protect\citeauthoryear{Radford et~al.}{2018}]{radford2018improving}
\begin{botherref}
\oauthor{\bsnm{Radford}, \binits{A.}},
\oauthor{\bsnm{Narasimhan}, \binits{K.}},
\oauthor{\bsnm{Salimans}, \binits{T.}},
\oauthor{\bsnm{Sutskever}, \binits{I.}}, et al.:
Improving language understanding by generative pre-training
(2018)
\end{botherref}
\endbibitem

\bibitem[\protect\citeauthoryear{Radford et~al.}{2019}]{radford2019language}
\begin{barticle}
\bauthor{\bsnm{Radford}, \binits{A.}},
\bauthor{\bsnm{Wu}, \binits{J.}},
\bauthor{\bsnm{Child}, \binits{R.}},
\bauthor{\bsnm{Luan}, \binits{D.}},
\bauthor{\bsnm{Amodei}, \binits{D.}},
\bauthor{\bsnm{Sutskever}, \binits{I.}}, \betal:
\batitle{Language models are unsupervised multitask learners}.
\bjtitle{OpenAI blog}
\bvolume{1}(\bissue{8}),
\bfpage{9}
(\byear{2019})
\end{barticle}
\endbibitem

\bibitem[\protect\citeauthoryear{Brown et~al.}{2020}]{brown2020language}
\begin{barticle}
\bauthor{\bsnm{Brown}, \binits{T.}},
\bauthor{\bsnm{Mann}, \binits{B.}},
\bauthor{\bsnm{Ryder}, \binits{N.}},
\bauthor{\bsnm{Subbiah}, \binits{M.}},
\bauthor{\bsnm{Kaplan}, \binits{J.D.}},
\bauthor{\bsnm{Dhariwal}, \binits{P.}},
\bauthor{\bsnm{Neelakantan}, \binits{A.}},
\bauthor{\bsnm{Shyam}, \binits{P.}},
\bauthor{\bsnm{Sastry}, \binits{G.}},
\bauthor{\bsnm{Askell}, \binits{A.}}, \betal:
\batitle{Language models are few-shot learners}.
\bjtitle{Advances in neural information processing systems}
\bvolume{33},
\bfpage{1877}--\blpage{1901}
(\byear{2020})
\end{barticle}
\endbibitem

\bibitem[\protect\citeauthoryear{Achiam et~al.}{2023}]{achiam2023gpt}
\begin{botherref}
\oauthor{\bsnm{Achiam}, \binits{J.}},
\oauthor{\bsnm{Adler}, \binits{S.}},
\oauthor{\bsnm{Agarwal}, \binits{S.}},
\oauthor{\bsnm{Ahmad}, \binits{L.}},
\oauthor{\bsnm{Akkaya}, \binits{I.}},
\oauthor{\bsnm{Aleman}, \binits{F.L.}},
\oauthor{\bsnm{Almeida}, \binits{D.}},
\oauthor{\bsnm{Altenschmidt}, \binits{J.}},
\oauthor{\bsnm{Altman}, \binits{S.}},
\oauthor{\bsnm{Anadkat}, \binits{S.}}, et al.:
Gpt-4 technical report.
arXiv preprint arXiv:2303.08774
(2023)
\end{botherref}
\endbibitem

\bibitem[\protect\citeauthoryear{Liu et~al.}{2023}]{liu2023pre}
\begin{barticle}
\bauthor{\bsnm{Liu}, \binits{P.}},
\bauthor{\bsnm{Yuan}, \binits{W.}},
\bauthor{\bsnm{Fu}, \binits{J.}},
\bauthor{\bsnm{Jiang}, \binits{Z.}},
\bauthor{\bsnm{Hayashi}, \binits{H.}},
\bauthor{\bsnm{Neubig}, \binits{G.}}:
\batitle{Pre-train, prompt, and predict: A systematic survey of prompting methods in natural language processing}.
\bjtitle{ACM Computing Surveys}
\bvolume{55}(\bissue{9}),
\bfpage{1}--\blpage{35}
(\byear{2023})
\end{barticle}
\endbibitem

\bibitem[\protect\citeauthoryear{Han et~al.}{2021}]{han2021pre}
\begin{barticle}
\bauthor{\bsnm{Han}, \binits{X.}},
\bauthor{\bsnm{Zhang}, \binits{Z.}},
\bauthor{\bsnm{Ding}, \binits{N.}},
\bauthor{\bsnm{Gu}, \binits{Y.}},
\bauthor{\bsnm{Liu}, \binits{X.}},
\bauthor{\bsnm{Huo}, \binits{Y.}},
\bauthor{\bsnm{Qiu}, \binits{J.}},
\bauthor{\bsnm{Yao}, \binits{Y.}},
\bauthor{\bsnm{Zhang}, \binits{A.}},
\bauthor{\bsnm{Zhang}, \binits{L.}}, \betal:
\batitle{Pre-trained models: Past, present and future}.
\bjtitle{AI Open}
\bvolume{2},
\bfpage{225}--\blpage{250}
(\byear{2021})
\end{barticle}
\endbibitem

\bibitem[\protect\citeauthoryear{Shanahan}{2024}]{shanahan2024talking}
\begin{barticle}
\bauthor{\bsnm{Shanahan}, \binits{M.}}:
\batitle{Talking about large language models}.
\bjtitle{Communications of the ACM}
\bvolume{67}(\bissue{2}),
\bfpage{68}--\blpage{79}
(\byear{2024})
\end{barticle}
\endbibitem

\bibitem[\protect\citeauthoryear{Dodge et~al.}{2021}]{dodge2021documenting}
\begin{bchapter}
\bauthor{\bsnm{Dodge}, \binits{J.}},
\bauthor{\bsnm{Sap}, \binits{M.}},
\bauthor{\bsnm{Marasovi{\'c}}, \binits{A.}},
\bauthor{\bsnm{Agnew}, \binits{W.}},
\bauthor{\bsnm{Ilharco}, \binits{G.}},
\bauthor{\bsnm{Groeneveld}, \binits{D.}},
\bauthor{\bsnm{Mitchell}, \binits{M.}},
\bauthor{\bsnm{Gardner}, \binits{M.}}:
\bctitle{Documenting large webtext corpora: A case study on the colossal clean crawled corpus}.
In: \bbtitle{Proceedings of the 2021 Conference on Empirical Methods in Natural Language Processing (EMNLP)},
pp. \bfpage{1286}--\blpage{1305}
(\byear{2021})
\end{bchapter}
\endbibitem

\bibitem[\protect\citeauthoryear{Sonkar et~al.}{2023}]{sonkar2023class}
\begin{bchapter}
\bauthor{\bsnm{Sonkar}, \binits{S.}},
\bauthor{\bsnm{Liu}, \binits{N.}},
\bauthor{\bsnm{Mallick}, \binits{D.}},
\bauthor{\bsnm{Baraniuk}, \binits{R.}}:
\bctitle{Class: A design framework for building intelligent tutoring systems based on learning science principles}.
In: \bbtitle{Findings of the Association for Computational Linguistics: EMNLP 2023},
pp. \bfpage{1941}--\blpage{1961}
(\byear{2023})
\end{bchapter}
\endbibitem

\bibitem[\protect\citeauthoryear{Kim et~al.}{2021}]{kim2021changes}
\begin{bchapter}
\bauthor{\bsnm{Kim}, \binits{B.}},
\bauthor{\bsnm{Kim}, \binits{H.}},
\bauthor{\bsnm{Lee}, \binits{S.-W.}},
\bauthor{\bsnm{Lee}, \binits{G.}},
\bauthor{\bsnm{Kwak}, \binits{D.}},
\bauthor{\bsnm{Hyeon}, \binits{J.D.}},
\bauthor{\bsnm{Park}, \binits{S.}},
\bauthor{\bsnm{Kim}, \binits{S.}},
\bauthor{\bsnm{Kim}, \binits{S.}},
\bauthor{\bsnm{Seo}, \binits{D.}}, \betal:
\bctitle{What changes can large-scale language models bring? intensive study on hyperclova: Billions-scale korean generative pretrained transformers}.
In: \bbtitle{Proceedings of the 2021 Conference on Empirical Methods in Natural Language Processing (EMNLP)},
pp. \bfpage{3405}--\blpage{3424}
(\byear{2021})
\end{bchapter}
\endbibitem

\bibitem[\protect\citeauthoryear{Tay et~al.}{2023}]{tay2023transcending}
\begin{bchapter}
\bauthor{\bsnm{Tay}, \binits{Y.}},
\bauthor{\bsnm{Wei}, \binits{J.}},
\bauthor{\bsnm{Chung}, \binits{H.}},
\bauthor{\bsnm{Tran}, \binits{V.}},
\bauthor{\bsnm{So}, \binits{D.}},
\bauthor{\bsnm{Shakeri}, \binits{S.}},
\bauthor{\bsnm{Garcia}, \binits{X.}},
\bauthor{\bsnm{Zheng}, \binits{S.}},
\bauthor{\bsnm{Rao}, \binits{J.}},
\bauthor{\bsnm{Chowdhery}, \binits{A.}}, \betal:
\bctitle{Transcending scaling laws with 0.1\% extra compute}.
In: \bbtitle{Proceedings of the 2023 Conference on Empirical Methods in Natural Language Processing (EMNLP)},
pp. \bfpage{1471}--\blpage{1486}
(\byear{2023})
\end{bchapter}
\endbibitem

\bibitem[\protect\citeauthoryear{Ahmad et~al.}{2021}]{ahmad2021unified}
\begin{bchapter}
\bauthor{\bsnm{Ahmad}, \binits{W.}},
\bauthor{\bsnm{Chakraborty}, \binits{S.}},
\bauthor{\bsnm{Ray}, \binits{B.}},
\bauthor{\bsnm{Chang}, \binits{K.}}:
\bctitle{Unified pre-training for program understanding and generation.}
In: \bbtitle{Proceedings of the 2021 Conference of the North American Chapter of the Association for Computational Linguistics: Human Language Technologies}
(\byear{2021})
\end{bchapter}
\endbibitem

\bibitem[\protect\citeauthoryear{Muennighoff et~al.}{2023}]{muennighoff2023crosslingual}
\begin{bchapter}
\bauthor{\bsnm{Muennighoff}, \binits{N.}},
\bauthor{\bsnm{Wang}, \binits{T.}},
\bauthor{\bsnm{Sutawika}, \binits{L.}},
\bauthor{\bsnm{Roberts}, \binits{A.}},
\bauthor{\bsnm{Biderman}, \binits{S.}},
\bauthor{\bsnm{Le~Scao}, \binits{T.}},
\bauthor{\bsnm{Bari}, \binits{M.S.}},
\bauthor{\bsnm{Shen}, \binits{S.}},
\bauthor{\bsnm{Yong}, \binits{Z.X.}},
\bauthor{\bsnm{Schoelkopf}, \binits{H.}}, \betal:
\bctitle{Crosslingual generalization through multitask finetuning}.
In: \bbtitle{Proceedings of the 61st Annual Meeting of the Association for Computational Linguistics (Volume 1: Long Papers)},
pp. \bfpage{15991}--\blpage{16111}
(\byear{2023})
\end{bchapter}
\endbibitem

\bibitem[\protect\citeauthoryear{Bang et~al.}{2023}]{bang2023multitask}
\begin{bchapter}
\bauthor{\bsnm{Bang}, \binits{Y.}},
\bauthor{\bsnm{Cahyawijaya}, \binits{S.}},
\bauthor{\bsnm{Lee}, \binits{N.}},
\bauthor{\bsnm{Dai}, \binits{W.}},
\bauthor{\bsnm{Su}, \binits{D.}},
\bauthor{\bsnm{Wilie}, \binits{B.}},
\bauthor{\bsnm{Lovenia}, \binits{H.}},
\bauthor{\bsnm{Ji}, \binits{Z.}},
\bauthor{\bsnm{Yu}, \binits{T.}},
\bauthor{\bsnm{Chung}, \binits{W.}}, \betal:
\bctitle{A multitask, multilingual, multimodal evaluation of chatgpt on reasoning, hallucination, and interactivity}.
In: \bbtitle{Proceedings of the 13th International Joint Conference on Natural Language Processing and the 3rd Conference of the Asia-Pacific Chapter of the Association for Computational Linguistics (Volume 1: Long Papers)},
pp. \bfpage{675}--\blpage{718}
(\byear{2023})
\end{bchapter}
\endbibitem

\bibitem[\protect\citeauthoryear{Sheng et~al.}{2021}]{sheng2021societal}
\begin{bchapter}
\bauthor{\bsnm{Sheng}, \binits{E.}},
\bauthor{\bsnm{Chang}, \binits{K.-W.}},
\bauthor{\bsnm{Natarajan}, \binits{P.}},
\bauthor{\bsnm{Peng}, \binits{N.}}:
\bctitle{Societal biases in language generation: Progress and challenges}.
In: \bbtitle{Proceedings of the 59th Annual Meeting of the Association for Computational Linguistics and the 11th International Joint Conference on Natural Language Processing (Volume 1: Long Papers)},
pp. \bfpage{4275}--\blpage{4293}
(\byear{2021})
\end{bchapter}
\endbibitem

\bibitem[\protect\citeauthoryear{Fried et~al.}{}]{fried2022incoder}
\begin{botherref}
\oauthor{\bsnm{Fried}, \binits{D.}},
\oauthor{\bsnm{Aghajanyan}, \binits{A.}},
\oauthor{\bsnm{Lin}, \binits{J.}},
\oauthor{\bsnm{Wang}, \binits{S.}},
\oauthor{\bsnm{Wallace}, \binits{E.}},
\oauthor{\bsnm{Shi}, \binits{F.}},
\oauthor{\bsnm{Zhong}, \binits{R.}},
\oauthor{\bsnm{Yih}, \binits{S.}},
\oauthor{\bsnm{Zettlemoyer}, \binits{L.}},
\oauthor{\bsnm{Lewis}, \binits{M.}}:
Incoder: A generative model for code infilling and synthesis.
In: The Eleventh International Conference on Learning Representations
\end{botherref}
\endbibitem

\bibitem[\protect\citeauthoryear{Nijkamp et~al.}{2022}]{nijkamp2022codegen}
\begin{bchapter}
\bauthor{\bsnm{Nijkamp}, \binits{E.}},
\bauthor{\bsnm{Pang}, \binits{B.}},
\bauthor{\bsnm{Hayashi}, \binits{H.}},
\bauthor{\bsnm{Tu}, \binits{L.}},
\bauthor{\bsnm{Wang}, \binits{H.}},
\bauthor{\bsnm{Zhou}, \binits{Y.}},
\bauthor{\bsnm{Savarese}, \binits{S.}},
\bauthor{\bsnm{Xiong}, \binits{C.}}:
\bctitle{Codegen: An open large language model for code with multi-turn program synthesis}.
In: \bbtitle{The Eleventh International Conference on Learning Representations}
(\byear{2022})
\end{bchapter}
\endbibitem

\bibitem[\protect\citeauthoryear{Wei et~al.}{2021}]{wei2021finetuned}
\begin{bchapter}
\bauthor{\bsnm{Wei}, \binits{J.}},
\bauthor{\bsnm{Bosma}, \binits{M.}},
\bauthor{\bsnm{Zhao}, \binits{V.}},
\bauthor{\bsnm{Guu}, \binits{K.}},
\bauthor{\bsnm{Yu}, \binits{A.W.}},
\bauthor{\bsnm{Lester}, \binits{B.}},
\bauthor{\bsnm{Du}, \binits{N.}},
\bauthor{\bsnm{Dai}, \binits{A.M.}},
\bauthor{\bsnm{Le}, \binits{Q.V.}}:
\bctitle{Finetuned language models are zero-shot learners}.
In: \bbtitle{International Conference on Learning Representations}
(\byear{2021})
\end{bchapter}
\endbibitem

\bibitem[\protect\citeauthoryear{Sanh et~al.}{2022}]{sanh2022multitask}
\begin{bchapter}
\bauthor{\bsnm{Sanh}, \binits{V.}},
\bauthor{\bsnm{Webson}, \binits{A.}},
\bauthor{\bsnm{Raffel}, \binits{C.}},
\bauthor{\bsnm{Bach}, \binits{S.H.}},
\bauthor{\bsnm{Sutawika}, \binits{L.}},
\bauthor{\bsnm{Alyafeai}, \binits{Z.}},
\bauthor{\bsnm{Chaffin}, \binits{A.}},
\bauthor{\bsnm{Stiegler}, \binits{A.}},
\bauthor{\bsnm{Le~Scao}, \binits{T.}},
\bauthor{\bsnm{Raja}, \binits{A.}}, \betal:
\bctitle{Multitask prompted training enables zero-shot task generalization}.
In: \bbtitle{ICLR 2022-Tenth International Conference on Learning Representations}
(\byear{2022})
\end{bchapter}
\endbibitem

\bibitem[\protect\citeauthoryear{Tay et~al.}{}]{tay2022ul2}
\begin{botherref}
\oauthor{\bsnm{Tay}, \binits{Y.}},
\oauthor{\bsnm{Dehghani}, \binits{M.}},
\oauthor{\bsnm{Tran}, \binits{V.Q.}},
\oauthor{\bsnm{Garcia}, \binits{X.}},
\oauthor{\bsnm{Wei}, \binits{J.}},
\oauthor{\bsnm{Wang}, \binits{X.}},
\oauthor{\bsnm{Chung}, \binits{H.W.}},
\oauthor{\bsnm{Bahri}, \binits{D.}},
\oauthor{\bsnm{Schuster}, \binits{T.}},
\oauthor{\bsnm{Zheng}, \binits{S.}}, et al.:
Ul2: Unifying language learning paradigms.
In: The Eleventh International Conference on Learning Representations
\end{botherref}
\endbibitem

\bibitem[\protect\citeauthoryear{Zeng et~al.}{}]{zeng2022glm}
\begin{botherref}
\oauthor{\bsnm{Zeng}, \binits{A.}},
\oauthor{\bsnm{Liu}, \binits{X.}},
\oauthor{\bsnm{Du}, \binits{Z.}},
\oauthor{\bsnm{Wang}, \binits{Z.}},
\oauthor{\bsnm{Lai}, \binits{H.}},
\oauthor{\bsnm{Ding}, \binits{M.}},
\oauthor{\bsnm{Yang}, \binits{Z.}},
\oauthor{\bsnm{Xu}, \binits{Y.}},
\oauthor{\bsnm{Zheng}, \binits{W.}},
\oauthor{\bsnm{Xia}, \binits{X.}}, et al.:
Glm-130b: An open bilingual pre-trained model.
In: The Eleventh International Conference on Learning Representations
\end{botherref}
\endbibitem

\bibitem[\protect\citeauthoryear{He et~al.}{}]{he2020deberta}
\begin{botherref}
\oauthor{\bsnm{He}, \binits{P.}},
\oauthor{\bsnm{Liu}, \binits{X.}},
\oauthor{\bsnm{Gao}, \binits{J.}},
\oauthor{\bsnm{Chen}, \binits{W.}}:
Deberta: Decoding-enhanced bert with disentangled attention.
In: International Conference on Learning Representations
\end{botherref}
\endbibitem

\bibitem[\protect\citeauthoryear{Merity et~al.}{2022}]{merity2022pointer}
\begin{botherref}
\oauthor{\bsnm{Merity}, \binits{S.}},
\oauthor{\bsnm{Xiong}, \binits{C.}},
\oauthor{\bsnm{Bradbury}, \binits{J.}},
\oauthor{\bsnm{Socher}, \binits{R.}}:
Pointer sentinel mixture models
(2022)
\end{botherref}
\endbibitem

\bibitem[\protect\citeauthoryear{Du et~al.}{2022}]{du2022glam}
\begin{bchapter}
\bauthor{\bsnm{Du}, \binits{N.}},
\bauthor{\bsnm{Huang}, \binits{Y.}},
\bauthor{\bsnm{Dai}, \binits{A.M.}},
\bauthor{\bsnm{Tong}, \binits{S.}},
\bauthor{\bsnm{Lepikhin}, \binits{D.}},
\bauthor{\bsnm{Xu}, \binits{Y.}},
\bauthor{\bsnm{Krikun}, \binits{M.}},
\bauthor{\bsnm{Zhou}, \binits{Y.}},
\bauthor{\bsnm{Yu}, \binits{A.W.}},
\bauthor{\bsnm{Firat}, \binits{O.}}, \betal:
\bctitle{Glam: Efficient scaling of language models with mixture-of-experts}.
In: \bbtitle{International Conference on Machine Learning},
pp. \bfpage{5547}--\blpage{5569}
(\byear{2022}).
\bcomment{PMLR}
\end{bchapter}
\endbibitem

\bibitem[\protect\citeauthoryear{Biderman et~al.}{2023}]{biderman2023pythia}
\begin{bchapter}
\bauthor{\bsnm{Biderman}, \binits{S.}},
\bauthor{\bsnm{Schoelkopf}, \binits{H.}},
\bauthor{\bsnm{Anthony}, \binits{Q.G.}},
\bauthor{\bsnm{Bradley}, \binits{H.}},
\bauthor{\bsnm{O’Brien}, \binits{K.}},
\bauthor{\bsnm{Hallahan}, \binits{E.}},
\bauthor{\bsnm{Khan}, \binits{M.A.}},
\bauthor{\bsnm{Purohit}, \binits{S.}},
\bauthor{\bsnm{Prashanth}, \binits{U.S.}},
\bauthor{\bsnm{Raff}, \binits{E.}}, \betal:
\bctitle{Pythia: A suite for analyzing large language models across training and scaling}.
In: \bbtitle{International Conference on Machine Learning},
pp. \bfpage{2397}--\blpage{2430}
(\byear{2023}).
\bcomment{PMLR}
\end{bchapter}
\endbibitem

\bibitem[\protect\citeauthoryear{Wang et~al.}{2022}]{wang2022language}
\begin{bchapter}
\bauthor{\bsnm{Wang}, \binits{T.}},
\bauthor{\bsnm{Roberts}, \binits{A.}},
\bauthor{\bsnm{Hesslow}, \binits{D.}},
\bauthor{\bsnm{Le~Scao}, \binits{T.}},
\bauthor{\bsnm{Chung}, \binits{H.W.}},
\bauthor{\bsnm{Beltagy}, \binits{I.}},
\bauthor{\bsnm{Launay}, \binits{J.}},
\bauthor{\bsnm{Raffel}, \binits{C.}}:
\bctitle{What language model architecture and pretraining objective works best for zero-shot generalization?}
In: \bbtitle{International Conference on Machine Learning},
pp. \bfpage{22964}--\blpage{22984}
(\byear{2022}).
\bcomment{PMLR}
\end{bchapter}
\endbibitem

\bibitem[\protect\citeauthoryear{Yu et~al.}{2023}]{yu2023codeipprompt}
\begin{bchapter}
\bauthor{\bsnm{Yu}, \binits{Z.}},
\bauthor{\bsnm{Wu}, \binits{Y.}},
\bauthor{\bsnm{Zhang}, \binits{N.}},
\bauthor{\bsnm{Wang}, \binits{C.}},
\bauthor{\bsnm{Vorobeychik}, \binits{Y.}},
\bauthor{\bsnm{Xiao}, \binits{C.}}:
\bctitle{Codeipprompt: intellectual property infringement assessment of code language models}.
In: \bbtitle{International Conference on Machine Learning},
pp. \bfpage{40373}--\blpage{40389}
(\byear{2023}).
\bcomment{PMLR}
\end{bchapter}
\endbibitem

\bibitem[\protect\citeauthoryear{Steenhoek et~al.}{2023}]{steenhoek2023empirical}
\begin{bchapter}
\bauthor{\bsnm{Steenhoek}, \binits{B.}},
\bauthor{\bsnm{Rahman}, \binits{M.M.}},
\bauthor{\bsnm{Jiles}, \binits{R.}},
\bauthor{\bsnm{Le}, \binits{W.}}:
\bctitle{An empirical study of deep learning models for vulnerability detection}.
In: \bbtitle{2023 IEEE/ACM 45th International Conference on Software Engineering (ICSE)},
pp. \bfpage{2237}--\blpage{2248}
(\byear{2023}).
\bcomment{IEEE}
\end{bchapter}
\endbibitem

\bibitem[\protect\citeauthoryear{Yin et~al.}{2024}]{yin2024improving}
\begin{bchapter}
\bauthor{\bsnm{Yin}, \binits{Z.}},
\bauthor{\bsnm{Wang}, \binits{Z.}},
\bauthor{\bsnm{Zhang}, \binits{W.}}:
\bctitle{Improving fairness in machine learning software via counterfactual fairness thinking}.
In: \bbtitle{Proceedings of the 2024 IEEE/ACM 46th International Conference on Software Engineering: Companion Proceedings},
pp. \bfpage{420}--\blpage{421}
(\byear{2024})
\end{bchapter}
\endbibitem

\bibitem[\protect\citeauthoryear{Li et~al.}{2023}]{li2023large}
\begin{bchapter}
\bauthor{\bsnm{Li}, \binits{Y.}},
\bauthor{\bsnm{Wang}, \binits{S.}},
\bauthor{\bsnm{Ding}, \binits{H.}},
\bauthor{\bsnm{Chen}, \binits{H.}}:
\bctitle{Large language models in finance: A survey}.
In: \bbtitle{Proceedings of the Fourth ACM International Conference on AI in Finance},
pp. \bfpage{374}--\blpage{382}
(\byear{2023})
\end{bchapter}
\endbibitem

\bibitem[\protect\citeauthoryear{Pagliaro et~al.}{2021}]{pagliaro2021investor}
\begin{bchapter}
\bauthor{\bsnm{Pagliaro}, \binits{C.}},
\bauthor{\bsnm{Mehta}, \binits{D.}},
\bauthor{\bsnm{Shiao}, \binits{H.-T.}},
\bauthor{\bsnm{Wang}, \binits{S.}},
\bauthor{\bsnm{Xiong}, \binits{L.}}:
\bctitle{Investor behavior modeling by analyzing financial advisor notes: a machine learning perspective}.
In: \bbtitle{Proceedings of the Second ACM International Conference on AI in Finance},
pp. \bfpage{1}--\blpage{8}
(\byear{2021})
\end{bchapter}
\endbibitem

\bibitem[\protect\citeauthoryear{Zhang and Weiss}{2021}]{zhang2021fair}
\begin{bchapter}
\bauthor{\bsnm{Zhang}, \binits{W.}},
\bauthor{\bsnm{Weiss}, \binits{J.}}:
\bctitle{Fair decision-making under uncertainty}.
In: \bbtitle{{2021 IEEE International Conference on Data Mining (ICDM)}}
(\byear{2021}).
\bcomment{IEEE}
\end{bchapter}
\endbibitem

\bibitem[\protect\citeauthoryear{Wang et~al.}{2023}]{wang2023mitigating}
\begin{bchapter}
\bauthor{\bsnm{Wang}, \binits{Z.}},
\bauthor{\bsnm{Narasimhan}, \binits{G.}},
\bauthor{\bsnm{Yao}, \binits{X.}},
\bauthor{\bsnm{Zhang}, \binits{W.}}:
\bctitle{Mitigating multisource biases in graph neural networks via real counterfactual samples}.
In: \bbtitle{2023 IEEE International Conference on Data Mining (ICDM)},
pp. \bfpage{638}--\blpage{647}
(\byear{2023}).
\bcomment{IEEE}
\end{bchapter}
\endbibitem

\bibitem[\protect\citeauthoryear{Chinta et~al.}{2023}]{chinta2023optimization}
\begin{bchapter}
\bauthor{\bsnm{Chinta}, \binits{S.V.}},
\bauthor{\bsnm{Fernandes}, \binits{K.}},
\bauthor{\bsnm{Cheng}, \binits{N.}},
\bauthor{\bsnm{Fernandez}, \binits{J.}},
\bauthor{\bsnm{Yazdani}, \binits{S.}},
\bauthor{\bsnm{Yin}, \binits{Z.}},
\bauthor{\bsnm{Wang}, \binits{Z.}},
\bauthor{\bsnm{Wang}, \binits{X.}},
\bauthor{\bsnm{Xu}, \binits{W.}},
\bauthor{\bsnm{Liu}, \binits{J.}}, \betal:
\bctitle{Optimization and improvement of fake news detection using voting technique for societal benefit}.
In: \bbtitle{2023 IEEE International Conference on Data Mining Workshops (ICDMW)},
pp. \bfpage{1565}--\blpage{1574}
(\byear{2023}).
\bcomment{IEEE}
\end{bchapter}
\endbibitem

\bibitem[\protect\citeauthoryear{Zhang and Ntoutsi}{2019}]{zhang2019faht}
\begin{bchapter}
\bauthor{\bsnm{Zhang}, \binits{W.}},
\bauthor{\bsnm{Ntoutsi}, \binits{E.}}:
\bctitle{Faht: an adaptive fairness-aware decision tree classifier}.
In: \bbtitle{International Joint Conference on Artificial Intelligence (IJCAI)},
pp. \bfpage{1480}--\blpage{1486}
(\byear{2019})
\end{bchapter}
\endbibitem

\bibitem[\protect\citeauthoryear{Zhang et~al.}{2021}]{zhang2021farf}
\begin{bchapter}
\bauthor{\bsnm{Zhang}, \binits{W.}},
\bauthor{\bsnm{Bifet}, \binits{A.}},
\bauthor{\bsnm{Zhang}, \binits{X.}},
\bauthor{\bsnm{Weiss}, \binits{J.C.}},
\bauthor{\bsnm{Nejdl}, \binits{W.}}:
\bctitle{Farf: A fair and adaptive random forests classifier}.
In: \bbtitle{Pacific-Asia Conference on Knowledge Discovery and Data Mining},
pp. \bfpage{245}--\blpage{256}
(\byear{2021}).
\bcomment{Springer}
\end{bchapter}
\endbibitem

\bibitem[\protect\citeauthoryear{Saxena et~al.}{2024}]{saxena2024unveiling}
\begin{botherref}
\oauthor{\bsnm{Saxena}, \binits{N.A.}},
\oauthor{\bsnm{Zhang}, \binits{W.}},
\oauthor{\bsnm{Shahabi}, \binits{C.}}:
Unveiling and mitigating bias in ride-hailing pricing for equitable policy making.
AI and Ethics,
1--12
(2024)
\end{botherref}
\endbibitem

\bibitem[\protect\citeauthoryear{Zhang and Weiss}{2022}]{zhang2022longitudinal}
\begin{bchapter}
\bauthor{\bsnm{Zhang}, \binits{W.}},
\bauthor{\bsnm{Weiss}, \binits{J.C.}}:
\bctitle{Longitudinal fairness with censorship}.
In: \bbtitle{Proceedings of the AAAI Conference on Artificial Intelligence},
vol. \bseriesno{36},
pp. \bfpage{12235}--\blpage{12243}
(\byear{2022})
\end{bchapter}
\endbibitem

\bibitem[\protect\citeauthoryear{Zhang}{2024}]{zhang2024fairness}
\begin{bchapter}
\bauthor{\bsnm{Zhang}, \binits{W.}}:
\bctitle{Fairness with censorship: Bridging the gap between fairness research and real-world deployment}.
In: \bbtitle{Proceedings of the AAAI Conference on Artificial Intelligence},
vol. \bseriesno{38},
pp. \bfpage{22685}--\blpage{22685}
(\byear{2024})
\end{bchapter}
\endbibitem

\bibitem[\protect\citeauthoryear{Zhang et~al.}{2023a}]{zhang2023censored}
\begin{bchapter}
\bauthor{\bsnm{Zhang}, \binits{W.}},
\bauthor{\bsnm{Hernandez-Boussard}, \binits{T.}},
\bauthor{\bsnm{Weiss}, \binits{J.}}:
\bctitle{Censored fairness through awareness}.
In: \bbtitle{Proceedings of the AAAI Conference on Artificial Intelligence},
vol. \bseriesno{37},
pp. \bfpage{14611}--\blpage{14619}
(\byear{2023})
\end{bchapter}
\endbibitem

\bibitem[\protect\citeauthoryear{Zhang et~al.}{2023b}]{zhang2023individual}
\begin{bchapter}
\bauthor{\bsnm{Zhang}, \binits{W.}},
\bauthor{\bsnm{Wang}, \binits{Z.}},
\bauthor{\bsnm{Kim}, \binits{J.}},
\bauthor{\bsnm{Cheng}, \binits{C.}},
\bauthor{\bsnm{Oommen}, \binits{T.}},
\bauthor{\bsnm{Ravikumar}, \binits{P.}},
\bauthor{\bsnm{Weiss}, \binits{J.}}:
\bctitle{Individual fairness under uncertainty}.
In: \bbtitle{26th European Conference on Artificial Intelligence},
pp. \bfpage{3042}--\blpage{3049}
(\byear{2023})
\end{bchapter}
\endbibitem

\bibitem[\protect\citeauthoryear{Wang and Zhang}{20234}]{wang2024group}
\begin{bchapter}
\bauthor{\bsnm{Wang}, \binits{Z.}},
\bauthor{\bsnm{Zhang}, \binits{W.}}:
\bctitle{Group fairness with individual and censorship constraints}.
In: \bbtitle{26th European Conference on Artificial Intelligence}
(\byear{20234})
\end{bchapter}
\endbibitem

\bibitem[\protect\citeauthoryear{Wang et~al.}{20234}]{wang2024individual1}
\begin{bchapter}
\bauthor{\bsnm{Wang}, \binits{Z.}},
\bauthor{\bsnm{Ulloa}, \binits{D.}},
\bauthor{\bsnm{Yu}, \binits{T.}},
\bauthor{\bsnm{Rangaswami}, \binits{R.}},
\bauthor{\bsnm{Yap}, \binits{R.}},
\bauthor{\bsnm{Zhang}, \binits{W.}}:
\bctitle{Individual fairness with group constraints in graph neural networks}.
In: \bbtitle{26th European Conference on Artificial Intelligence}
(\byear{20234})
\end{bchapter}
\endbibitem

\bibitem[\protect\citeauthoryear{Xiao et~al.}{2023}]{xiao2023evaluating}
\begin{bchapter}
\bauthor{\bsnm{Xiao}, \binits{C.}},
\bauthor{\bsnm{Xu}, \binits{S.X.}},
\bauthor{\bsnm{Zhang}, \binits{K.}},
\bauthor{\bsnm{Wang}, \binits{Y.}},
\bauthor{\bsnm{Xia}, \binits{L.}}:
\bctitle{Evaluating reading comprehension exercises generated by llms: A showcase of chatgpt in education applications}.
In: \bbtitle{Proceedings of the 18th Workshop on Innovative Use of NLP for Building Educational Applications (BEA 2023)},
pp. \bfpage{610}--\blpage{625}
(\byear{2023})
\end{bchapter}
\endbibitem

\bibitem[\protect\citeauthoryear{Gupta et~al.}{2020}]{gupta2020comprehensive}
\begin{barticle}
\bauthor{\bsnm{Gupta}, \binits{A.}},
\bauthor{\bsnm{Dengre}, \binits{V.}},
\bauthor{\bsnm{Kheruwala}, \binits{H.A.}},
\bauthor{\bsnm{Shah}, \binits{M.}}:
\batitle{Comprehensive review of text-mining applications in finance}.
\bjtitle{Financial Innovation}
\bvolume{6},
\bfpage{1}--\blpage{25}
(\byear{2020})
\end{barticle}
\endbibitem

\bibitem[\protect\citeauthoryear{Kung et~al.}{2023}]{kung2023performance}
\begin{barticle}
\bauthor{\bsnm{Kung}, \binits{T.H.}},
\bauthor{\bsnm{Cheatham}, \binits{M.}},
\bauthor{\bsnm{Medenilla}, \binits{A.}},
\bauthor{\bsnm{Sillos}, \binits{C.}},
\bauthor{\bsnm{De~Leon}, \binits{L.}},
\bauthor{\bsnm{Elepa{\~n}o}, \binits{C.}},
\bauthor{\bsnm{Madriaga}, \binits{M.}},
\bauthor{\bsnm{Aggabao}, \binits{R.}},
\bauthor{\bsnm{Diaz-Candido}, \binits{G.}},
\bauthor{\bsnm{Maningo}, \binits{J.}}, \betal:
\batitle{Performance of chatgpt on usmle: potential for ai-assisted medical education using large language models}.
\bjtitle{PLoS digital health}
\bvolume{2}(\bissue{2}),
\bfpage{0000198}
(\byear{2023})
\end{barticle}
\endbibitem

\bibitem[\protect\citeauthoryear{Mozafari et~al.}{2020}]{mozafari2020hate}
\begin{barticle}
\bauthor{\bsnm{Mozafari}, \binits{M.}},
\bauthor{\bsnm{Farahbakhsh}, \binits{R.}},
\bauthor{\bsnm{Crespi}, \binits{N.}}:
\batitle{Hate speech detection and racial bias mitigation in social media based on bert model}.
\bjtitle{PloS one}
\bvolume{15}(\bissue{8}),
\bfpage{0237861}
(\byear{2020})
\end{barticle}
\endbibitem

\bibitem[\protect\citeauthoryear{Jin et~al.}{2021}]{jin2021disease}
\begin{barticle}
\bauthor{\bsnm{Jin}, \binits{D.}},
\bauthor{\bsnm{Pan}, \binits{E.}},
\bauthor{\bsnm{Oufattole}, \binits{N.}},
\bauthor{\bsnm{Weng}, \binits{W.-H.}},
\bauthor{\bsnm{Fang}, \binits{H.}},
\bauthor{\bsnm{Szolovits}, \binits{P.}}:
\batitle{What disease does this patient have? a large-scale open domain question answering dataset from medical exams}.
\bjtitle{Applied Sciences}
\bvolume{11}(\bissue{14}),
\bfpage{6421}
(\byear{2021})
\end{barticle}
\endbibitem

\bibitem[\protect\citeauthoryear{Kombrink et~al.}{2011}]{kombrink2011recurrent}
\begin{bchapter}
\bauthor{\bsnm{Kombrink}, \binits{S.}},
\bauthor{\bsnm{Mikolov}, \binits{T.}},
\bauthor{\bsnm{Karafi{\'a}t}, \binits{M.}},
\bauthor{\bsnm{Burget}, \binits{L.}}:
\bctitle{Recurrent neural network based language modeling in meeting recognition.}
In: \bbtitle{Interspeech},
vol. \bseriesno{11},
pp. \bfpage{2877}--\blpage{2880}
(\byear{2011})
\end{bchapter}
\endbibitem

\bibitem[\protect\citeauthoryear{Mikolov et~al.}{2010}]{mikolov2010recurrent}
\begin{bchapter}
\bauthor{\bsnm{Mikolov}, \binits{T.}},
\bauthor{\bsnm{Karafi{\'a}t}, \binits{M.}},
\bauthor{\bsnm{Burget}, \binits{L.}},
\bauthor{\bsnm{Cernock{\`y}}, \binits{J.}},
\bauthor{\bsnm{Khudanpur}, \binits{S.}}:
\bctitle{Recurrent neural network based language model.}
In: \bbtitle{Interspeech},
vol. \bseriesno{2},
pp. \bfpage{1045}--\blpage{1048}
(\byear{2010}).
\bcomment{Makuhari}
\end{bchapter}
\endbibitem

\bibitem[\protect\citeauthoryear{Stolcke et~al.}{2002}]{stolcke2002srilm}
\begin{bchapter}
\bauthor{\bsnm{Stolcke}, \binits{A.}}, \betal:
\bctitle{Srilm-an extensible language modeling toolkit.}
In: \bbtitle{Interspeech},
vol. \bseriesno{2002},
p. \bfpage{2002}
(\byear{2002})
\end{bchapter}
\endbibitem

\bibitem[\protect\citeauthoryear{Chung et~al.}{2024}]{chung2024scaling}
\begin{barticle}
\bauthor{\bsnm{Chung}, \binits{H.W.}},
\bauthor{\bsnm{Hou}, \binits{L.}},
\bauthor{\bsnm{Longpre}, \binits{S.}},
\bauthor{\bsnm{Zoph}, \binits{B.}},
\bauthor{\bsnm{Tay}, \binits{Y.}},
\bauthor{\bsnm{Fedus}, \binits{W.}},
\bauthor{\bsnm{Li}, \binits{Y.}},
\bauthor{\bsnm{Wang}, \binits{X.}},
\bauthor{\bsnm{Dehghani}, \binits{M.}},
\bauthor{\bsnm{Brahma}, \binits{S.}}, \betal:
\batitle{Scaling instruction-finetuned language models}.
\bjtitle{Journal of Machine Learning Research}
\bvolume{25}(\bissue{70}),
\bfpage{1}--\blpage{53}
(\byear{2024})
\end{barticle}
\endbibitem

\bibitem[\protect\citeauthoryear{Chowdhery et~al.}{2023}]{chowdhery2023palm}
\begin{barticle}
\bauthor{\bsnm{Chowdhery}, \binits{A.}},
\bauthor{\bsnm{Narang}, \binits{S.}},
\bauthor{\bsnm{Devlin}, \binits{J.}},
\bauthor{\bsnm{Bosma}, \binits{M.}},
\bauthor{\bsnm{Mishra}, \binits{G.}},
\bauthor{\bsnm{Roberts}, \binits{A.}},
\bauthor{\bsnm{Barham}, \binits{P.}},
\bauthor{\bsnm{Chung}, \binits{H.W.}},
\bauthor{\bsnm{Sutton}, \binits{C.}},
\bauthor{\bsnm{Gehrmann}, \binits{S.}}, \betal:
\batitle{Palm: Scaling language modeling with pathways}.
\bjtitle{Journal of Machine Learning Research}
\bvolume{24}(\bissue{240}),
\bfpage{1}--\blpage{113}
(\byear{2023})
\end{barticle}
\endbibitem

\bibitem[\protect\citeauthoryear{Fedus et~al.}{2022}]{fedus2022switch}
\begin{barticle}
\bauthor{\bsnm{Fedus}, \binits{W.}},
\bauthor{\bsnm{Zoph}, \binits{B.}},
\bauthor{\bsnm{Shazeer}, \binits{N.}}:
\batitle{Switch transformers: Scaling to trillion parameter models with simple and efficient sparsity}.
\bjtitle{Journal of Machine Learning Research}
\bvolume{23}(\bissue{120}),
\bfpage{1}--\blpage{39}
(\byear{2022})
\end{barticle}
\endbibitem

\bibitem[\protect\citeauthoryear{Raffel et~al.}{2020}]{raffel2020exploring}
\begin{barticle}
\bauthor{\bsnm{Raffel}, \binits{C.}},
\bauthor{\bsnm{Shazeer}, \binits{N.}},
\bauthor{\bsnm{Roberts}, \binits{A.}},
\bauthor{\bsnm{Lee}, \binits{K.}},
\bauthor{\bsnm{Narang}, \binits{S.}},
\bauthor{\bsnm{Matena}, \binits{M.}},
\bauthor{\bsnm{Zhou}, \binits{Y.}},
\bauthor{\bsnm{Li}, \binits{W.}},
\bauthor{\bsnm{Liu}, \binits{P.J.}}:
\batitle{Exploring the limits of transfer learning with a unified text-to-text transformer}.
\bjtitle{Journal of machine learning research}
\bvolume{21}(\bissue{140}),
\bfpage{1}--\blpage{67}
(\byear{2020})
\end{barticle}
\endbibitem

\bibitem[\protect\citeauthoryear{Baidoo-Anu and Ansah}{2023}]{baidoo2023education}
\begin{barticle}
\bauthor{\bsnm{Baidoo-Anu}, \binits{D.}},
\bauthor{\bsnm{Ansah}, \binits{L.O.}}:
\batitle{Education in the era of generative artificial intelligence (ai): Understanding the potential benefits of chatgpt in promoting teaching and learning}.
\bjtitle{Journal of AI}
\bvolume{7}(\bissue{1}),
\bfpage{52}--\blpage{62}
(\byear{2023})
\end{barticle}
\endbibitem

\bibitem[\protect\citeauthoryear{Chen et~al.}{2024}]{chen2024survey}
\begin{botherref}
\oauthor{\bsnm{Chen}, \binits{Z.Z.}},
\oauthor{\bsnm{Ma}, \binits{J.}},
\oauthor{\bsnm{Zhang}, \binits{X.}},
\oauthor{\bsnm{Hao}, \binits{N.}},
\oauthor{\bsnm{Yan}, \binits{A.}},
\oauthor{\bsnm{Nourbakhsh}, \binits{A.}},
\oauthor{\bsnm{Yang}, \binits{X.}},
\oauthor{\bsnm{McAuley}, \binits{J.}},
\oauthor{\bsnm{Petzold}, \binits{L.}},
\oauthor{\bsnm{Wang}, \binits{W.Y.}}:
A survey on large language models for critical societal domains: Finance, healthcare, and law.
arXiv preprint arXiv:2405.01769
(2024)
\end{botherref}
\endbibitem

\bibitem[\protect\citeauthoryear{Yao et~al.}{2024}]{yao2024survey}
\begin{botherref}
\oauthor{\bsnm{Yao}, \binits{Y.}},
\oauthor{\bsnm{Duan}, \binits{J.}},
\oauthor{\bsnm{Xu}, \binits{K.}},
\oauthor{\bsnm{Cai}, \binits{Y.}},
\oauthor{\bsnm{Sun}, \binits{Z.}},
\oauthor{\bsnm{Zhang}, \binits{Y.}}:
A survey on large language model (llm) security and privacy: The good, the bad, and the ugly.
High-Confidence Computing,
100211
(2024)
\end{botherref}
\endbibitem

\bibitem[\protect\citeauthoryear{Huang et~al.}{2023}]{huang2023survey}
\begin{botherref}
\oauthor{\bsnm{Huang}, \binits{L.}},
\oauthor{\bsnm{Yu}, \binits{W.}},
\oauthor{\bsnm{Ma}, \binits{W.}},
\oauthor{\bsnm{Zhong}, \binits{W.}},
\oauthor{\bsnm{Feng}, \binits{Z.}},
\oauthor{\bsnm{Wang}, \binits{H.}},
\oauthor{\bsnm{Chen}, \binits{Q.}},
\oauthor{\bsnm{Peng}, \binits{W.}},
\oauthor{\bsnm{Feng}, \binits{X.}},
\oauthor{\bsnm{Qin}, \binits{B.}}, et al.:
A survey on hallucination in large language models: Principles, taxonomy, challenges, and open questions.
arXiv preprint arXiv:2311.05232
(2023)
\end{botherref}
\endbibitem

\bibitem[\protect\citeauthoryear{Jelinek}{1998}]{jelinek1998statistical}
\begin{botherref}
\oauthor{\bsnm{Jelinek}, \binits{F.}}:
Statistical methods for speech recognition.
MIT Press
(1998)
\end{botherref}
\endbibitem

\bibitem[\protect\citeauthoryear{Rosenfeld}{2000}]{rosenfeld2000two}
\begin{barticle}
\bauthor{\bsnm{Rosenfeld}, \binits{R.}}:
\batitle{Two decades of statistical language modeling: Where do we go from here?}
\bjtitle{Proceedings of the IEEE}
\bvolume{88}(\bissue{8}),
\bfpage{1270}--\blpage{1278}
(\byear{2000})
\end{barticle}
\endbibitem

\bibitem[\protect\citeauthoryear{Bengio et~al.}{2000}]{bengio2000neural}
\begin{botherref}
\oauthor{\bsnm{Bengio}, \binits{Y.}},
\oauthor{\bsnm{Ducharme}, \binits{R.}},
\oauthor{\bsnm{Vincent}, \binits{P.}}:
A neural probabilistic language model.
Advances in neural information processing systems
\textbf{13}
(2000)
\end{botherref}
\endbibitem

\bibitem[\protect\citeauthoryear{Mikolov et~al.}{2013}]{mikolov2013distributed}
\begin{botherref}
\oauthor{\bsnm{Mikolov}, \binits{T.}},
\oauthor{\bsnm{Sutskever}, \binits{I.}},
\oauthor{\bsnm{Chen}, \binits{K.}},
\oauthor{\bsnm{Corrado}, \binits{G.S.}},
\oauthor{\bsnm{Dean}, \binits{J.}}:
Distributed representations of words and phrases and their compositionality.
Advances in neural information processing systems
\textbf{26}
(2013)
\end{botherref}
\endbibitem

\bibitem[\protect\citeauthoryear{Devlin et~al.}{2018}]{devlin2018bert}
\begin{botherref}
\oauthor{\bsnm{Devlin}, \binits{J.}},
\oauthor{\bsnm{Chang}, \binits{M.-W.}},
\oauthor{\bsnm{Lee}, \binits{K.}},
\oauthor{\bsnm{Toutanova}, \binits{K.}}:
Bert: Pre-training of deep bidirectional transformers for language understanding.
arXiv preprint arXiv:1810.04805
(2018)
\end{botherref}
\endbibitem

\bibitem[\protect\citeauthoryear{Touvron et~al.}{2023}]{touvron2023llama}
\begin{botherref}
\oauthor{\bsnm{Touvron}, \binits{H.}},
\oauthor{\bsnm{Lavril}, \binits{T.}},
\oauthor{\bsnm{Izacard}, \binits{G.}},
\oauthor{\bsnm{Martinet}, \binits{X.}},
\oauthor{\bsnm{Lachaux}, \binits{M.-A.}},
\oauthor{\bsnm{Lacroix}, \binits{T.}},
\oauthor{\bsnm{Roziere}, \binits{B.}},
\oauthor{\bsnm{Goyal}, \binits{N.}},
\oauthor{\bsnm{Hambro}, \binits{E.}},
\oauthor{\bsnm{Azhar}, \binits{F.}}, et al.:
Llama: Open and efficient foundation language models.
arXiv preprint arXiv:2302.13971
(2023)
\end{botherref}
\endbibitem

\bibitem[\protect\citeauthoryear{Jozefowicz et~al.}{2016}]{jozefowicz2016exploring}
\begin{botherref}
\oauthor{\bsnm{Jozefowicz}, \binits{R.}},
\oauthor{\bsnm{Vinyals}, \binits{O.}},
\oauthor{\bsnm{Schuster}, \binits{M.}},
\oauthor{\bsnm{Shazeer}, \binits{N.}},
\oauthor{\bsnm{Wu}, \binits{Y.}}:
Exploring the limits of language modeling.
arXiv preprint arXiv:1602.02410
(2016)
\end{botherref}
\endbibitem

\bibitem[\protect\citeauthoryear{Kiros et~al.}{2015}]{kiros2015skip}
\begin{botherref}
\oauthor{\bsnm{Kiros}, \binits{R.}},
\oauthor{\bsnm{Zhu}, \binits{Y.}},
\oauthor{\bsnm{Salakhutdinov}, \binits{R.R.}},
\oauthor{\bsnm{Zemel}, \binits{R.}},
\oauthor{\bsnm{Urtasun}, \binits{R.}},
\oauthor{\bsnm{Torralba}, \binits{A.}},
\oauthor{\bsnm{Fidler}, \binits{S.}}:
Skip-thought vectors.
Advances in neural information processing systems
\textbf{28}
(2015)
\end{botherref}
\endbibitem

\bibitem[\protect\citeauthoryear{Gao et~al.}{2020}]{gao2020pile}
\begin{botherref}
\oauthor{\bsnm{Gao}, \binits{L.}},
\oauthor{\bsnm{Biderman}, \binits{S.}},
\oauthor{\bsnm{Black}, \binits{S.}},
\oauthor{\bsnm{Golding}, \binits{L.}},
\oauthor{\bsnm{Hoppe}, \binits{T.}},
\oauthor{\bsnm{Foster}, \binits{C.}},
\oauthor{\bsnm{Phang}, \binits{J.}},
\oauthor{\bsnm{He}, \binits{H.}},
\oauthor{\bsnm{Thite}, \binits{A.}},
\oauthor{\bsnm{Nabeshima}, \binits{N.}}, et al.:
The pile: An 800gb dataset of diverse text for language modeling. arxiv.
arXiv preprint arXiv:2101.00027
(2020)
\end{botherref}
\endbibitem

\bibitem[\protect\citeauthoryear{Kaplan et~al.}{2020}]{kaplan2020scaling}
\begin{botherref}
\oauthor{\bsnm{Kaplan}, \binits{J.}},
\oauthor{\bsnm{McCandlish}, \binits{S.}},
\oauthor{\bsnm{Henighan}, \binits{T.}},
\oauthor{\bsnm{Brown}, \binits{T.B.}},
\oauthor{\bsnm{Chess}, \binits{B.}},
\oauthor{\bsnm{Child}, \binits{R.}},
\oauthor{\bsnm{Gray}, \binits{S.}},
\oauthor{\bsnm{Radford}, \binits{A.}},
\oauthor{\bsnm{Wu}, \binits{J.}},
\oauthor{\bsnm{Amodei}, \binits{D.}}:
Scaling laws for neural language models.
arXiv preprint arXiv:2001.08361
(2020)
\end{botherref}
\endbibitem

\bibitem[\protect\citeauthoryear{Vaswani et~al.}{2017}]{vaswani2017attention}
\begin{botherref}
\oauthor{\bsnm{Vaswani}, \binits{A.}},
\oauthor{\bsnm{Shazeer}, \binits{N.}},
\oauthor{\bsnm{Parmar}, \binits{N.}},
\oauthor{\bsnm{Uszkoreit}, \binits{J.}},
\oauthor{\bsnm{Jones}, \binits{L.}},
\oauthor{\bsnm{Gomez}, \binits{A.N.}},
\oauthor{\bsnm{Kaiser}, \binits{{\L}.}},
\oauthor{\bsnm{Polosukhin}, \binits{I.}}:
Attention is all you need.
Advances in neural information processing systems
\textbf{30}
(2017)
\end{botherref}
\endbibitem

\bibitem[\protect\citeauthoryear{LeCun et~al.}{1998}]{lecun1998gradient}
\begin{barticle}
\bauthor{\bsnm{LeCun}, \binits{Y.}},
\bauthor{\bsnm{Bottou}, \binits{L.}},
\bauthor{\bsnm{Bengio}, \binits{Y.}},
\bauthor{\bsnm{Haffner}, \binits{P.}}:
\batitle{Gradient-based learning applied to document recognition}.
\bjtitle{Proceedings of the IEEE}
\bvolume{86}(\bissue{11}),
\bfpage{2278}--\blpage{2324}
(\byear{1998})
\end{barticle}
\endbibitem

\bibitem[\protect\citeauthoryear{Cui et~al.}{2023}]{cui2023chatlaw}
\begin{botherref}
\oauthor{\bsnm{Cui}, \binits{J.}},
\oauthor{\bsnm{Li}, \binits{Z.}},
\oauthor{\bsnm{Yan}, \binits{Y.}},
\oauthor{\bsnm{Chen}, \binits{B.}},
\oauthor{\bsnm{Yuan}, \binits{L.}}:
Chatlaw: Open-source legal large language model with integrated external knowledge bases.
arXiv preprint arXiv:2306.16092
(2023)
\end{botherref}
\endbibitem

\bibitem[\protect\citeauthoryear{Yang et~al.}{2019}]{yang2019xlnet}
\begin{botherref}
\oauthor{\bsnm{Yang}, \binits{Z.}},
\oauthor{\bsnm{Dai}, \binits{Z.}},
\oauthor{\bsnm{Yang}, \binits{Y.}},
\oauthor{\bsnm{Carbonell}, \binits{J.}},
\oauthor{\bsnm{Salakhutdinov}, \binits{R.R.}},
\oauthor{\bsnm{Le}, \binits{Q.V.}}:
Xlnet: Generalized autoregressive pretraining for language understanding.
Advances in neural information processing systems
\textbf{32}
(2019)
\end{botherref}
\endbibitem

\bibitem[\protect\citeauthoryear{Lan et~al.}{2019}]{lan2019albert}
\begin{botherref}
\oauthor{\bsnm{Lan}, \binits{Z.}},
\oauthor{\bsnm{Chen}, \binits{M.}},
\oauthor{\bsnm{Goodman}, \binits{S.}},
\oauthor{\bsnm{Gimpel}, \binits{K.}},
\oauthor{\bsnm{Sharma}, \binits{P.}},
\oauthor{\bsnm{Soricut}, \binits{R.}}:
Albert: A lite bert for self-supervised learning of language representations.
arXiv preprint arXiv:1909.11942
(2019)
\end{botherref}
\endbibitem

\bibitem[\protect\citeauthoryear{Kaiming et~al.}{2016}]{kaiming2016deep}
\begin{bchapter}
\bauthor{\bsnm{Kaiming}, \binits{H.}},
\bauthor{\bsnm{Xiangyu}, \binits{Z.}},
\bauthor{\bsnm{Shaoqing}, \binits{R.}},
\bauthor{\bsnm{Jian}, \binits{S.}}, \betal:
\bctitle{Deep residual learning for image recognition}.
In: \bbtitle{Proceedings of the IEEE Conference on Computer Vision and Pattern Recognition},
vol. \bseriesno{34},
pp. \bfpage{770}--\blpage{778}
(\byear{2016})
\end{bchapter}
\endbibitem

\bibitem[\protect\citeauthoryear{Ba et~al.}{2016}]{ba2016layer}
\begin{botherref}
\oauthor{\bsnm{Ba}, \binits{J.L.}},
\oauthor{\bsnm{Kiros}, \binits{J.R.}},
\oauthor{\bsnm{Hinton}, \binits{G.E.}}:
Layer normalization.
arXiv preprint arXiv:1607.06450
(2016)
\end{botherref}
\endbibitem

\bibitem[\protect\citeauthoryear{Wang et~al.}{2019}]{wang2019language}
\begin{botherref}
\oauthor{\bsnm{Wang}, \binits{C.}},
\oauthor{\bsnm{Li}, \binits{M.}},
\oauthor{\bsnm{Smola}, \binits{A.J.}}:
Language models with transformers.
arXiv preprint arXiv:1904.09408
(2019)
\end{botherref}
\endbibitem

\bibitem[\protect\citeauthoryear{Chen et~al.}{2021}]{chen2021evaluating}
\begin{botherref}
\oauthor{\bsnm{Chen}, \binits{M.}},
\oauthor{\bsnm{Tworek}, \binits{J.}},
\oauthor{\bsnm{Jun}, \binits{H.}},
\oauthor{\bsnm{Yuan}, \binits{Q.}},
\oauthor{\bsnm{Pinto}, \binits{H.P.D.O.}},
\oauthor{\bsnm{Kaplan}, \binits{J.}},
\oauthor{\bsnm{Edwards}, \binits{H.}},
\oauthor{\bsnm{Burda}, \binits{Y.}},
\oauthor{\bsnm{Joseph}, \binits{N.}},
\oauthor{\bsnm{Brockman}, \binits{G.}}, et al.:
Evaluating large language models trained on code.
arXiv preprint arXiv:2107.03374
(2021)
\end{botherref}
\endbibitem

\bibitem[\protect\citeauthoryear{Clark}{2020}]{clark2020electra}
\begin{botherref}
\oauthor{\bsnm{Clark}, \binits{K.}}:
Electra: Pre-training text encoders as discriminators rather than generators.
arXiv preprint arXiv:2003.10555
(2020)
\end{botherref}
\endbibitem

\bibitem[\protect\citeauthoryear{Zhuang et~al.}{2021}]{zhuang2021robustly}
\begin{bchapter}
\bauthor{\bsnm{Zhuang}, \binits{L.}},
\bauthor{\bsnm{Wayne}, \binits{L.}},
\bauthor{\bsnm{Ya}, \binits{S.}},
\bauthor{\bsnm{Jun}, \binits{Z.}}:
\bctitle{A robustly optimized bert pre-training approach with post-training}.
In: \bbtitle{Proceedings of the 20th Chinese National Conference on Computational Linguistics},
pp. \bfpage{1218}--\blpage{1227}
(\byear{2021})
\end{bchapter}
\endbibitem

\bibitem[\protect\citeauthoryear{Conneau}{2019}]{conneau2019unsupervised}
\begin{botherref}
\oauthor{\bsnm{Conneau}, \binits{A.}}:
Unsupervised cross-lingual representation learning at scale.
arXiv preprint arXiv:1911.02116
(2019)
\end{botherref}
\endbibitem

\bibitem[\protect\citeauthoryear{Sun et~al.}{2021}]{sun2021ernie}
\begin{botherref}
\oauthor{\bsnm{Sun}, \binits{Y.}},
\oauthor{\bsnm{Wang}, \binits{S.}},
\oauthor{\bsnm{Feng}, \binits{S.}},
\oauthor{\bsnm{Ding}, \binits{S.}},
\oauthor{\bsnm{Pang}, \binits{C.}},
\oauthor{\bsnm{Shang}, \binits{J.}},
\oauthor{\bsnm{Liu}, \binits{J.}},
\oauthor{\bsnm{Chen}, \binits{X.}},
\oauthor{\bsnm{Zhao}, \binits{Y.}},
\oauthor{\bsnm{Lu}, \binits{Y.}}, et al.:
Ernie 3.0: Large-scale knowledge enhanced pre-training for language understanding and generation.
arXiv preprint arXiv:2107.02137
(2021)
\end{botherref}
\endbibitem

\bibitem[\protect\citeauthoryear{Soltan et~al.}{2022}]{soltan2022alexatm}
\begin{botherref}
\oauthor{\bsnm{Soltan}, \binits{S.}},
\oauthor{\bsnm{Ananthakrishnan}, \binits{S.}},
\oauthor{\bsnm{FitzGerald}, \binits{J.}},
\oauthor{\bsnm{Gupta}, \binits{R.}},
\oauthor{\bsnm{Hamza}, \binits{W.}},
\oauthor{\bsnm{Khan}, \binits{H.}},
\oauthor{\bsnm{Peris}, \binits{C.}},
\oauthor{\bsnm{Rawls}, \binits{S.}},
\oauthor{\bsnm{Rosenbaum}, \binits{A.}},
\oauthor{\bsnm{Rumshisky}, \binits{A.}}, et al.:
Alexatm 20b: Few-shot learning using a large-scale multilingual seq2seq model.
arXiv preprint arXiv:2208.01448
(2022)
\end{botherref}
\endbibitem

\bibitem[\protect\citeauthoryear{Li et~al.}{2022}]{li2022competition}
\begin{barticle}
\bauthor{\bsnm{Li}, \binits{Y.}},
\bauthor{\bsnm{Choi}, \binits{D.}},
\bauthor{\bsnm{Chung}, \binits{J.}},
\bauthor{\bsnm{Kushman}, \binits{N.}},
\bauthor{\bsnm{Schrittwieser}, \binits{J.}},
\bauthor{\bsnm{Leblond}, \binits{R.}},
\bauthor{\bsnm{Eccles}, \binits{T.}},
\bauthor{\bsnm{Keeling}, \binits{J.}},
\bauthor{\bsnm{Gimeno}, \binits{F.}},
\bauthor{\bsnm{Dal~Lago}, \binits{A.}}, \betal:
\batitle{Competition-level code generation with alphacode}.
\bjtitle{Science}
\bvolume{378}(\bissue{6624}),
\bfpage{1092}--\blpage{1097}
(\byear{2022})
\end{barticle}
\endbibitem

\bibitem[\protect\citeauthoryear{Askell et~al.}{2021}]{askell2021general}
\begin{botherref}
\oauthor{\bsnm{Askell}, \binits{A.}},
\oauthor{\bsnm{Bai}, \binits{Y.}},
\oauthor{\bsnm{Chen}, \binits{A.}},
\oauthor{\bsnm{Drain}, \binits{D.}},
\oauthor{\bsnm{Ganguli}, \binits{D.}},
\oauthor{\bsnm{Henighan}, \binits{T.}},
\oauthor{\bsnm{Jones}, \binits{A.}},
\oauthor{\bsnm{Joseph}, \binits{N.}},
\oauthor{\bsnm{Mann}, \binits{B.}},
\oauthor{\bsnm{DasSarma}, \binits{N.}}, et al.:
A general language assistant as a laboratory for alignment.
arXiv preprint arXiv:2112.00861
(2021)
\end{botherref}
\endbibitem

\bibitem[\protect\citeauthoryear{Costa-juss{\`a} et~al.}{2022}]{costa2022no}
\begin{botherref}
\oauthor{\bsnm{Costa-juss{\`a}}, \binits{M.R.}},
\oauthor{\bsnm{Cross}, \binits{J.}},
\oauthor{\bsnm{{\c{C}}elebi}, \binits{O.}},
\oauthor{\bsnm{Elbayad}, \binits{M.}},
\oauthor{\bsnm{Heafield}, \binits{K.}},
\oauthor{\bsnm{Heffernan}, \binits{K.}},
\oauthor{\bsnm{Kalbassi}, \binits{E.}},
\oauthor{\bsnm{Lam}, \binits{J.}},
\oauthor{\bsnm{Licht}, \binits{D.}},
\oauthor{\bsnm{Maillard}, \binits{J.}}, et al.:
No language left behind: Scaling human-centered machine translation.
arXiv preprint arXiv:2207.04672
(2022)
\end{botherref}
\endbibitem

\bibitem[\protect\citeauthoryear{Glaese et~al.}{2022}]{glaese2022improving}
\begin{botherref}
\oauthor{\bsnm{Glaese}, \binits{A.}},
\oauthor{\bsnm{McAleese}, \binits{N.}},
\oauthor{\bsnm{Tr{\k{e}}bacz}, \binits{M.}},
\oauthor{\bsnm{Aslanides}, \binits{J.}},
\oauthor{\bsnm{Firoiu}, \binits{V.}},
\oauthor{\bsnm{Ewalds}, \binits{T.}},
\oauthor{\bsnm{Rauh}, \binits{M.}},
\oauthor{\bsnm{Weidinger}, \binits{L.}},
\oauthor{\bsnm{Chadwick}, \binits{M.}},
\oauthor{\bsnm{Thacker}, \binits{P.}}, et al.:
Improving alignment of dialogue agents via targeted human judgements.
arXiv preprint arXiv:2209.14375
(2022)
\end{botherref}
\endbibitem

\bibitem[\protect\citeauthoryear{Hoffmann et~al.}{2022}]{hoffmann2022training}
\begin{bchapter}
\bauthor{\bsnm{Hoffmann}, \binits{J.}},
\bauthor{\bsnm{Borgeaud}, \binits{S.}},
\bauthor{\bsnm{Mensch}, \binits{A.}},
\bauthor{\bsnm{Buchatskaya}, \binits{E.}},
\bauthor{\bsnm{Cai}, \binits{T.}},
\bauthor{\bsnm{Rutherford}, \binits{E.}},
\bauthor{\bsnm{Las~Casas}, \binits{D.}},
\bauthor{\bsnm{Hendricks}, \binits{L.A.}},
\bauthor{\bsnm{Welbl}, \binits{J.}},
\bauthor{\bsnm{Clark}, \binits{A.}}, \betal:
\bctitle{Training compute-optimal large language models}.
In: \bbtitle{Proceedings of the 36th International Conference on Neural Information Processing Systems},
pp. \bfpage{30016}--\blpage{30030}
(\byear{2022})
\end{bchapter}
\endbibitem

\bibitem[\protect\citeauthoryear{Iyer et~al.}{2022}]{iyer2022opt}
\begin{botherref}
\oauthor{\bsnm{Iyer}, \binits{S.}},
\oauthor{\bsnm{Lin}, \binits{X.V.}},
\oauthor{\bsnm{Pasunuru}, \binits{R.}},
\oauthor{\bsnm{Mihaylov}, \binits{T.}},
\oauthor{\bsnm{Simig}, \binits{D.}},
\oauthor{\bsnm{Yu}, \binits{P.}},
\oauthor{\bsnm{Shuster}, \binits{K.}},
\oauthor{\bsnm{Wang}, \binits{T.}},
\oauthor{\bsnm{Liu}, \binits{Q.}},
\oauthor{\bsnm{Koura}, \binits{P.S.}}, et al.:
Opt-iml: Scaling language model instruction meta learning through the lens of generalization.
arXiv preprint arXiv:2212.12017
(2022)
\end{botherref}
\endbibitem

\bibitem[\protect\citeauthoryear{Rae et~al.}{2021}]{rae2021scaling}
\begin{botherref}
\oauthor{\bsnm{Rae}, \binits{J.W.}},
\oauthor{\bsnm{Borgeaud}, \binits{S.}},
\oauthor{\bsnm{Cai}, \binits{T.}},
\oauthor{\bsnm{Millican}, \binits{K.}},
\oauthor{\bsnm{Hoffmann}, \binits{J.}},
\oauthor{\bsnm{Song}, \binits{F.}},
\oauthor{\bsnm{Aslanides}, \binits{J.}},
\oauthor{\bsnm{Henderson}, \binits{S.}},
\oauthor{\bsnm{Ring}, \binits{R.}},
\oauthor{\bsnm{Young}, \binits{S.}}, et al.:
Scaling language models: Methods, analysis \& insights from training gopher.
arXiv preprint arXiv:2112.11446
(2021)
\end{botherref}
\endbibitem

\bibitem[\protect\citeauthoryear{Zheng et~al.}{2023}]{zheng2023codegeex}
\begin{botherref}
\oauthor{\bsnm{Zheng}, \binits{Q.}},
\oauthor{\bsnm{Xia}, \binits{X.}},
\oauthor{\bsnm{Zou}, \binits{X.}},
\oauthor{\bsnm{Dong}, \binits{Y.}},
\oauthor{\bsnm{Wang}, \binits{S.}},
\oauthor{\bsnm{Xue}, \binits{Y.}},
\oauthor{\bsnm{Wang}, \binits{Z.}},
\oauthor{\bsnm{Shen}, \binits{L.}},
\oauthor{\bsnm{Wang}, \binits{A.}},
\oauthor{\bsnm{Li}, \binits{Y.}}, et al.:
Codegeex: A pre-trained model for code generation with multilingual evaluations on humaneval-x.
arXiv preprint arXiv:2303.17568
(2023)
\end{botherref}
\endbibitem

\bibitem[\protect\citeauthoryear{Wei et~al.}{2023}]{wei2023skywork}
\begin{botherref}
\oauthor{\bsnm{Wei}, \binits{T.}},
\oauthor{\bsnm{Zhao}, \binits{L.}},
\oauthor{\bsnm{Zhang}, \binits{L.}},
\oauthor{\bsnm{Zhu}, \binits{B.}},
\oauthor{\bsnm{Wang}, \binits{L.}},
\oauthor{\bsnm{Yang}, \binits{H.}},
\oauthor{\bsnm{Li}, \binits{B.}},
\oauthor{\bsnm{Cheng}, \binits{C.}},
\oauthor{\bsnm{L{\"u}}, \binits{W.}},
\oauthor{\bsnm{Hu}, \binits{R.}}, et al.:
Skywork: A more open bilingual foundation model.
arXiv preprint arXiv:2310.19341
(2023)
\end{botherref}
\endbibitem

\bibitem[\protect\citeauthoryear{Bai et~al.}{2023}]{bai2023qwen}
\begin{botherref}
\oauthor{\bsnm{Bai}, \binits{J.}},
\oauthor{\bsnm{Bai}, \binits{S.}},
\oauthor{\bsnm{Chu}, \binits{Y.}},
\oauthor{\bsnm{Cui}, \binits{Z.}},
\oauthor{\bsnm{Dang}, \binits{K.}},
\oauthor{\bsnm{Deng}, \binits{X.}},
\oauthor{\bsnm{Fan}, \binits{Y.}},
\oauthor{\bsnm{Ge}, \binits{W.}},
\oauthor{\bsnm{Han}, \binits{Y.}},
\oauthor{\bsnm{Huang}, \binits{F.}}, et al.:
Qwen technical report.
arXiv preprint arXiv:2309.16609
(2023)
\end{botherref}
\endbibitem

\bibitem[\protect\citeauthoryear{Nijkamp et~al.}{2023}]{nijkamp2023codegen2}
\begin{botherref}
\oauthor{\bsnm{Nijkamp}, \binits{E.}},
\oauthor{\bsnm{Hayashi}, \binits{H.}},
\oauthor{\bsnm{Xiong}, \binits{C.}},
\oauthor{\bsnm{Savarese}, \binits{S.}},
\oauthor{\bsnm{Zhou}, \binits{Y.}}:
Codegen2: Lessons for training llms on programming and natural languages.
arXiv preprint arXiv:2305.02309
(2023)
\end{botherref}
\endbibitem

\bibitem[\protect\citeauthoryear{Black et~al.}{}]{black2022gpt}
\begin{botherref}
\oauthor{\bsnm{Black}, \binits{S.}},
\oauthor{\bsnm{Biderman}, \binits{S.}},
\oauthor{\bsnm{Hallahan}, \binits{E.}},
\oauthor{\bsnm{Anthony}, \binits{Q.G.}},
\oauthor{\bsnm{Gao}, \binits{L.}},
\oauthor{\bsnm{Golding}, \binits{L.}},
\oauthor{\bsnm{He}, \binits{H.}},
\oauthor{\bsnm{Leahy}, \binits{C.}},
\oauthor{\bsnm{McDonell}, \binits{K.}},
\oauthor{\bsnm{Phang}, \binits{J.}}, et al.:
Gpt-neox-20b: An open-source autoregressive language model.
In: Challenges $\{$$\backslash$\&$\}$ Perspectives in Creating Large Language Models
\end{botherref}
\endbibitem

\bibitem[\protect\citeauthoryear{Reid et~al.}{2024}]{reid2024gemini}
\begin{botherref}
\oauthor{\bsnm{Reid}, \binits{M.}},
\oauthor{\bsnm{Savinov}, \binits{N.}},
\oauthor{\bsnm{Teplyashin}, \binits{D.}},
\oauthor{\bsnm{Lepikhin}, \binits{D.}},
\oauthor{\bsnm{Lillicrap}, \binits{T.}},
\oauthor{\bsnm{Alayrac}, \binits{J.-b.}},
\oauthor{\bsnm{Soricut}, \binits{R.}},
\oauthor{\bsnm{Lazaridou}, \binits{A.}},
\oauthor{\bsnm{Firat}, \binits{O.}},
\oauthor{\bsnm{Schrittwieser}, \binits{J.}}, et al.:
Gemini 1.5: Unlocking multimodal understanding across millions of tokens of context.
arXiv preprint arXiv:2403.05530
(2024)
\end{botherref}
\endbibitem

\bibitem[\protect\citeauthoryear{Dubey et~al.}{2024}]{dubey2024llama}
\begin{botherref}
\oauthor{\bsnm{Dubey}, \binits{A.}},
\oauthor{\bsnm{Jauhri}, \binits{A.}},
\oauthor{\bsnm{Pandey}, \binits{A.}},
\oauthor{\bsnm{Kadian}, \binits{A.}},
\oauthor{\bsnm{Al-Dahle}, \binits{A.}},
\oauthor{\bsnm{Letman}, \binits{A.}},
\oauthor{\bsnm{Mathur}, \binits{A.}},
\oauthor{\bsnm{Schelten}, \binits{A.}},
\oauthor{\bsnm{Yang}, \binits{A.}},
\oauthor{\bsnm{Fan}, \binits{A.}}, et al.:
The llama 3 herd of models.
arXiv preprint arXiv:2407.21783
(2024)
\end{botherref}
\endbibitem

\bibitem[\protect\citeauthoryear{Taylor et~al.}{2022}]{taylor2022galactica}
\begin{botherref}
\oauthor{\bsnm{Taylor}, \binits{R.}},
\oauthor{\bsnm{Kardas}, \binits{M.}},
\oauthor{\bsnm{Cucurull}, \binits{G.}},
\oauthor{\bsnm{Scialom}, \binits{T.}},
\oauthor{\bsnm{Hartshorn}, \binits{A.}},
\oauthor{\bsnm{Saravia}, \binits{E.}},
\oauthor{\bsnm{Poulton}, \binits{A.}},
\oauthor{\bsnm{Kerkez}, \binits{V.}},
\oauthor{\bsnm{Stojnic}, \binits{R.}}:
Galactica: A large language model for science.
arXiv preprint arXiv:2211.09085
(2022)
\end{botherref}
\endbibitem

\bibitem[\protect\citeauthoryear{Thoppilan et~al.}{2022}]{thoppilan2022lamda}
\begin{botherref}
\oauthor{\bsnm{Thoppilan}, \binits{R.}},
\oauthor{\bsnm{De~Freitas}, \binits{D.}},
\oauthor{\bsnm{Hall}, \binits{J.}},
\oauthor{\bsnm{Shazeer}, \binits{N.}},
\oauthor{\bsnm{Kulshreshtha}, \binits{A.}},
\oauthor{\bsnm{Cheng}, \binits{H.-T.}},
\oauthor{\bsnm{Jin}, \binits{A.}},
\oauthor{\bsnm{Bos}, \binits{T.}},
\oauthor{\bsnm{Baker}, \binits{L.}},
\oauthor{\bsnm{Du}, \binits{Y.}}, et al.:
Lamda: Language models for dialog applications.
arXiv preprint arXiv:2201.08239
(2022)
\end{botherref}
\endbibitem

\bibitem[\protect\citeauthoryear{Lieber et~al.}{2021}]{lieber2021jurassic}
\begin{botherref}
\oauthor{\bsnm{Lieber}, \binits{O.}},
\oauthor{\bsnm{Sharir}, \binits{O.}},
\oauthor{\bsnm{Lenz}, \binits{B.}},
\oauthor{\bsnm{Shoham}, \binits{Y.}}:
Jurassic-1: Technical details and evaluation.
White Paper. AI21 Labs
\textbf{1}(9)
(2021)
\end{botherref}
\endbibitem

\bibitem[\protect\citeauthoryear{Nakano et~al.}{2021}]{nakano2021webgpt}
\begin{botherref}
\oauthor{\bsnm{Nakano}, \binits{R.}},
\oauthor{\bsnm{Hilton}, \binits{J.}},
\oauthor{\bsnm{Balaji}, \binits{S.}},
\oauthor{\bsnm{Wu}, \binits{J.}},
\oauthor{\bsnm{Ouyang}, \binits{L.}},
\oauthor{\bsnm{Kim}, \binits{C.}},
\oauthor{\bsnm{Hesse}, \binits{C.}},
\oauthor{\bsnm{Jain}, \binits{S.}},
\oauthor{\bsnm{Kosaraju}, \binits{V.}},
\oauthor{\bsnm{Saunders}, \binits{W.}}, et al.:
Webgpt: Browser-assisted question-answering with human feedback.
arXiv preprint arXiv:2112.09332
(2021)
\end{botherref}
\endbibitem

\bibitem[\protect\citeauthoryear{Ouyang et~al.}{2022}]{ouyang2022training}
\begin{barticle}
\bauthor{\bsnm{Ouyang}, \binits{L.}},
\bauthor{\bsnm{Wu}, \binits{J.}},
\bauthor{\bsnm{Jiang}, \binits{X.}},
\bauthor{\bsnm{Almeida}, \binits{D.}},
\bauthor{\bsnm{Wainwright}, \binits{C.}},
\bauthor{\bsnm{Mishkin}, \binits{P.}},
\bauthor{\bsnm{Zhang}, \binits{C.}},
\bauthor{\bsnm{Agarwal}, \binits{S.}},
\bauthor{\bsnm{Slama}, \binits{K.}},
\bauthor{\bsnm{Ray}, \binits{A.}}, \betal:
\batitle{Training language models to follow instructions with human feedback}.
\bjtitle{Advances in neural information processing systems}
\bvolume{35},
\bfpage{27730}--\blpage{27744}
(\byear{2022})
\end{barticle}
\endbibitem

\bibitem[\protect\citeauthoryear{Le~Scao et~al.}{2023}]{le2023bloom}
\begin{botherref}
\oauthor{\bsnm{Le~Scao}, \binits{T.}},
\oauthor{\bsnm{Fan}, \binits{A.}},
\oauthor{\bsnm{Akiki}, \binits{C.}},
\oauthor{\bsnm{Pavlick}, \binits{E.}},
\oauthor{\bsnm{Ili{\'c}}, \binits{S.}},
\oauthor{\bsnm{Hesslow}, \binits{D.}},
\oauthor{\bsnm{Castagn{\'e}}, \binits{R.}},
\oauthor{\bsnm{Luccioni}, \binits{A.S.}},
\oauthor{\bsnm{Yvon}, \binits{F.}},
\oauthor{\bsnm{Gall{\'e}}, \binits{M.}}, et al.:
Bloom: A 176b-parameter open-access multilingual language model
(2023)
\end{botherref}
\endbibitem

\bibitem[\protect\citeauthoryear{Vavekanand and Sam}{}]{vavekanandllama}
\begin{botherref}
\oauthor{\bsnm{Vavekanand}, \binits{R.}},
\oauthor{\bsnm{Sam}, \binits{K.}}:
Llama 3.1: An in-depth analysis of the next-generation large language model
\end{botherref}
\endbibitem

\bibitem[\protect\citeauthoryear{Smith et~al.}{2022}]{smith2022using}
\begin{botherref}
\oauthor{\bsnm{Smith}, \binits{S.}},
\oauthor{\bsnm{Patwary}, \binits{M.}},
\oauthor{\bsnm{Norick}, \binits{B.}},
\oauthor{\bsnm{LeGresley}, \binits{P.}},
\oauthor{\bsnm{Rajbhandari}, \binits{S.}},
\oauthor{\bsnm{Casper}, \binits{J.}},
\oauthor{\bsnm{Liu}, \binits{Z.}},
\oauthor{\bsnm{Prabhumoye}, \binits{S.}},
\oauthor{\bsnm{Zerveas}, \binits{G.}},
\oauthor{\bsnm{Korthikanti}, \binits{V.}}, et al.:
Using deepspeed and megatron to train megatron-turing nlg 530b, a large-scale generative language model.
arXiv preprint arXiv:2201.11990
(2022)
\end{botherref}
\endbibitem

\bibitem[\protect\citeauthoryear{Team et~al.}{2024}]{team2024gemma}
\begin{botherref}
\oauthor{\bsnm{Team}, \binits{G.}},
\oauthor{\bsnm{Mesnard}, \binits{T.}},
\oauthor{\bsnm{Hardin}, \binits{C.}},
\oauthor{\bsnm{Dadashi}, \binits{R.}},
\oauthor{\bsnm{Bhupatiraju}, \binits{S.}},
\oauthor{\bsnm{Pathak}, \binits{S.}},
\oauthor{\bsnm{Sifre}, \binits{L.}},
\oauthor{\bsnm{Rivi{\`e}re}, \binits{M.}},
\oauthor{\bsnm{Kale}, \binits{M.S.}},
\oauthor{\bsnm{Love}, \binits{J.}}, et al.:
Gemma: Open models based on gemini research and technology.
arXiv preprint arXiv:2403.08295
(2024)
\end{botherref}
\endbibitem

\bibitem[\protect\citeauthoryear{Li et~al.}{2022}]{li2022automating}
\begin{bchapter}
\bauthor{\bsnm{Li}, \binits{Z.}},
\bauthor{\bsnm{Lu}, \binits{S.}},
\bauthor{\bsnm{Guo}, \binits{D.}},
\bauthor{\bsnm{Duan}, \binits{N.}},
\bauthor{\bsnm{Jannu}, \binits{S.}},
\bauthor{\bsnm{Jenks}, \binits{G.}},
\bauthor{\bsnm{Majumder}, \binits{D.}},
\bauthor{\bsnm{Green}, \binits{J.}},
\bauthor{\bsnm{Svyatkovskiy}, \binits{A.}},
\bauthor{\bsnm{Fu}, \binits{S.}}, \betal:
\bctitle{Automating code review activities by large-scale pre-training}.
In: \bbtitle{Proceedings of the 30th ACM Joint European Software Engineering Conference and Symposium on the Foundations of Software Engineering},
pp. \bfpage{1035}--\blpage{1047}
(\byear{2022})
\end{bchapter}
\endbibitem

\bibitem[\protect\citeauthoryear{He et~al.}{2024}]{he2024representation}
\begin{barticle}
\bauthor{\bsnm{He}, \binits{J.}},
\bauthor{\bsnm{Zhou}, \binits{X.}},
\bauthor{\bsnm{Xu}, \binits{B.}},
\bauthor{\bsnm{Zhang}, \binits{T.}},
\bauthor{\bsnm{Kim}, \binits{K.}},
\bauthor{\bsnm{Yang}, \binits{Z.}},
\bauthor{\bsnm{Thung}, \binits{F.}},
\bauthor{\bsnm{Irsan}, \binits{I.C.}},
\bauthor{\bsnm{Lo}, \binits{D.}}:
\batitle{Representation learning for stack overflow posts: How far are we?}
\bjtitle{ACM Transactions on Software Engineering and Methodology}
\bvolume{33}(\bissue{3}),
\bfpage{1}--\blpage{24}
(\byear{2024})
\end{barticle}
\endbibitem

\bibitem[\protect\citeauthoryear{He et~al.}{2022}]{he2022ptm4tag}
\begin{bchapter}
\bauthor{\bsnm{He}, \binits{J.}},
\bauthor{\bsnm{Xu}, \binits{B.}},
\bauthor{\bsnm{Yang}, \binits{Z.}},
\bauthor{\bsnm{Han}, \binits{D.}},
\bauthor{\bsnm{Yang}, \binits{C.}},
\bauthor{\bsnm{Lo}, \binits{D.}}:
\bctitle{Ptm4tag: sharpening tag recommendation of stack overflow posts with pre-trained models}.
In: \bbtitle{Proceedings of the 30th IEEE/ACM International Conference on Program Comprehension},
pp. \bfpage{1}--\blpage{11}
(\byear{2022})
\end{bchapter}
\endbibitem

\bibitem[\protect\citeauthoryear{Yang et~al.}{2022}]{yang2022answer}
\begin{bchapter}
\bauthor{\bsnm{Yang}, \binits{C.}},
\bauthor{\bsnm{Xu}, \binits{B.}},
\bauthor{\bsnm{Thung}, \binits{F.}},
\bauthor{\bsnm{Shi}, \binits{Y.}},
\bauthor{\bsnm{Zhang}, \binits{T.}},
\bauthor{\bsnm{Yang}, \binits{Z.}},
\bauthor{\bsnm{Zhou}, \binits{X.}},
\bauthor{\bsnm{Shi}, \binits{J.}},
\bauthor{\bsnm{He}, \binits{J.}},
\bauthor{\bsnm{Han}, \binits{D.}}, \betal:
\bctitle{Answer summarization for technical queries: Benchmark and new approach}.
In: \bbtitle{Proceedings of the 37th IEEE/ACM International Conference on Automated Software Engineering},
pp. \bfpage{1}--\blpage{13}
(\byear{2022})
\end{bchapter}
\endbibitem

\bibitem[\protect\citeauthoryear{Roziere et~al.}{2023}]{roziere2023code}
\begin{botherref}
\oauthor{\bsnm{Roziere}, \binits{B.}},
\oauthor{\bsnm{Gehring}, \binits{J.}},
\oauthor{\bsnm{Gloeckle}, \binits{F.}},
\oauthor{\bsnm{Sootla}, \binits{S.}},
\oauthor{\bsnm{Gat}, \binits{I.}},
\oauthor{\bsnm{Tan}, \binits{X.E.}},
\oauthor{\bsnm{Adi}, \binits{Y.}},
\oauthor{\bsnm{Liu}, \binits{J.}},
\oauthor{\bsnm{Remez}, \binits{T.}},
\oauthor{\bsnm{Rapin}, \binits{J.}}, et al.:
Code llama: Open foundation models for code.
arXiv preprint arXiv:2308.12950
(2023)
\end{botherref}
\endbibitem

\bibitem[\protect\citeauthoryear{Le et~al.}{2022}]{le2022coderl}
\begin{barticle}
\bauthor{\bsnm{Le}, \binits{H.}},
\bauthor{\bsnm{Wang}, \binits{Y.}},
\bauthor{\bsnm{Gotmare}, \binits{A.D.}},
\bauthor{\bsnm{Savarese}, \binits{S.}},
\bauthor{\bsnm{Hoi}, \binits{S.C.H.}}:
\batitle{Coderl: Mastering code generation through pretrained models and deep reinforcement learning}.
\bjtitle{Advances in Neural Information Processing Systems}
\bvolume{35},
\bfpage{21314}--\blpage{21328}
(\byear{2022})
\end{barticle}
\endbibitem

\bibitem[\protect\citeauthoryear{Sadybekov and Katritch}{2023}]{sadybekov2023computational}
\begin{barticle}
\bauthor{\bsnm{Sadybekov}, \binits{A.V.}},
\bauthor{\bsnm{Katritch}, \binits{V.}}:
\batitle{Computational approaches streamlining drug discovery}.
\bjtitle{Nature}
\bvolume{616}(\bissue{7958}),
\bfpage{673}--\blpage{685}
(\byear{2023})
\end{barticle}
\endbibitem

\bibitem[\protect\citeauthoryear{Gorgulla et~al.}{2022}]{gorgulla2022emerging}
\begin{barticle}
\bauthor{\bsnm{Gorgulla}, \binits{C.}},
\bauthor{\bsnm{Jayaraj}, \binits{A.}},
\bauthor{\bsnm{Fackeldey}, \binits{K.}},
\bauthor{\bsnm{Arthanari}, \binits{H.}}:
\batitle{Emerging frontiers in virtual drug discovery: From quantum mechanical methods to deep learning approaches}.
\bjtitle{Current opinion in chemical biology}
\bvolume{69},
\bfpage{102156}
(\byear{2022})
\end{barticle}
\endbibitem

\bibitem[\protect\citeauthoryear{Savage}{2023}]{savage2023drug}
\begin{barticle}
\bauthor{\bsnm{Savage}, \binits{N.}}:
\batitle{Drug discovery companies are customizing chatgpt: here’s how}.
\bjtitle{Nat Biotechnol}
\bvolume{41}(\bissue{5}),
\bfpage{585}--\blpage{586}
(\byear{2023})
\end{barticle}
\endbibitem

\bibitem[\protect\citeauthoryear{Haley and Roudnicky}{2020}]{haley2020functional}
\begin{barticle}
\bauthor{\bsnm{Haley}, \binits{B.}},
\bauthor{\bsnm{Roudnicky}, \binits{F.}}:
\batitle{Functional genomics for cancer drug target discovery}.
\bjtitle{Cancer Cell}
\bvolume{38}(\bissue{1}),
\bfpage{31}--\blpage{43}
(\byear{2020})
\end{barticle}
\endbibitem

\bibitem[\protect\citeauthoryear{Paananen and Fortino}{2020}]{paananen2020omics}
\begin{barticle}
\bauthor{\bsnm{Paananen}, \binits{J.}},
\bauthor{\bsnm{Fortino}, \binits{V.}}:
\batitle{An omics perspective on drug target discovery platforms}.
\bjtitle{Briefings in bioinformatics}
\bvolume{21}(\bissue{6}),
\bfpage{1937}--\blpage{1953}
(\byear{2020})
\end{barticle}
\endbibitem

\bibitem[\protect\citeauthoryear{Zhang et~al.}{2020}]{zhang2020deep}
\begin{botherref}
\oauthor{\bsnm{Zhang}, \binits{Z.}},
\oauthor{\bsnm{Zohren}, \binits{S.}},
\oauthor{\bsnm{Roberts}, \binits{S.}}:
Deep learning for portfolio optimization.
The Journal of Financial Data Science
(2020)
\end{botherref}
\endbibitem

\bibitem[\protect\citeauthoryear{Mashrur et~al.}{2020}]{mashrur2020machine}
\begin{barticle}
\bauthor{\bsnm{Mashrur}, \binits{A.}},
\bauthor{\bsnm{Luo}, \binits{W.}},
\bauthor{\bsnm{Zaidi}, \binits{N.A.}},
\bauthor{\bsnm{Robles-Kelly}, \binits{A.}}:
\batitle{Machine learning for financial risk management: a survey}.
\bjtitle{Ieee Access}
\bvolume{8},
\bfpage{203203}--\blpage{203223}
(\byear{2020})
\end{barticle}
\endbibitem

\bibitem[\protect\citeauthoryear{Shah et~al.}{}]{shahfinaid}
\begin{botherref}
\oauthor{\bsnm{Shah}, \binits{A.}},
\oauthor{\bsnm{Raj}, \binits{P.}},
\oauthor{\bsnm{Pushpam~Kumar}, \binits{S.P.}},
\oauthor{\bsnm{Asha}, \binits{H.}}:
Finaid, a financial advisor application using ai
\end{botherref}
\endbibitem

\bibitem[\protect\citeauthoryear{Misischia et~al.}{2022}]{misischia2022chatbots}
\begin{barticle}
\bauthor{\bsnm{Misischia}, \binits{C.V.}},
\bauthor{\bsnm{Poecze}, \binits{F.}},
\bauthor{\bsnm{Strauss}, \binits{C.}}:
\batitle{Chatbots in customer service: Their relevance and impact on service quality}.
\bjtitle{Procedia Computer Science}
\bvolume{201},
\bfpage{421}--\blpage{428}
(\byear{2022})
\end{barticle}
\endbibitem

\bibitem[\protect\citeauthoryear{Wu et~al.}{2023}]{wu2023bloomberggpt}
\begin{botherref}
\oauthor{\bsnm{Wu}, \binits{S.}},
\oauthor{\bsnm{Irsoy}, \binits{O.}},
\oauthor{\bsnm{Lu}, \binits{S.}},
\oauthor{\bsnm{Dabravolski}, \binits{V.}},
\oauthor{\bsnm{Dredze}, \binits{M.}},
\oauthor{\bsnm{Gehrmann}, \binits{S.}},
\oauthor{\bsnm{Kambadur}, \binits{P.}},
\oauthor{\bsnm{Rosenberg}, \binits{D.}},
\oauthor{\bsnm{Mann}, \binits{G.}}:
Bloomberggpt: A large language model for finance.
arXiv preprint arXiv:2303.17564
(2023)
\end{botherref}
\endbibitem

\bibitem[\protect\citeauthoryear{Thirunavukarasu et~al.}{2023}]{thirunavukarasu2023large}
\begin{barticle}
\bauthor{\bsnm{Thirunavukarasu}, \binits{A.J.}},
\bauthor{\bsnm{Ting}, \binits{D.S.J.}},
\bauthor{\bsnm{Elangovan}, \binits{K.}},
\bauthor{\bsnm{Gutierrez}, \binits{L.}},
\bauthor{\bsnm{Tan}, \binits{T.F.}},
\bauthor{\bsnm{Ting}, \binits{D.S.W.}}:
\batitle{Large language models in medicine}.
\bjtitle{Nature medicine}
\bvolume{29}(\bissue{8}),
\bfpage{1930}--\blpage{1940}
(\byear{2023})
\end{barticle}
\endbibitem

\bibitem[\protect\citeauthoryear{Chinta et~al.}{2024}]{chinta2024ai}
\begin{botherref}
\oauthor{\bsnm{Chinta}, \binits{S.V.}},
\oauthor{\bsnm{Wang}, \binits{Z.}},
\oauthor{\bsnm{Zhang}, \binits{X.}},
\oauthor{\bsnm{Viet}, \binits{T.D.}},
\oauthor{\bsnm{Kashif}, \binits{A.}},
\oauthor{\bsnm{Smith}, \binits{M.A.}},
\oauthor{\bsnm{Zhang}, \binits{W.}}:
Ai-driven healthcare: A survey on ensuring fairness and mitigating bias.
arXiv preprint arXiv:2407.19655
(2024)
\end{botherref}
\endbibitem

\bibitem[\protect\citeauthoryear{Singhal et~al.}{2023}]{singhal2023towards}
\begin{botherref}
\oauthor{\bsnm{Singhal}, \binits{K.}},
\oauthor{\bsnm{Tu}, \binits{T.}},
\oauthor{\bsnm{Gottweis}, \binits{J.}},
\oauthor{\bsnm{Sayres}, \binits{R.}},
\oauthor{\bsnm{Wulczyn}, \binits{E.}},
\oauthor{\bsnm{Hou}, \binits{L.}},
\oauthor{\bsnm{Clark}, \binits{K.}},
\oauthor{\bsnm{Pfohl}, \binits{S.}},
\oauthor{\bsnm{Cole-Lewis}, \binits{H.}},
\oauthor{\bsnm{Neal}, \binits{D.}}, et al.:
Towards expert-level medical question answering with large language models.
arXiv preprint arXiv:2305.09617
(2023)
\end{botherref}
\endbibitem

\bibitem[\protect\citeauthoryear{Arora and Arora}{2023}]{arora2023promise}
\begin{barticle}
\bauthor{\bsnm{Arora}, \binits{A.}},
\bauthor{\bsnm{Arora}, \binits{A.}}:
\batitle{The promise of large language models in health care}.
\bjtitle{The Lancet}
\bvolume{401}(\bissue{10377}),
\bfpage{641}
(\byear{2023})
\end{barticle}
\endbibitem

\bibitem[\protect\citeauthoryear{Bommarito~II and Katz}{2022}]{bommarito2022gpt}
\begin{botherref}
\oauthor{\bsnm{Bommarito~II}, \binits{M.}},
\oauthor{\bsnm{Katz}, \binits{D.M.}}:
Gpt takes the bar exam.
arXiv preprint arXiv:2212.14402
(2022)
\end{botherref}
\endbibitem

\bibitem[\protect\citeauthoryear{Iu and Wong}{2023}]{iu2023chatgpt}
\begin{botherref}
\oauthor{\bsnm{Iu}, \binits{K.Y.}},
\oauthor{\bsnm{Wong}, \binits{V.M.-Y.}}:
Chatgpt by openai: The end of litigation lawyers?
Available at SSRN 4339839
(2023)
\end{botherref}
\endbibitem

\bibitem[\protect\citeauthoryear{Lee et~al.}{}]{lee23generative}
\begin{botherref}
\oauthor{\bsnm{Lee}, \binits{U.}},
\oauthor{\bsnm{Lee}, \binits{S.}},
\oauthor{\bsnm{Koh}, \binits{J.}},
\oauthor{\bsnm{Jeong}, \binits{Y.}},
\oauthor{\bsnm{Jung}, \binits{H.}},
\oauthor{\bsnm{Byun}, \binits{G.}},
\oauthor{\bsnm{Lee}, \binits{Y.}},
\oauthor{\bsnm{Moon}, \binits{J.}},
\oauthor{\bsnm{Lim}, \binits{J.}},
\oauthor{\bsnm{Kim}, \binits{H.}}:
Generative Agent for Teacher Training: Designing Educational Problem-Solving Simulations with Large Language Model-based Agents for Pre-Service Teachers.
NeurIPS
\end{botherref}
\endbibitem

\bibitem[\protect\citeauthoryear{Markel et~al.}{2023}]{markel2023gpteach}
\begin{bchapter}
\bauthor{\bsnm{Markel}, \binits{J.M.}},
\bauthor{\bsnm{Opferman}, \binits{S.G.}},
\bauthor{\bsnm{Landay}, \binits{J.A.}},
\bauthor{\bsnm{Piech}, \binits{C.}}:
\bctitle{Gpteach: Interactive ta training with gpt-based students}.
In: \bbtitle{Proceedings of the Tenth Acm Conference on Learning@ Scale},
pp. \bfpage{226}--\blpage{236}
(\byear{2023})
\end{bchapter}
\endbibitem

\bibitem[\protect\citeauthoryear{Chinta et~al.}{2024}]{chinta2024fairaied}
\begin{botherref}
\oauthor{\bsnm{Chinta}, \binits{S.V.}},
\oauthor{\bsnm{Wang}, \binits{Z.}},
\oauthor{\bsnm{Yin}, \binits{Z.}},
\oauthor{\bsnm{Hoang}, \binits{N.}},
\oauthor{\bsnm{Gonzalez}, \binits{M.}},
\oauthor{\bsnm{Quy}, \binits{T.L.}},
\oauthor{\bsnm{Zhang}, \binits{W.}}:
Fairaied: Navigating fairness, bias, and ethics in educational ai applications.
arXiv preprint arXiv:2407.18745
(2024)
\end{botherref}
\endbibitem

\bibitem[\protect\citeauthoryear{Tu et~al.}{2023}]{tu2023littlemu}
\begin{bchapter}
\bauthor{\bsnm{Tu}, \binits{S.}},
\bauthor{\bsnm{Zhang}, \binits{Z.}},
\bauthor{\bsnm{Yu}, \binits{J.}},
\bauthor{\bsnm{Li}, \binits{C.}},
\bauthor{\bsnm{Zhang}, \binits{S.}},
\bauthor{\bsnm{Yao}, \binits{Z.}},
\bauthor{\bsnm{Hou}, \binits{L.}},
\bauthor{\bsnm{Li}, \binits{J.}}:
\bctitle{Littlemu: Deploying an online virtual teaching assistant via heterogeneous sources integration and chain of teach prompts}.
In: \bbtitle{Proceedings of the 32nd ACM International Conference on Information and Knowledge Management},
pp. \bfpage{4843}--\blpage{4849}
(\byear{2023})
\end{bchapter}
\endbibitem

\bibitem[\protect\citeauthoryear{Chen et~al.}{2023}]{chen2023empowering}
\begin{botherref}
\oauthor{\bsnm{Chen}, \binits{Y.}},
\oauthor{\bsnm{Ding}, \binits{N.}},
\oauthor{\bsnm{Zheng}, \binits{H.-T.}},
\oauthor{\bsnm{Liu}, \binits{Z.}},
\oauthor{\bsnm{Sun}, \binits{M.}},
\oauthor{\bsnm{Zhou}, \binits{B.}}:
Empowering private tutoring by chaining large language models.
arXiv preprint arXiv:2309.08112
(2023)
\end{botherref}
\endbibitem

\bibitem[\protect\citeauthoryear{Zentner}{2022}]{zentner2022applied}
\begin{botherref}
\oauthor{\bsnm{Zentner}, \binits{A.}}:
Applied innovation: Artificial intelligence in higher education.
Available at SSRN 4314180
(2022)
\end{botherref}
\endbibitem

\bibitem[\protect\citeauthoryear{Zhang}{2023}]{zhang2023preparing}
\begin{botherref}
\oauthor{\bsnm{Zhang}, \binits{B.}}:
Preparing educators and students for chatgpt and ai technology in higher education.
ResearchGate
(2023)
\end{botherref}
\endbibitem

\bibitem[\protect\citeauthoryear{Dwivedi et~al.}{2023}]{dwivedi2023opinion}
\begin{barticle}
\bauthor{\bsnm{Dwivedi}, \binits{Y.K.}},
\bauthor{\bsnm{Kshetri}, \binits{N.}},
\bauthor{\bsnm{Hughes}, \binits{L.}},
\bauthor{\bsnm{Slade}, \binits{E.L.}},
\bauthor{\bsnm{Jeyaraj}, \binits{A.}},
\bauthor{\bsnm{Kar}, \binits{A.K.}},
\bauthor{\bsnm{Baabdullah}, \binits{A.M.}},
\bauthor{\bsnm{Koohang}, \binits{A.}},
\bauthor{\bsnm{Raghavan}, \binits{V.}},
\bauthor{\bsnm{Ahuja}, \binits{M.}}, \betal:
\batitle{Opinion paper:“so what if chatgpt wrote it?” multidisciplinary perspectives on opportunities, challenges and implications of generative conversational ai for research, practice and policy}.
\bjtitle{International Journal of Information Management}
\bvolume{71},
\bfpage{102642}
(\byear{2023})
\end{barticle}
\endbibitem

\bibitem[\protect\citeauthoryear{Chen et~al.}{2023}]{chen2023artificial}
\begin{barticle}
\bauthor{\bsnm{Chen}, \binits{Y.}},
\bauthor{\bsnm{Jensen}, \binits{S.}},
\bauthor{\bsnm{Albert}, \binits{L.J.}},
\bauthor{\bsnm{Gupta}, \binits{S.}},
\bauthor{\bsnm{Lee}, \binits{T.}}:
\batitle{Artificial intelligence (ai) student assistants in the classroom: Designing chatbots to support student success}.
\bjtitle{Information Systems Frontiers}
\bvolume{25}(\bissue{1}),
\bfpage{161}--\blpage{182}
(\byear{2023})
\end{barticle}
\endbibitem

\bibitem[\protect\citeauthoryear{Yan et~al.}{2024}]{yan2024protecting}
\begin{botherref}
\oauthor{\bsnm{Yan}, \binits{B.}},
\oauthor{\bsnm{Li}, \binits{K.}},
\oauthor{\bsnm{Xu}, \binits{M.}},
\oauthor{\bsnm{Dong}, \binits{Y.}},
\oauthor{\bsnm{Zhang}, \binits{Y.}},
\oauthor{\bsnm{Ren}, \binits{Z.}},
\oauthor{\bsnm{Cheng}, \binits{X.}}:
On protecting the data privacy of large language models (llms): A survey.
arXiv preprint arXiv:2403.05156
(2024)
\end{botherref}
\endbibitem

\bibitem[\protect\citeauthoryear{Carlini et~al.}{2021}]{carlini2021extracting}
\begin{bchapter}
\bauthor{\bsnm{Carlini}, \binits{N.}},
\bauthor{\bsnm{Tramer}, \binits{F.}},
\bauthor{\bsnm{Wallace}, \binits{E.}},
\bauthor{\bsnm{Jagielski}, \binits{M.}},
\bauthor{\bsnm{Herbert-Voss}, \binits{A.}},
\bauthor{\bsnm{Lee}, \binits{K.}},
\bauthor{\bsnm{Roberts}, \binits{A.}},
\bauthor{\bsnm{Brown}, \binits{T.}},
\bauthor{\bsnm{Song}, \binits{D.}},
\bauthor{\bsnm{Erlingsson}, \binits{U.}}, \betal:
\bctitle{Extracting training data from large language models}.
In: \bbtitle{30th USENIX Security Symposium (USENIX Security 21)},
pp. \bfpage{2633}--\blpage{2650}
(\byear{2021})
\end{bchapter}
\endbibitem

\bibitem[\protect\citeauthoryear{Fredrikson et~al.}{2015}]{fredrikson2015model}
\begin{bchapter}
\bauthor{\bsnm{Fredrikson}, \binits{M.}},
\bauthor{\bsnm{Jha}, \binits{S.}},
\bauthor{\bsnm{Ristenpart}, \binits{T.}}:
\bctitle{Model inversion attacks that exploit confidence information and basic countermeasures}.
In: \bbtitle{Proceedings of the 22nd ACM SIGSAC Conference on Computer and Communications Security},
pp. \bfpage{1322}--\blpage{1333}
(\byear{2015})
\end{bchapter}
\endbibitem

\bibitem[\protect\citeauthoryear{Leboukh et~al.}{2023}]{leboukh2023balancing}
\begin{barticle}
\bauthor{\bsnm{Leboukh}, \binits{F.}},
\bauthor{\bsnm{Aduku}, \binits{E.B.}},
\bauthor{\bsnm{Ali}, \binits{O.}}:
\batitle{Balancing chatgpt and data protection in germany: challenges and opportunities for policy makers}.
\bjtitle{Journal of Politics and Ethics in New Technologies and AI}
\bvolume{2}(\bissue{1}),
\bfpage{35166}--\blpage{35166}
(\byear{2023})
\end{barticle}
\endbibitem

\bibitem[\protect\citeauthoryear{Falade}{2023}]{falade2023decoding}
\begin{botherref}
\oauthor{\bsnm{Falade}, \binits{P.V.}}:
Decoding the threat landscape: Chatgpt, fraudgpt, and wormgpt in social engineering attacks.
arXiv preprint arXiv:2310.05595
(2023)
\end{botherref}
\endbibitem

\bibitem[\protect\citeauthoryear{Amos}{2023}]{amos2023fraudgpt}
\begin{botherref}
\oauthor{\bsnm{Amos}, \binits{Z.}}:
What is fraudgpt?
(2023)
\end{botherref}
\endbibitem

\bibitem[\protect\citeauthoryear{Delley}{2023}]{delley2023wormgpt}
\begin{barticle}
\bauthor{\bsnm{Delley}, \binits{D.}}:
\batitle{Wormgpt--the generative ai tool cybercriminals are using to launch business email compromise attacks}.
\bjtitle{SlashNext. Retrieved August}
\bvolume{24},
\bfpage{2023}
(\byear{2023})
\end{barticle}
\endbibitem

\bibitem[\protect\citeauthoryear{Saxena et~al.}{2023}]{saxena2023missed}
\begin{bchapter}
\bauthor{\bsnm{Saxena}, \binits{N.A.}},
\bauthor{\bsnm{Zhang}, \binits{W.}},
\bauthor{\bsnm{Shahabi}, \binits{C.}}:
\bctitle{Missed opportunities in fair ai}.
In: \bbtitle{Proceedings of the 2023 SIAM International Conference on Data Mining (SDM)},
pp. \bfpage{961}--\blpage{964}
(\byear{2023}).
\bcomment{SIAM}
\end{bchapter}
\endbibitem

\bibitem[\protect\citeauthoryear{Wang et~al.}{2024a}]{wang2024individual}
\begin{bchapter}
\bauthor{\bsnm{Wang}, \binits{Z.}},
\bauthor{\bsnm{Dzuong}, \binits{J.}},
\bauthor{\bsnm{Yuan}, \binits{X.}},
\bauthor{\bsnm{Chen}, \binits{Z.}},
\bauthor{\bsnm{Wu}, \binits{Y.}},
\bauthor{\bsnm{Yao}, \binits{X.}},
\bauthor{\bsnm{Zhang}, \binits{W.}}:
\bctitle{Individual fairness with group awareness under uncertainty}.
In: \bbtitle{Joint European Conference on Machine Learning and Knowledge Discovery in Databases}
(\byear{2024}).
\bcomment{Springer Nature Switzerland}
\end{bchapter}
\endbibitem

\bibitem[\protect\citeauthoryear{Wang et~al.}{2024b}]{wang2024toward}
\begin{botherref}
\oauthor{\bsnm{Wang}, \binits{Z.}},
\oauthor{\bsnm{Qiu}, \binits{M.}},
\oauthor{\bsnm{Chen}, \binits{M.}},
\oauthor{\bsnm{Salem}, \binits{M.B.}},
\oauthor{\bsnm{Yao}, \binits{X.}},
\oauthor{\bsnm{Zhang}, \binits{W.}}:
Toward fair graph neural networks via real counterfactual samples.
Knowledge and Information Systems,
1--25
(2024)
\end{botherref}
\endbibitem

\bibitem[\protect\citeauthoryear{Wang et~al.}{2023a}]{wang2023fg2an}
\begin{bchapter}
\bauthor{\bsnm{Wang}, \binits{Z.}},
\bauthor{\bsnm{Wallace}, \binits{C.}},
\bauthor{\bsnm{Bifet}, \binits{A.}},
\bauthor{\bsnm{Yao}, \binits{X.}},
\bauthor{\bsnm{Zhang}, \binits{W.}}:
\bctitle{Fg$^2$an: Fairness-aware graph generative adversarial networks}.
In: \bbtitle{Joint European Conference on Machine Learning and Knowledge Discovery in Databases},
pp. \bfpage{259}--\blpage{275}
(\byear{2023}).
\bcomment{Springer Nature Switzerland}
\end{bchapter}
\endbibitem

\bibitem[\protect\citeauthoryear{Wang et~al.}{2023b}]{wang2023preventing}
\begin{bchapter}
\bauthor{\bsnm{Wang}, \binits{Z.}},
\bauthor{\bsnm{Saxena}, \binits{N.}},
\bauthor{\bsnm{Yu}, \binits{T.}},
\bauthor{\bsnm{Karki}, \binits{S.}},
\bauthor{\bsnm{Zetty}, \binits{T.}},
\bauthor{\bsnm{Haque}, \binits{I.}},
\bauthor{\bsnm{Zhou}, \binits{S.}},
\bauthor{\bsnm{Kc}, \binits{D.}},
\bauthor{\bsnm{Stockwell}, \binits{I.}},
\bauthor{\bsnm{Bifet}, \binits{A.}}, \betal:
\bctitle{Preventing discriminatory decision-making in evolving data streams}.
In: \bbtitle{Proceedings of the 2023 ACM Conference on Fairness, Accountability, and Transparency (FAccT)}
(\byear{2023})
\end{bchapter}
\endbibitem

\bibitem[\protect\citeauthoryear{Chu et~al.}{2024}]{chu2024fairness}
\begin{botherref}
\oauthor{\bsnm{Chu}, \binits{Z.}},
\oauthor{\bsnm{Wang}, \binits{Z.}},
\oauthor{\bsnm{Zhang}, \binits{W.}}:
Fairness in large language models: A taxonomic survey.
ACM SIGKDD Explorations Newsletter, 2024,
34--48
(2024)
\end{botherref}
\endbibitem

\bibitem[\protect\citeauthoryear{Doan et~al.}{2024}]{doan2024fairness1}
\begin{bchapter}
\bauthor{\bsnm{Doan}, \binits{T.V.}},
\bauthor{\bsnm{Wang}, \binits{Z.}},
\bauthor{\bsnm{Nguyen}, \binits{M.N.}},
\bauthor{\bsnm{Zhang}, \binits{W.}}:
\bctitle{Fairness in large language models in three hours}.
In: \bbtitle{Proceedings of the 33rd ACM International Conference on Information \& Knowledge Management (Boise, USA)}
(\byear{2024})
\end{bchapter}
\endbibitem

\bibitem[\protect\citeauthoryear{Zhang and Weiss}{2023}]{zhang2023fairness}
\begin{botherref}
\oauthor{\bsnm{Zhang}, \binits{W.}},
\oauthor{\bsnm{Weiss}, \binits{J.C.}}:
Fairness with censorship and group constraints.
Knowledge and Information Systems,
1--24
(2023)
\end{botherref}
\endbibitem

\bibitem[\protect\citeauthoryear{Doan et~al.}{2024}]{doan2024fairness}
\begin{botherref}
\oauthor{\bsnm{Doan}, \binits{T.}},
\oauthor{\bsnm{Chu}, \binits{Z.}},
\oauthor{\bsnm{Wang}, \binits{Z.}},
\oauthor{\bsnm{Zhang}, \binits{W.}}:
Fairness definitions in language models explained.
arXiv preprint arXiv:2407.18454
(2024)
\end{botherref}
\endbibitem

\bibitem[\protect\citeauthoryear{Zhang}{2024}]{zhang2024inpractice}
\begin{botherref}
\oauthor{\bsnm{Zhang}, \binits{W.}}:
Ai fairness in practice: Paradigm, challenges, and prospects.
Ai Magazine
(2024)
\end{botherref}
\endbibitem

\bibitem[\protect\citeauthoryear{Bender et~al.}{2021}]{bender2021dangers}
\begin{bchapter}
\bauthor{\bsnm{Bender}, \binits{E.M.}},
\bauthor{\bsnm{Gebru}, \binits{T.}},
\bauthor{\bsnm{McMillan-Major}, \binits{A.}},
\bauthor{\bsnm{Shmitchell}, \binits{S.}}:
\bctitle{On the dangers of stochastic parrots: Can language models be too big?}
In: \bbtitle{Proceedings of the 2021 ACM Conference on Fairness, Accountability, and Transparency},
pp. \bfpage{610}--\blpage{623}
(\byear{2021})
\end{bchapter}
\endbibitem

\bibitem[\protect\citeauthoryear{Meade et~al.}{2021}]{meade2021empirical}
\begin{botherref}
\oauthor{\bsnm{Meade}, \binits{N.}},
\oauthor{\bsnm{Poole-Dayan}, \binits{E.}},
\oauthor{\bsnm{Reddy}, \binits{S.}}:
An empirical survey of the effectiveness of debiasing techniques for pre-trained language models.
arXiv preprint arXiv:2110.08527
(2021)
\end{botherref}
\endbibitem

\bibitem[\protect\citeauthoryear{Gallegos et~al.}{2024}]{gallegos2024bias}
\begin{botherref}
\oauthor{\bsnm{Gallegos}, \binits{I.O.}},
\oauthor{\bsnm{Rossi}, \binits{R.A.}},
\oauthor{\bsnm{Barrow}, \binits{J.}},
\oauthor{\bsnm{Tanjim}, \binits{M.M.}},
\oauthor{\bsnm{Kim}, \binits{S.}},
\oauthor{\bsnm{Dernoncourt}, \binits{F.}},
\oauthor{\bsnm{Yu}, \binits{T.}},
\oauthor{\bsnm{Zhang}, \binits{R.}},
\oauthor{\bsnm{Ahmed}, \binits{N.K.}}:
Bias and fairness in large language models: A survey.
Computational Linguistics,
1--79
(2024)
\end{botherref}
\endbibitem

\bibitem[\protect\citeauthoryear{Wang et~al.}{2024}]{wang2024advancing}
\begin{bchapter}
\bauthor{\bsnm{Wang}, \binits{Z.}},
\bauthor{\bsnm{Chu}, \binits{Z.}},
\bauthor{\bsnm{Blanco}, \binits{R.}},
\bauthor{\bsnm{Chen}, \binits{Z.}},
\bauthor{\bsnm{Chen}, \binits{S.-C.}},
\bauthor{\bsnm{Zhang}, \binits{W.}}:
\bctitle{Advancing graph counterfactual fairness through fair representation learning}.
In: \bbtitle{Joint European Conference on Machine Learning and Knowledge Discovery in Databases}
(\byear{2024}).
\bcomment{Springer Nature Switzerland}
\end{bchapter}
\endbibitem

\bibitem[\protect\citeauthoryear{Blodgett and O'Connor}{2017}]{blodgett2017racial}
\begin{botherref}
\oauthor{\bsnm{Blodgett}, \binits{S.L.}},
\oauthor{\bsnm{O'Connor}, \binits{B.}}:
Racial disparity in natural language processing: A case study of social media african-american english.
arXiv preprint arXiv:1707.00061
(2017)
\end{botherref}
\endbibitem

\bibitem[\protect\citeauthoryear{Mei et~al.}{2023}]{mei2023bias}
\begin{bchapter}
\bauthor{\bsnm{Mei}, \binits{K.}},
\bauthor{\bsnm{Fereidooni}, \binits{S.}},
\bauthor{\bsnm{Caliskan}, \binits{A.}}:
\bctitle{Bias against 93 stigmatized groups in masked language models and downstream sentiment classification tasks}.
In: \bbtitle{Proceedings of the 2023 ACM Conference on Fairness, Accountability, and Transparency},
pp. \bfpage{1699}--\blpage{1710}
(\byear{2023})
\end{bchapter}
\endbibitem

\bibitem[\protect\citeauthoryear{Dash et~al.}{2023}]{dash2023evaluation}
\begin{botherref}
\oauthor{\bsnm{Dash}, \binits{D.}},
\oauthor{\bsnm{Thapa}, \binits{R.}},
\oauthor{\bsnm{Banda}, \binits{J.M.}},
\oauthor{\bsnm{Swaminathan}, \binits{A.}},
\oauthor{\bsnm{Cheatham}, \binits{M.}},
\oauthor{\bsnm{Kashyap}, \binits{M.}},
\oauthor{\bsnm{Kotecha}, \binits{N.}},
\oauthor{\bsnm{Chen}, \binits{J.H.}},
\oauthor{\bsnm{Gombar}, \binits{S.}},
\oauthor{\bsnm{Downing}, \binits{L.}}, et al.:
Evaluation of gpt-3.5 and gpt-4 for supporting real-world information needs in healthcare delivery.
arXiv preprint arXiv:2304.13714
(2023)
\end{botherref}
\endbibitem

\bibitem[\protect\citeauthoryear{Pal et~al.}{2023}]{umapathi2023med}
\begin{bchapter}
\bauthor{\bsnm{Pal}, \binits{A.}},
\bauthor{\bsnm{Umapathi}, \binits{L.K.}},
\bauthor{\bsnm{Sankarasubbu}, \binits{M.}}:
\bctitle{Med-halt: Medical domain hallucination test for large language models}.
In: \bbtitle{Proceedings of the 27th Conference on Computational Natural Language Learning (CoNLL)},
pp. \bfpage{314}--\blpage{334}
(\byear{2023})
\end{bchapter}
\endbibitem

\bibitem[\protect\citeauthoryear{Dzuong et~al.}{2024}]{dzuong2024uncertain}
\begin{botherref}
\oauthor{\bsnm{Dzuong}, \binits{J.}},
\oauthor{\bsnm{Wang}, \binits{Z.}},
\oauthor{\bsnm{Zhang}, \binits{W.}}:
Uncertain boundaries: Multidisciplinary approaches to copyright issues in generative ai.
arXiv preprint arXiv:2404.08221
(2024)
\end{botherref}
\endbibitem

\bibitem[\protect\citeauthoryear{Yazdani et~al.}{2024}]{yazdani2024comprehensive}
\begin{botherref}
\oauthor{\bsnm{Yazdani}, \binits{S.}},
\oauthor{\bsnm{Saxena}, \binits{N.}},
\oauthor{\bsnm{Wang}, \binits{Z.}},
\oauthor{\bsnm{Wu}, \binits{Y.}},
\oauthor{\bsnm{Zhang}, \binits{W.}}:
A comprehensive survey of image and video generative ai: Recent advances, variants, and applications
(2024)
\end{botherref}
\endbibitem

\bibitem[\protect\citeauthoryear{Small}{2023}]{novelistLawsuit}
\begin{botherref}
\oauthor{\bsnm{Small}, \binits{Z.}}:
Sarah silverman sues openai and meta over copyright infringement.
The New York Times
(2023)
\end{botherref}
\endbibitem

\bibitem[\protect\citeauthoryear{Stempel}{2023}]{Stempel_2023}
\begin{botherref}
\oauthor{\bsnm{Stempel}, \binits{J.}}:
NY Times sues openai, Microsoft for infringing copyrighted works ...
Thomson Reuters Corporation
(2023).
\url{https://www.reuters.com/legal/transactional/ny-times-sues-openai-microsoft-infringing-copyrighted-work-2023-12-27/}
\end{botherref}
\endbibitem

\bibitem[\protect\citeauthoryear{Li et~al.}{2023}]{li2023protecting}
\begin{bchapter}
\bauthor{\bsnm{Li}, \binits{Z.}},
\bauthor{\bsnm{Wang}, \binits{C.}},
\bauthor{\bsnm{Wang}, \binits{S.}},
\bauthor{\bsnm{Gao}, \binits{C.}}:
\bctitle{Protecting intellectual property of large language model-based code generation apis via watermarks}.
In: \bbtitle{Proceedings of the 2023 ACM SIGSAC Conference on Computer and Communications Security},
pp. \bfpage{2336}--\blpage{2350}
(\byear{2023})
\end{bchapter}
\endbibitem

\end{thebibliography}

\end{document}